\documentclass[runningheads]{llncs}

\PassOptionsToPackage{table}{xcolor}
\usepackage{eccv}

\usepackage{eccvabbrv}

\usepackage{graphicx}
\usepackage{booktabs}

\usepackage{graphicx}
\usepackage{amsmath}
\usepackage{amssymb}
\usepackage{booktabs}
\usepackage{multirow}
\usepackage{comment}
\usepackage{algorithm}
\usepackage{algpseudocode}
\usepackage{subcaption}
\usepackage{float}

\usepackage{wrapfig}

\usepackage{todonotes}
\usepackage[normalem]{ulem}
\useunder{\uline}{\ul}{}
\usepackage{pgfplots}
\pgfplotsset{compat=1.18}
\usepackage{placeins}   % \FloatBarrier used in merged supplementary

\definecolor{best}{RGB}{144,238,144}      % Light green for best results
\definecolor{second}{RGB}{211,211,211}    % Light gray for second best

\newcommand{\crh}[1]{#1}% highlighting disabled (no-blue version)

\DeclareMathOperator*{\argmax}{arg\,max}

\usepackage[accsupp]{axessibility}  % Improves PDF readability for those with disabilities.

\usepackage{hyperref}

\usepackage{orcidlink}

\begin{document}

% ---------------------------------------------------------------
\title{SIMPLER: Efficient Foundation Model Adaptation via Similarity-Guided Layer Pruning for Earth Observation}

\titlerunning{SIMPLER}

\author{V{\'\i}ctor Barreiro\inst{1,2}\orcidlink{0009-0008-2167-2005} \and
Johannes Jakubik\inst{3}\orcidlink{0000-0002-6235-0300} \and
Francisco Arg{\"u}ello\inst{2}\orcidlink{0000-0001-9279-5426} \and
Dora B.~Heras\inst{1,2}\orcidlink{0000-0002-5304-1426}}

\authorrunning{V.~Barreiro et al.}

\institute{Centro Singular de Investigación en Tecnoloxías Intelixentes (CiTIUS), Universidade de Santiago de Compostela, Spain \and
Departmento de Electrónica e Computación, Universidade de Santiago de Compostela, Spain \and
IBM Research Europe, Zurich, Switzerland}

\maketitle

\begin{abstract}
Fine-tuning foundation models for Earth Observation is computationally expensive, with high training time and memory demands for both training and deployment. Parameter-efficient methods reduce training cost but retain full inference complexity, while post-hoc compression optimizes inference only after costly full fine-tuning. We introduce SIMPLER, a pre–fine-tuning architecture selection method that reduces inference and deployment costs by identifying an effective model depth before adaptation. SIMPLER exploits stabilization of representations in deeper layers of pre-trained vision transformers: it computes layer-wise representation similarity on unlabeled task data and applies an automated scoring function to select redundant layers, with no gradients, magnitude heuristics, or hyperparameter tuning required. On Prithvi-EO-2, SIMPLER prunes up to 79\% of parameters while retaining 94\% of baseline performance, yielding a 2.1× training speedup and 2.6× inference speedup. The method generalizes to TerraMind (a multimodal EO foundation model) and ImageNet-pretrained ViT-MAE, demonstrating applicability across tasks, architectures, and spectral modalities. \crh{Code is available at \url{https://gitlab.citius.gal/hpc4rs/simpler}.}

\end{abstract}

\section{Introduction}
\label{sec:intro}

\begin{figure}[t]
  \centering

  \begin{subfigure}[t]{\textwidth}
    \centering
    \includegraphics[width=0.7\linewidth]{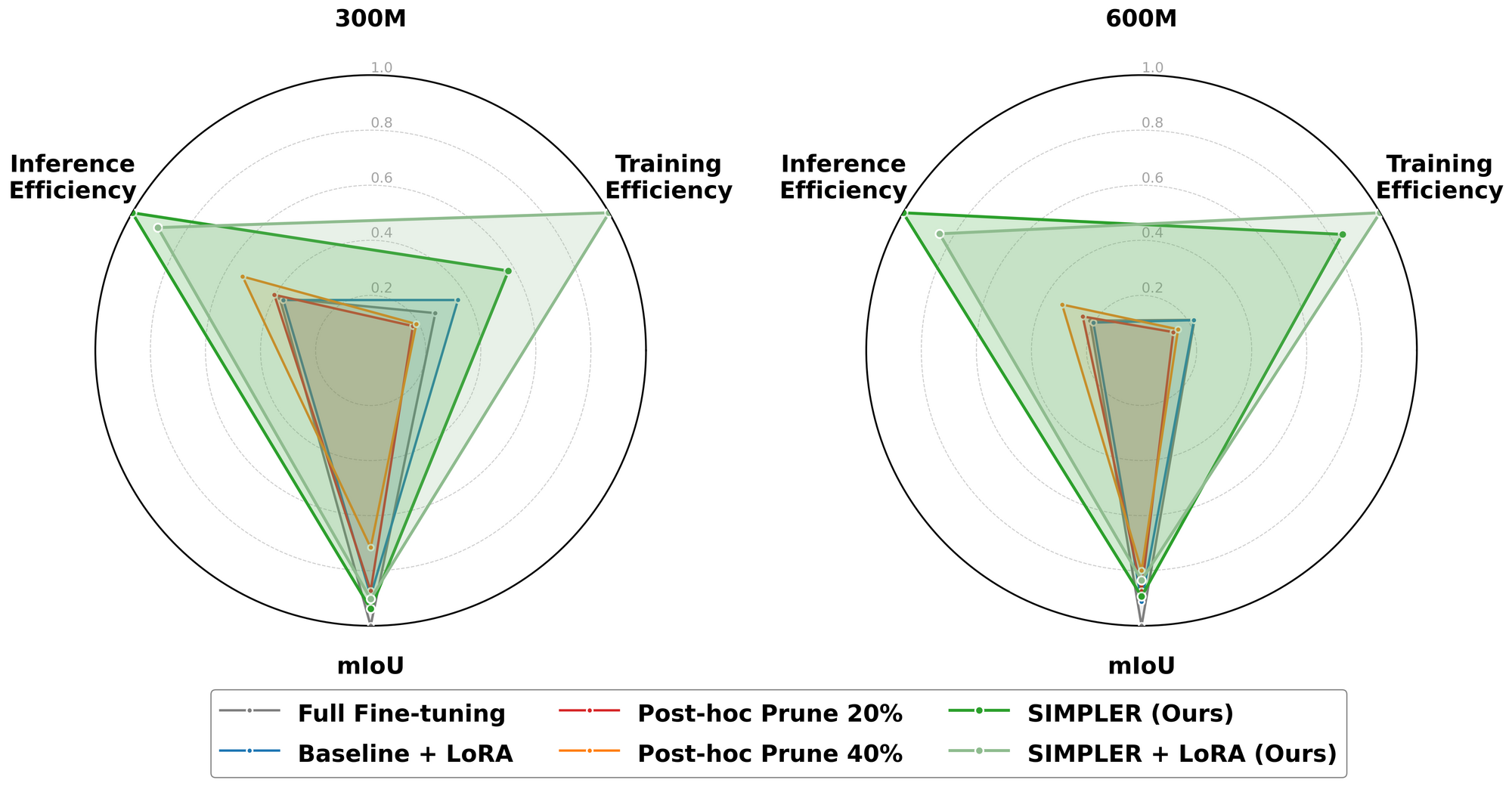}
    \caption{Trade-off between training/inference efficiency and mIoU on MADOS for Prithvi-EO-2.   Training and inference efficiency are computed as the reciprocal of their respective times (1/time),
   normalized to [0,1] %
   Higher values indicate better 
  efficiency. }
    \label{fig:radar}
  \end{subfigure}

  \begin{subfigure}[t]{\textwidth}
    \centering
    \includegraphics[width=\linewidth]{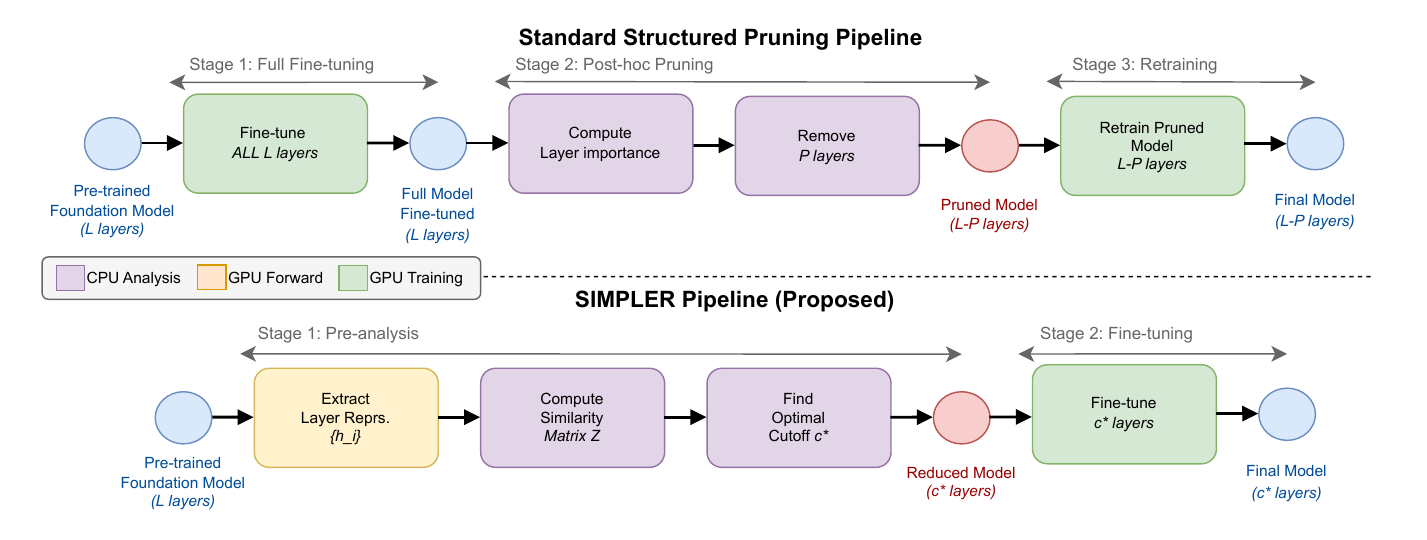}
    \caption{Complete pipeline from layer representation extraction to fine-tuning.}
    \label{fig:pipeline_overview}
  \end{subfigure}

  \caption{Overview of the proposed SIMPLER method. The upper radar plot summarizes the trade-off between training/inference efficiency and mIoU performance on the MADOS dataset. The lower panel illustrates the complete pipeline, from layer representation extraction to fine-tuning of the reduced model. }
  \label{fig:summary}
\end{figure}

Training and deploying large-scale Earth Observation (EO) foundation models imposes substantial computational costs~\cite{strubell2019energy}. For example, fine-tuning Prithvi-EO-2~\cite{prithvi2} (ViT~\cite{vit} model, 300M parameters) on BigEarthNetv2 dataset requires nearly 2.81 hours on a 4-GPU H200 cluster with 128 GB-VRAM total memory, while inference costs scale linearly with model depth, limiting deployment on satellites, drones, and edge devices that are critical for applications including disaster response and precision agriculture~\cite{geobench}. These constraints motivate efficiency research: reducing both training and inference costs without sacrificing the generalization capabilities that make foundation models~\cite{bommasani2021} valuable for Earth observation tasks~\cite{eo_models,satmae} (\cref{fig:radar}).

Existing solutions address only training or inference cost. Parameter-efficient fine-tuning methods constrain updates to low-rank subspaces or adapter modules, dramatically reducing training memory and time while leaving inference complexity unchanged; all transformer blocks remain active during deployment. Conversely, structured pruning compresses models for faster inference but operates post-hoc: practitioners must complete expensive full fine-tuning, analyze task-specific weight statistics, iteratively prune and retrain, and finally deploy (\cref{fig:pipeline_overview}). This sequential workflow incurs high computational cost before delivering efficiency gains and relies on post-adaptation parameters rather than pre-trained representational structure. No existing method reduces both training and inference costs in a unified manner.

We propose SIMPLER (\textbf{SIM}ilarity-based \textbf{P}arameter \textbf{L}ightweight \textbf{E}fficient \textbf{R}eduction) based on exploiting a key observation: deep layers in pre-trained vision transformers produce nearly identical representations when processing downstream task samples~\cite{see}, revealing redundancy before any adaptation occurs. SIMPLER (Fig. \ref{fig:pipeline_overview}) selects optimal architecture depth by computing layer-wise representation similarity on unlabeled task samples from the pre-trained model. Using Centered Kernel Alignment~\cite{cka}, we measure how each layer transforms input distributions, partition the similarity matrix at candidate cutoff points, and apply a scoring function  that balances representational diversity in retained layers against stability in pruned layers. This automatically identifies the cutoff without magnitude thresholds, gradient computation, or hyperparameter search, enabling architecture selection \textit{before} fine-tuning begins.

In particular, we make three contributions:
\begin{itemize}
\item We show that representation similarity on pre-trained features predicts post-fine-tuning layer importance. This fact is validated by ablation studies showing that pruned architectures retain full capacity when trained from scratch while pre-training provides gains of 42-43\%.
\item The proposed automated scoring criterion identifies optimal depth without hyperparameter tuning, with CKA-selected cutoffs (5 blocks, 94\% performance) substantially outperforming alternative metrics (2 blocks, 76\% performance). \item The approach generalizes across foundation models (Prithvi-EO-2, TerraMind, ViT-MAE~\cite{mae}), task types (segmentation, classification, time series), and spectral modalities (multispectral EO, RGB natural images).
\end{itemize}

\section{Related Work}
\label{sec:related}

\textbf{Parameter-efficient fine-tuning (PEFT)} methods like LoRA~\cite{lora}, adapters~\cite{adapters,adaptformer}, and visual prompts~\cite{vpt} constrain updates to low-rank subspaces or bottleneck modules, dramatically reducing training costs while maintaining task performance. However, these methods operate in weight space, assuming all layers contribute meaningfully. At inference, the full model depth remains active, offering no reduction in deployment costs.

\textbf{Model compression} addresses inference costs through complementary \linebreak strategies. Structured pruning~\cite{fang2023depgraph,nvit,xpruner,michel2019sixteen} removes entire architectural components (layers, attention heads, channels), achieving substantial compression while maintaining competitive performance on standard hardware. The critical limitation is timing~\cite{tay2022efficient}: structured pruning operates post-hoc, after completing expensive full fine-tuning. Practitioners must analyze task-specific weight statistics, iteratively prune and retrain, incurring full training costs upfront before efficiency gains are realized. Magnitude-based criteria can misidentify layer importance, removing layers with small weights that perform valuable transformations~\cite{lottery}.

Knowledge distillation~\cite{distillation,deit,tinyvit,vitkd} transfers knowledge from large teacher models to compact students, achieving impressive compression ratios but requiring training the new models from scratch. Quantization~\cite{ptq4vit,qvit,fqvit} reduces numerical precision, compressing models and accelerating inference through specialized hardware. While effective and orthogonal to our approach, quantization does not address architectural depth or training efficiency.

\textbf{Adaptive depth methods} enable flexible model capacity at deployment. LayerDrop~\cite{layerdrop} applies structured dropout during training, allowing arbitrary depth sub-networks to be extracted at inference without additional fine-tuning. Early exiting methods~\cite{branchynet,early_exit_vit,lgvit} attach auxiliary classifiers to intermediate layers, enabling samples to terminate inference early once a confidence criterion is met. While these approaches offer deployment flexibility, and potential inference savings, they increase training complexity and do not explicitly identify a single optimal architecture based on pre-trained representations prior to fine-tuning.

\textbf{Representation similarity metrics} provide tools for analyzing neural network internals. Centered Kernel Alignment (CKA)~\cite{cka} measures similarity through kernel methods with invariance to orthogonal transformations. \linebreak SVCCA~\cite{svcca} applies singular value decomposition before canonical correlation analysis. Jaccard similarity~\cite{similarity_survey} quantifies overlap in k-nearest neighbor graphs. These metrics reveal that vision transformers produce increasingly uniform representations in deeper layers~\cite{see,platonic,dino}\crh{, a deep-layer redundancy also reported for large language models~\cite{gromov2024unreasonable}}, as self-attention enables early global information aggregation while residual connections propagate features across depth. However, prior work uses these metrics exclusively for post-hoc analysis of trained models~\cite{nguyen2021wide}. The crucial distinction is the stage of analysis: representation similarity patterns visible in pre-trained models on downstream task data are never exploited to inform architecture decisions before fine-tuning begins.

\begin{table}[h!]
\centering
\caption{Comparison of adaptation strategies and computational cost.}
\label{tab:conceptual_comparison}
\resizebox{0.5\linewidth}{!}{
\begin{tabular}{@{}lccc@{}}
\toprule
\textbf{Method} & \textbf{Stage} & \textbf{Train Cost} & \textbf{Inf. Cost} \\ \midrule
LoRA & During & \textbf{Low} & Baseline \\
Post-hoc Pruning & After & High & \textbf{Low} \\
\textbf{SIMPLER} & \textbf{Before} & \textbf{Low} & \textbf{Low} \\ \bottomrule
\end{tabular}
}

\end{table}

\textbf{Adaptation Stage and Cost Comparison.} As shown in \Cref{tab:conceptual_comparison}, we compare methods according to the stage at which adaptation occurs (before, during, or after fine-tuning) and their training and inference costs; SIMPLER acts in pretrained representation space, yielding low train and inference cost while avoiding layer-level and weight-specific constraints.

\section{Methodology}
\label{sec:methodology}

SIMPLER selects optimal architecture depth for foundation model adaptation through representation similarity analysis before fine-tuning begins.

\subsection{Problem Formulation}

Given pre-trained model $\mathcal{F}_{\text{pre}}$ with $L$ layers and representations $\mathbf{h}_{\ell}$, downstream task $\mathcal{T}$ with dataset $\mathcal{D}$, we identify optimal cutoff $c^* \in \{2, \ldots, L-2\}$ minimizing depth while preserving data-relevant features:
\begin{equation}
c^* = \argmax_{c \in \{2, \ldots, L-2\}} \text{score}(c; \mathcal{S}, \mathcal{F}_{\text{pre}})
\end{equation}
where $\mathcal{S} \subset \mathcal{D}$ is a small unlabeled sample set. We then fine-tune $\mathcal{F}_{c^*}$ (layers $1$ to $c^*$) with data head $\mathcal{H}_{\mathcal{D}}$, avoiding full model adaptation costs.

\subsection{Representation Similarity Metrics}

We compute layer similarity matrix $\mathbf{Z} \in \mathbb{R}^{L \times L}$ where $Z_{i,j}$ quantifies redundancy between layers $i$ and $j$. We evaluate three metrics: \textbf{CKA}~\cite{cka} (invariant to orthogonal transforms, $\mathcal{O}(N^2 D^2)$ complexity); \textbf{Jaccard}~\cite{similarity_survey} (nearest-neighbor overlap with parameter $k$); \textbf{SVCCA}~\cite{svcca} (SVD-based canonical correlation, $\mathcal{O}(ND^2)$ complexity). While our approach is metric-agnostic, CKA provides the most consistent results across datasets and architectures (see Suppl.~Mat.~Sec.~A for detailed analysis).

\textbf{Motivation.} Pre-trained representation similarity predicts post-fine-tuning layer importance through two complementary mechanisms. When consecutive layers produce highly similar representations on downstream task data, their transformations are redundant for that distribution. During fine-tuning via gradient descent, the loss landscape exhibits flat directions along these aligned representations, making it difficult to differentiate them through task-specific learning. Consequently, one layer captures the essential transformation while other layers contribute marginally.
We empirically validate this hypothesis in \cref{sec:experiments} through ablations showing that pruned architectures retain full capacity when trained from scratch (\cref{tab:ablation_300m}).

This contrasts with magnitude-based pruning, which risks discarding layers with small weights that perform valuable transformations, and gradient-based pruning, which requires expensive backpropagation and estimates task-specific loss sensitivity rather than intrinsic representational redundancy.

\begin{figure}[t]
    \centering
    \includegraphics[width=\linewidth]{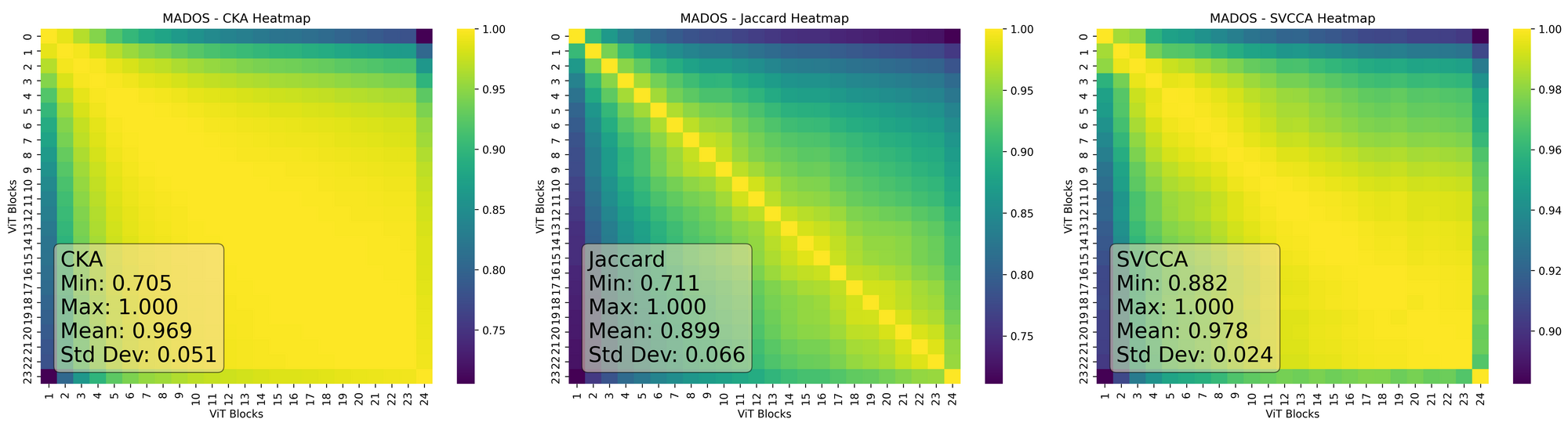}
    \caption{Similarity metrics comparison (CKA, Jaccard, SVCCA) on MADOS dataset (weak semantic segmentation) with Prithvi-EO-2 300M. Higher values (yellow) indicate greater similarity between layer representations, while lower values (blue) indicate greater divergence.}
    \label{fig:mados-metrics}
\end{figure}

\subsection{Automated Layer Selection}
We partition similarity matrix $\mathbf{Z}$ at cutoff $c$ into
$\mathbf{Z}_{TL} = \mathbf{Z}[0{:}c-1, 0{:}c-1]$ (retained layers, $c \times c$) and
$\mathbf{Z}_{BR} = \mathbf{Z}[c{:}L, c{:}L]$  (pruned layers, $(L{-}c) \times (L{-}c)$).
For a square block $M \in \mathbb{R}^{k \times k}$, let $\delta(M) = \frac{1}{(k{-}1)\,k}\sum_{i=0}^{k-2}\sum_{j=0}^{k-1}|M_{i,j} - M_{i+1,j}|$ denote the mean absolute consecutive-row difference.
Variability measures are then:
\begin{equation}
    \Delta_{TL} = \delta(\mathbf{Z}_{TL}), \qquad \Delta_{BR} = \delta(\mathbf{Z}_{BR})
\end{equation}
where $\Delta_{TL}$ captures diversity in retained layers (high indicates rich features) and $\Delta_{BR}$ captures stability in pruned layers (low indicates redundancy). The optimal cutoff maximizes $c^* = \argmax_{c \in \{2,\ldots,L-2\}} (\Delta_{TL} - \Delta_{BR})$, requiring no hyperparameter tuning. \textbf{Procedure:} (1) Extract representations $\{\mathbf{h}_1, \ldots, \mathbf{h}_L\}$ from $\mathcal{S}$; (2) Compute $\mathbf{Z}$; (3) Select $c^*$ maximizing score; (4) Fine-tune $\mathcal{F}_{c^*}$ with head $\mathcal{H}_{\mathcal{D}}$. Complete pseudocode is provided in Suppl.~Mat.~Algorithm~S1.

\subsection{Fine-Tuning Strategies}

Once optimal architecture $\mathcal{F}_{c^*}$ has been selected via representation similarity analysis, any fine-tuning strategy can be applied to adapt the reduced model to the downstream task. SIMPLER operates in architecture space (layer selection) rather than weight space (parameter updates), making it orthogonal and complementary to parameter-efficient fine-tuning methods. In our experiments, we evaluate two strategies: \textbf{(1) Full fine-tuning} trains all parameters of $\mathcal{F}_{c^*}$ and task head $\mathcal{H}_{\mathcal{T}}$ to maximize performance on the downstream task; \textbf{(2) SIMPLER with LoRA} applies low-rank adaptation~\cite{lora} to the reduced encoder $\mathcal{F}_{c^*}$, achieving compound efficiency gains by reducing both architectural depth (fewer layers) and trainable parameters (low-rank updates). SIMPLER is compatible with other PEFT methods such as adapters~\cite{adapters}, prefix tuning~\cite{li2021prefix}, or prompt tuning~\cite{lester-etal-2021-power}, as architecture selection does not constrain weight-space optimization strategies.

An important practical advantage over post-hoc structured pruning is that SIMPLER produces standard dense models avoiding the need for specialized sparse inference libraries. The reduced architecture can be deployed directly on any standard PyTorch or TensorFlow runtimes without any modification, facilitating  integration across HPC, cloud, and edge environments.

\section{Experiments}
\label{sec:experiments}

We conduct comprehensive experiments to evaluate SIMPLER across multiple downstream tasks, datasets, and model configurations. Our experimental design addresses three key objectives: (1)~confirming that similarity patterns in EO foundation models exhibit the representation stabilization observed in prior work; (2)~benchmarking SIMPLER against existing techniques in terms of performance-efficiency trade-offs; and (3)~validating the generalization of the approach across different foundation model architectures.

\subsection{Experimental Setup}

\textbf{Tasks and Datasets.} We evaluate SIMPLER across three diverse tasks: \textit{semantic segmentation} on MADOS~\cite{mados} for Marine Debris/Oil Spill detection with weak supervision; \textit{multi-label classification} on BigEarthNetv2~\cite{bigearthnetv2} (19 co-occurring land cover classes); \textit{time series analysis} on Sen4Map~\cite{sen4map} for land-cover classification. We primarily use Prithvi-EO-2~\cite{prithvi2} (300M and 600M parameters). The results are validated on TerraMind~\cite{terramind} for generalizability.

\textbf{Baselines.} We compare against: (1)~\textit{Full Fine-tuning} (all parameters); \linebreak (2)~\textit{LoRA}~\cite{lora} (encoder frozen; decoder and LoRA parameters trained; configuration details in Suppl.~Mat.~Sec.~F);
(3)~\textit{Post-hoc Structured Pruning}~\cite{li2017pruning,fang2023depgraph} (20\% and 40\% compression, magnitude-based, with retraining; reported times include full fine-tuning + pruning + retraining); (4)~\textit{SIMPLER with LoRA} (combining both approaches, only LoRA parameters and decoder are fine-tuned). \crh{An adaptive-depth baseline (LayerDrop~\cite{layerdrop}) is also reported in Suppl.\ Mat., Sec.~G.}

\textbf{Implementation Details.} We extract representations from 500 sampled images per dataset (see sample size analysis in \cref{fig:combined}). Jaccard uses $k=20$ neighbors. Cutoff selection maximizes $\text{score}(c) = \Delta_{TL} - \Delta_{BR}$ without hyperparameter tuning. Complete training hyperparameters, LoRA configurations, and computational requirements are provided in Suppl.~Mat.~Sec.~F.

\textbf{Evaluation Metrics.} We report task-specific performance (mIoU for segmentation, accuracy for classification) and efficiency metrics (total/trainable parameters, training/inference speedup, FLOPs), measured on identical hardware.

\subsection{Analysis of Similarity Metrics}

We begin by investigating whether representation stabilization patterns documented in vision transformers models~\cite{see} manifest in EO foundation models. All three similarity metrics (CKA, Jaccard, SVCCA) reveal distinct progressive stabilization of representations in Prithvi-EO-2 when computed on MADOS samples from the pre-trained model (\cref{fig:mados-metrics}). Early layers (0-5) exhibit low self-similarity and high inter-layer variability (indicated by the darker off-diagonal regions), reflecting the rapid evolution of features as the model processes low-level visual information. In contrast, the middle layers (6-15) show a gradual increase in similarity, serving as a transitional phase. Crucially, the deep layers (16-24) display high block-diagonal similarity (bright yellow regions in CKA and SVCCA), indicating that representations have stabilized and consecutive layers are performing redundant transformations. This varying redundancy profile across depth confirms the hypothesis of representation stabilization in EO foundation models and justifies the use of our layer selection strategy.

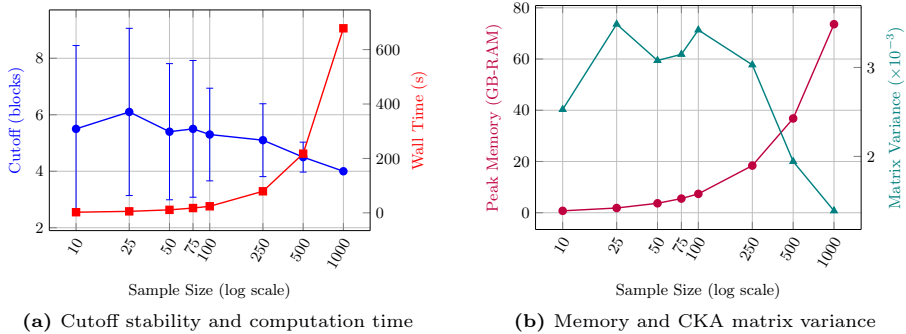
\begin{figure}[t!]
    \centering
    \begin{subfigure}{0.48\linewidth}
        \centering
        \resizebox{\linewidth}{!}{
        \begin{tikzpicture}
        \begin{axis}[
            xlabel={Sample Size (log scale)},
            ylabel={Cutoff (blocks)},
            ylabel style={blue},
            axis y line*=left,
            xmode=log,
            log basis x=10,
            xtick={10,25,50,75,100,250,500,1000},
            xticklabels={10,25,50,75,100,250,500,1000},
            xticklabel style={rotate=60, anchor=east, yshift=-4pt},
            width=8cm,
            height=6cm,
            legend style={at={(0.97,0.97)},anchor=north east,font=\footnotesize},
            grid=major,
            ymajorgrids=true,
            error bars/y dir=both,
            error bars/y explicit,
        ]
        \addplot[blue,thick,mark=*] coordinates {
            (10,5.50) +- (0,2.95)
            (25,6.10) +- (0,2.96)
            (50,5.40) +- (0,2.41)
            (75,5.50) +- (0,2.42)
            (100,5.30) +- (0,1.64)
            (250,5.10) +- (0,1.29)
            (500,4.50) +- (0,0.53)
            (1000,4.00) +- (0,0.00)
        };
        \end{axis}

        \begin{axis}[
            xlabel={Sample Size (log scale)},
            ylabel={Wall Time (s)},
            ylabel style={red},
            axis y line*=right,
            axis x line=none,
            xmode=log,
            log basis x=10,
            width=8cm,
            height=6cm,
        ]
        \addplot[red,thick,mark=square*] coordinates {
            (10,2.35)
            (25,5.47)
            (50,11.00)
            (75,17.62)
            (100,24.09)
            (250,79.24)
            (500,217.52)
            (1000,678.23)
        };
        \end{axis}
        \end{tikzpicture}
        }
        \caption{Cutoff stability and computation time}
        \label{fig:cutoff_walltime}
    \end{subfigure}
    \hfill
    \begin{subfigure}{0.48\linewidth}
        \centering
        \resizebox{\linewidth}{!}{ 
        \begin{tikzpicture}
        \begin{axis}[
            xlabel={Sample Size (log scale)},
            ylabel={Peak Memory (GB-RAM)},
            ylabel style={purple},
            axis y line*=left,
            xmode=log,
            log basis x=10,
            xtick={10,25,50,75,100,250,500,1000},
            xticklabels={10,25,50,75,100,250,500,1000},
            xticklabel style={rotate=60, anchor=east, yshift=-4pt},
            width=8cm,
            height=6cm,
            legend style={at={(0.03,0.97)},anchor=north west,font=\footnotesize},
            grid=major,
            ymajorgrids=true,
        ]
        \addplot[purple,thick,mark=*] coordinates {
            (10,0.736)
            (25,1.840)
            (50,3.680)
            (75,5.520)
            (100,7.359)
            (250,18.398)
            (500,36.797)
            (1000,73.594)
        };
        \end{axis}

        \begin{axis}[
            xlabel={Sample Size (log scale)},
            ylabel={Matrix Variance ($\times 10^{-3}$)},
            ylabel style={teal},
            axis y line*=right,
            axis x line=none,
            xmode=log,
            log basis x=10,
            width=8cm,
            height=6cm,
        ]
        \addplot[teal,thick,mark=triangle*] coordinates {
            (10,2.527)
            (25,3.486)
            (50,3.078)
            (75,3.147)
            (100,3.421)
            (250,3.029)
            (500,1.943)
            (1000,1.388)
        };
        \end{axis}
        \end{tikzpicture}
        }
        \caption{Memory and CKA matrix variance}
        \label{fig:variance_memory}
    \end{subfigure}
    \caption{Sample size sensitivity analysis for CKA computation on MADOS (Prithvi-300M). Left: Cutoff selection stabilizes at 500 samples (std=0.53, 5.6$\times$ reduction from 10 samples) with acceptable computation time (218s vs. 678s for 1000 samples). Right: Memory consumption scales linearly (36.8GB at 500 samples), while CKA variance remains stable. The 500-sample configuration provides optimal balance between cutoff stability, computational efficiency, and memory footprint. Notice that the computation time is in CPU and RAM, not VRAM}
    \label{fig:combined}
\end{figure}

\textbf{Sample size selection for CKA analysis.} We investigate the sensitivity of cutoff selection to the number of samples used for CKA computation (\cref{fig:combined}). With only 10 samples, the selected cutoff exhibits high variance (mean=5.5, std=2.95 blocks), leading to unstable layer selection across random draws. Increasing to 500 samples dramatically stabilizes selection (mean=4.5, std=0.53 blocks, 5.6$\times$ variance reduction), while maintaining practical computational cost (218s) and memory consumption (36.8GB RAM). Further increasing to 1000 samples yields full convergence (std=0.00) at 3.1$\times$ higher computational cost (678s) and 2$\times$ memory consumption (74GB-RAM). The matrix variance metric (right panel) remains consistent across sample sizes (mean $\approx$ 0.003), confirming that CKA similarity patterns converge with sufficient samples. Based on this empirical analysis, we use 500 samples for all experiments, achieving stable automated layer selection without excessive computational overhead.

\subsection{Performance-Efficiency Trade-offs}

We evaluate SIMPLER's effectiveness in balancing task performance with computational efficiency across three downstream tasks, demonstrating consistent compression with competitive accuracy.

\textbf{Semantic Segmentation (MADOS).} SIMPLER (\cref{tab:mados_results}) retains 94\% baseline performance (mIoU 62.8\% vs 66.9\%) while reducing the 300M Prithvi-EO-2 model to just 64.57M parameters (79\% reduction), achieving 2.1$\times$ training speedup and 2.6$\times$ inference speedup. Compared to LoRA (which reduces trainable parameters but maintains full inference architecture), SIMPLER provides 2.7$\times$ inference speedup at comparable performance. Combining SIMPLER with LoRA provides both benefits: 0.55M trainable parameters (0.2\% of original), fastest training (4.31 min), 90\% baseline performance.

\begin{table*}[t!]
\centering
\caption{Task 1 - Semantic Segmentation: Results on MADOS dataset comparing SIMPLER against baseline methods. Results show mean $\pm$ std over 5 runs. Best results highlighted with \colorbox{best}{light green}, second best with \colorbox{second}{light gray}.}
\label{tab:mados_results}
\resizebox{\textwidth}{!}{%
\begin{tabular}{@{}llccccccccc@{}}
\toprule
\multicolumn{3}{l}{}
& \multicolumn{3}{c}{\textbf{Training Cost}}
& \multicolumn{3}{c}{\textbf{Inference Cost}}
& \multicolumn{2}{c}{\textbf{Performance}}            \\ \midrule
\textbf{Model} & \textbf{Method} & \multicolumn{1}{c|}{\begin{tabular}[c]{@{}c@{}}Params\\ (M)\end{tabular}} & \begin{tabular}[c]{@{}c@{}}Train\\ (M)\end{tabular} & \begin{tabular}[c]{@{}c@{}}Time\\ (min)\end{tabular} & \multicolumn{1}{c|}{\begin{tabular}[c]{@{}c@{}}Mem\\ (GB)\end{tabular}} & \begin{tabular}[c]{@{}c@{}}FLOPs\\ (G)\end{tabular} & \begin{tabular}[c]{@{}c@{}}Thr.\\ (img/s)\end{tabular} & \multicolumn{1}{c|}{\begin{tabular}[c]{@{}c@{}}Inf\\ (s)\end{tabular}} & \begin{tabular}[c]{@{}c@{}}mIoU\\ (\%)\end{tabular} & \begin{tabular}[c]{@{}c@{}}Acc\\ (\%)\end{tabular} \\ \midrule
\multirow{6}{*}{\textbf{300M}} & Baseline                       & \multicolumn{1}{c|}{303.90} & 303.90 & 15.90$\pm$4.80 & \multicolumn{1}{c|}{11.70$\pm$0.47} & 238.50 & 33.02$\pm$1.94 & \multicolumn{1}{c|}{3.04$\pm$0.18} & \cellcolor{best}66.9$\pm$2.5 & \cellcolor{best}95.3$\pm$1.2 \\
 & Baseline with LoRA             & \multicolumn{1}{c|}{306.32} & \cellcolor{second}2.42 & 11.77$\pm$2.17 & \multicolumn{1}{c|}{8.76$\pm$0.50} & 238.50 & 31.62$\pm$1.13 & \multicolumn{1}{c|}{3.17$\pm$0.11} & 59.6$\pm$1.5 & 94.1$\pm$0.6 \\
 & Baseline + Prune 20\%          & \multicolumn{1}{c|}{240.92} & 240.92 & 24.34$\pm$5.09 & \multicolumn{1}{c|}{11.70$\pm$0.47} & 189.07 & 35.33$\pm$4.02 & \multicolumn{1}{c|}{2.87$\pm$0.36} & 58.4$\pm$1.6 & 93.2$\pm$2.0 \\
 & Baseline + Prune 40\%          & \multicolumn{1}{c|}{177.94} & 177.94 & 22.51$\pm$4.83 & \multicolumn{1}{c|}{11.70$\pm$0.47} & 139.64 & 47.03$\pm$6.03 & \multicolumn{1}{c|}{2.16$\pm$0.25} & 47.9$\pm$3.7 & 87.2$\pm$3.6 \\
 & SIMPLER (Ours)                 & \multicolumn{1}{c|}{\cellcolor{best}64.57} & 64.57 & \cellcolor{second}7.46$\pm$1.62 & \multicolumn{1}{c|}{\cellcolor{second}2.83$\pm$0.08} & \cellcolor{best}50.67 & \cellcolor{best}88.72$\pm$15.04 & \multicolumn{1}{c|}{\cellcolor{best}1.16$\pm$0.21} & \cellcolor{second}62.8$\pm$1.2 & \cellcolor{second}94.2$\pm$1.1 \\
 & SIMPLER with LoRA (Ours)       & \multicolumn{1}{c|}{\cellcolor{second}65.12} & \cellcolor{best}0.55 & \cellcolor{best}4.31$\pm$0.28 & \multicolumn{1}{c|}{\cellcolor{best}2.46$\pm$0.10} & \cellcolor{best}50.67 & \cellcolor{second}79.51$\pm$14.44 & \multicolumn{1}{c|}{\cellcolor{second}1.30$\pm$0.21} & 60.4$\pm$1.4 & 91.8$\pm$1.2 \\
\midrule
\multirow{6}{*}{\textbf{600M}} & Baseline                       & \multicolumn{1}{c|}{631.21} & 631.21 & 29.20$\pm$8.11 & \multicolumn{1}{c|}{25.06$\pm$1.09} & 646.85 & 16.28$\pm$0.31 & \multicolumn{1}{c|}{6.15$\pm$0.12} & \cellcolor{best}69.6$\pm$2.2 & \cellcolor{second}94.1$\pm$1.5 \\
 & Baseline with LoRA             & \multicolumn{1}{c|}{635.20} & \cellcolor{second}3.99 & 29.55$\pm$4.85 & \multicolumn{1}{c|}{17.99$\pm$1.05} & 646.85 & 15.22$\pm$0.63 & \multicolumn{1}{c|}{6.58$\pm$0.28} & \cellcolor{second}63.5$\pm$1.5 & \cellcolor{best}95.1$\pm$0.3 \\
 & Baseline + Prune 20\%          & \multicolumn{1}{c|}{513.14} & 513.14 & 48.75$\pm$10.08 & \multicolumn{1}{c|}{25.06$\pm$1.09} & 525.86 & 18.62$\pm$1.15 & \multicolumn{1}{c|}{5.39$\pm$0.34} & 60.8$\pm$5.5 & 92.8$\pm$2.7 \\
 & Baseline + Prune 40\%          & \multicolumn{1}{c|}{375.40} & 375.40 & 42.34$\pm$8.84 & \multicolumn{1}{c|}{25.06$\pm$1.09} & 384.70 & 25.08$\pm$0.90 & \multicolumn{1}{c|}{3.99$\pm$0.14} & 55.6$\pm$2.0 & 90.2$\pm$3.5 \\
 & SIMPLER (Ours)                 & \multicolumn{1}{c|}{\cellcolor{best}80.24} & 80.24 & \cellcolor{second}7.70$\pm$1.90 & \multicolumn{1}{c|}{\cellcolor{second}3.97$\pm$0.16} & \cellcolor{best}82.22 & \cellcolor{best}77.26$\pm$11.27 & \multicolumn{1}{c|}{\cellcolor{best}1.33$\pm$0.24} & 62.2$\pm$2.0 & 90.5$\pm$3.4 \\
 & SIMPLER with LoRA (Ours)       & \multicolumn{1}{c|}{\cellcolor{second}80.79} & \cellcolor{best}0.55 & \cellcolor{best}6.49$\pm$1.18 & \multicolumn{1}{c|}{\cellcolor{best}3.27$\pm$0.12} & \cellcolor{best}82.22 & \cellcolor{second}64.92$\pm$9.56 & \multicolumn{1}{c|}{\cellcolor{second}1.57$\pm$0.19} & 58.1$\pm$1.3 & 88.8$\pm$2.1 \\
\bottomrule
\end{tabular}%
}
\end{table*}

Post-hoc pruning requires full fine-tuning + retraining (22-24 min for 300M vs SIMPLER's 7.46 min). While 40\% pruning achieves 72\% of baseline (mIoU 47.9\%), it shows higher variance ($\pm$3.7), suggesting sensitivity to magnitude-based criteria. For 600M, SIMPLER retains 89\% of baseline (mIoU 62.2\% vs 69.6\%) with 87\% parameter reduction (80.24M) and 3.8$\times$ training speedup, demonstrating scalability and the advantage of pre-fine-tuning architecture selection.

\textbf{Multi-label Classification (BigEarthNetv2).} On the more complex multi-label classification task with 19 co-occurring classes, SIMPLER achieves 83\% compression (51.98M parameters) while retaining 97\% baseline mAP (71.2\% vs 73.4\%), with 4.2$\times$ training speedup and 2.9$\times$ inference speedup (\cref{tab:bigearthnet_results}).

\begin{table*}[t!]
\centering
\caption{Task 2 - Multi-label Classification: Results on BigEarthNetv2 dataset comparing SIMPLER against baseline methods. Results show mean $\pm$ std over 5 runs. Best results are highlighted with \colorbox{best}{light green}, second best with \colorbox{second}{light gray}.}
\label{tab:bigearthnet_results}
\resizebox{\textwidth}{!}{%
\begin{tabular}{ @{}llrrrrrrrrlll @{}}
\toprule
\multicolumn{3}{c}{}                                                                                                                     & \multicolumn{3}{c}{\textbf{Training Cost}}                                                                                                                                           & \multicolumn{3}{c}{\textbf{Inference Cost}}                                                                                                                                           & \multicolumn{3}{c}{\textbf{Performance}}            \\ \midrule
\multicolumn{1}{c}{\textbf{Model}} & \multicolumn{1}{c}{\textbf{Method}} & \multicolumn{1}{c|}{\begin{tabular}[c]{@{}c@{}}Params\\ (M)\end{tabular}} & \multicolumn{1}{c}{\begin{tabular}[c]{@{}c@{}}Train\\ (M)\end{tabular}} & \multicolumn{1}{c}{\begin{tabular}[c]{@{}c@{}}Time\\ (min)\end{tabular}} & \multicolumn{1}{c|}{\begin{tabular}[c]{@{}c@{}}Mem\\ (GB)\end{tabular}} & \multicolumn{1}{c}{\begin{tabular}[c]{@{}c@{}}FLOPs\\ (G)\end{tabular}} & \multicolumn{1}{c}{\begin{tabular}[c]{@{}c@{}}Thr.\\ (img/s)\end{tabular}} & \multicolumn{1}{c|}{\begin{tabular}[c]{@{}c@{}}Inf\\ (s)\end{tabular}} & \multicolumn{1}{c}{\begin{tabular}[c]{@{}c@{}}mAP\\ (\%)\end{tabular}} & \multicolumn{1}{c}{\begin{tabular}[c]{@{}c@{}}F1-macro\\ (\%)\end{tabular}} & \multicolumn{1}{c}{\begin{tabular}[c]{@{}c@{}}F1-micro\\ (\%)\end{tabular}} \\ \midrule
\multirow{4}{*}{\textbf{300M}} & Baseline                       & \multicolumn{1}{r|}{303.91} & 303.91 & 168.87$\pm$6.60 & \multicolumn{1}{r|}{127.84$\pm$0.76} & 238.49 & 37.67$\pm$0.04 & \multicolumn{1}{r|}{2.65$\pm$0.00} & \cellcolor{best}73.4$\pm$0.5 & \cellcolor{best}66.8$\pm$0.7 & \cellcolor{best}77.7$\pm$0.2 \\
 & Baseline with LoRA             & \multicolumn{1}{r|}{306.33} & \cellcolor{second}2.42 & 230.65$\pm$17.90 & \multicolumn{1}{r|}{93.8$\pm$0.88} & 240.38 & 34.89$\pm$0.08 & \multicolumn{1}{r|}{2.87$\pm$0.01} & \cellcolor{second}72.2$\pm$0.4 & \cellcolor{second}66.3$\pm$0.7 & \cellcolor{second}77.6$\pm$0.1 \\
 & SIMPLER (Ours)                 & \multicolumn{1}{r|}{\cellcolor{best}51.98} & 51.98 & \cellcolor{best}40.66$\pm$1.64 & \multicolumn{1}{r|}{\cellcolor{second}23.76$\pm$0.00} & \cellcolor{best}40.78 & \cellcolor{best}110.05$\pm$0.36 & \multicolumn{1}{r|}{\cellcolor{best}0.91$\pm$0.00} & 71.2$\pm$0.3 & 65.3$\pm$0.3 & 76.4$\pm$0.2 \\
 & SIMPLER with LoRA (Ours)       & \multicolumn{1}{r|}{\cellcolor{second}52.43} & \cellcolor{best}0.45 & \cellcolor{second}92.29$\pm$16.33 & \multicolumn{1}{r|}{\cellcolor{best}19.04$\pm$0.40} & \cellcolor{second}41.12 & \cellcolor{second}105.11$\pm$0.42 & \multicolumn{1}{r|}{\cellcolor{second}0.95$\pm$0.00} & 70.1$\pm$0.2 & 64.2$\pm$0.4 & 76.5$\pm$0.1 \\
\bottomrule
\end{tabular}%
}
\end{table*}

For BigEarthNetv2 (300M), SIMPLER reduces the number of parameters to 17\% (51.98M) while retaining 97\% of baseline mAP (71.2\% vs 73.4\%), with 2.9$\times$ of inference speedup and 4.2$\times$ of training speedup. The slightly larger performance gap vs MADOS reflects increased task complexity (19 co-occurring land-cover classes). Combining SIMPLER with LoRA achieves 0.45M trainable parameters (0.1\% of original) with comparable performance (mAP 70.1\%), validating compatibility with parameter-efficient fine-tuning and automatically adapting model capacity to task complexity.

\textbf{Time Series Analysis (Sen4Map).} For time series land-cover classification, SIMPLER compresses the model by 70\% (89.76M parameters) while retaining 96\% baseline F1-macro (63.8\% vs 66.6\%), demonstrating 3.3$\times$ inference speedup and 2.4$\times$ training speedup (\cref{tab:sen4map_results}).

\begin{table*}[t!]
\centering
\caption{Task 3 - Time Series Classification: Results on Sen4Map dataset comparing SIMPLER against baseline methods. Results show mean $\pm$ std over 5 runs. Best results are highlighted with \colorbox{best}{light green}, second best with \colorbox{second}{light gray}.}
\label{tab:sen4map_results}
\resizebox{\textwidth}{!}{%
\begin{tabular}{ @{}llrrrrrrrrllll @{}}
\toprule
\multicolumn{3}{c}{}                                                                                                                     & \multicolumn{3}{c}{\textbf{Training Cost}}                                                                                                                                           & \multicolumn{3}{c}{\textbf{Inference Cost}}                                                                                                                                           & \multicolumn{4}{c}{\textbf{Performance}}            \\ \midrule
\multicolumn{1}{c}{\textbf{Model}} & \multicolumn{1}{c}{\textbf{Method}} & \multicolumn{1}{c|}{\begin{tabular}[c]{@{}c@{}}Params\\ (M)\end{tabular}} & \multicolumn{1}{c}{\begin{tabular}[c]{@{}c@{}}Train\\ (M)\end{tabular}} & \multicolumn{1}{c}{\begin{tabular}[c]{@{}c@{}}Time\\ (min)\end{tabular}} & \multicolumn{1}{c|}{\begin{tabular}[c]{@{}c@{}}Mem\\ (GB)\end{tabular}} & \multicolumn{1}{c}{\begin{tabular}[c]{@{}c@{}}FLOPs\\ (G)\end{tabular}} & \multicolumn{1}{c}{\begin{tabular}[c]{@{}c@{}}Thr.\\ (img/s)\end{tabular}} & \multicolumn{1}{c|}{\begin{tabular}[c]{@{}c@{}}Inf\\ (s)\end{tabular}} & \multicolumn{1}{c}{\begin{tabular}[c]{@{}c@{}}OA\\ (\%)\end{tabular}} & \multicolumn{1}{c}{\begin{tabular}[c]{@{}c@{}}AA\\ (\%)\end{tabular}} & \multicolumn{1}{c}{\begin{tabular}[c]{@{}c@{}}F1-macro\\ (\%)\end{tabular}} & \multicolumn{1}{c}{\begin{tabular}[c]{@{}c@{}}Kappa\\ (\%)\end{tabular}} \\ \midrule
\multirow{4}{*}{\textbf{300M}} & Baseline                       & \multicolumn{1}{r|}{303.90} & 303.90 & \cellcolor{second}133.81$\pm$8.48 & \multicolumn{1}{r|}{75.72$\pm$2.88} & 238.49 & 4.72$\pm$0.00 & \multicolumn{1}{r|}{21.16$\pm$0.01} & \cellcolor{second}75.4$\pm$0.2 & \cellcolor{second}65.0$\pm$0.4 & \cellcolor{second}66.6$\pm$0.3 & \cellcolor{second}69.7$\pm$0.3 \\
 & Baseline with LoRA             & \multicolumn{1}{r|}{306.31} & \cellcolor{second}2.41 & 930.42$\pm$257.38 & \multicolumn{1}{r|}{53.60$\pm$2.04} & 240.38 & 3.79$\pm$0.09 & \multicolumn{1}{r|}{26.40$\pm$0.65} & \cellcolor{best}75.8$\pm$0.3 & \cellcolor{best}65.1$\pm$0.7 & \cellcolor{best}66.9$\pm$0.6 & \cellcolor{best}70.1$\pm$0.4 \\
 & SIMPLER (Ours)                 & \multicolumn{1}{r|}{\cellcolor{best}89.76} & 89.76 & \cellcolor{best}55.62$\pm$3.83 & \multicolumn{1}{r|}{\cellcolor{second}23.28$\pm$1.08} & \cellcolor{best}70.44 & \cellcolor{best}15.55$\pm$0.03 & \multicolumn{1}{r|}{\cellcolor{best}6.43$\pm$0.01} & 73.6$\pm$0.1 & 61.8$\pm$0.7 & 63.8$\pm$0.4 & 67.4$\pm$0.1 \\
 & SIMPLER with LoRA (Ours)       & \multicolumn{1}{r|}{\cellcolor{second}90.50} & \cellcolor{best}0.74 & 398.17$\pm$40.71 & \multicolumn{1}{r|}{\cellcolor{best}17.72$\pm$0.56} & \cellcolor{second}71.01 & \cellcolor{second}12.32$\pm$0.08 & \multicolumn{1}{r|}{\cellcolor{second}8.12$\pm$0.05} & 73.9$\pm$0.2 & 61.5$\pm$0.6 & 63.8$\pm$0.4 & 67.6$\pm$0.2 \\
\bottomrule
\end{tabular}%
}
\end{table*}

For Sen4Map (300M), SIMPLER reduces parameters by 70\% (89.76M) while retaining 96\% of baseline F1-macro (63.8\% vs 66.6\%), with 3.3$\times$ inference speedup and 2.4$\times$ training speedup. The results demonstrate consistent parameter reduction (70-83\%) with 94-97\% performance retention across all three task types (segmentation, classification, time series), validating SIMPLER's broad applicability to diverse EO applications.

\textit{Performance Trade-offs:} For the time series task (Sen4Map), the baseline with LoRA achieves slightly superior performance (66.9\% F1-macro) compared to SIMPLER (63.8\% F1). However, this comes at the cost of 16.7$\times$ slower training (930.42 min vs 55.62 min) and offers \textit{zero} reduction in inference cost (FLOPs remain 240.38G). In contrast, SIMPLER provides a substantial reduction in both training time (2.4$\times$ speedup) and deployment costs (FLOPs reduced to 70.44G, a 3.4$\times$ decrease) while retaining 96\% of baseline performance. This trade-off reflects different optimization priorities: time series tasks with complex temporal dependencies may benefit from full-depth architectures when accuracy is paramount and inference constraints are relaxed, while SIMPLER is better suited for deployment-critical scenarios requiring both training efficiency and low-latency inference, such as on-device processing for disaster response or real-time edge monitoring. For resource-constrained scenarios requiring both efficient training and low-latency edge deployment, SIMPLER's combined efficiency advantages are decisive.

\subsection{Ablation Studies}

\textbf{Similarity Metric Comparison.} CKA outperforms Jaccard and SVCCA for layer selection (\cref{tab:ablation_300m}). While Jaccard/SVCCA select aggressive cutoff (layer 2, 9\% parameters), they severely degrade performance to 76\% baseline mIoU. CKA selects conservative cutoff (layer 5, 21\% parameters) while preserving 94\% of baseline performance, demonstrating superior task-relevant depth identification. The 3-layer difference yields 18\% performance gap, validating CKA as default metric. CKA's advantage arises from its invariance to orthogonal transformations and smooth gradient properties with respect to layer depth, enabling robust identification of semantic similarity rather than superficial feature alignment. In contrast, Jaccard's discrete nearest-neighbor structure and SVCCA's sensitivity to low-variance dimensions can lead to overly aggressive pruning that removes layers still contributing meaningful transformations. 
CKA's consistent performance across three model families (Prithvi-EO-2, TerraMind, ViT-MAE) and four downstream tasks demonstrates its generality for pre-fine-tuning layer selection. Linear probing analysis (Suppl.~Mat.~Sec.~E) provides complementary evidence: while Block 2 (Jaccard/SVCCA) achieves 33.00\% probe mIoU vs.~Block 5 (CKA) at 30.26\%, the fine-tuning results demonstrate that CKA's selection better captures task-relevant features during adaptation.

\textbf{Pre-training Impact Analysis.} SIMPLER's 5-block architecture (\cref{tab:ablation_300m}) achieves nearly identical from-scratch performance as the full 24-block baseline (mIoU 44.1\% vs 46.7\%), indicating removed layers contribute minimal architectural capacity. However, pre-training provides substantial gains (43\% and 42\% relative improvement), confirming SIMPLER preserves the most valuable pre-trained features. Linear probing on frozen representations (Suppl.~Mat.~Sec.~E) corroborates this finding: Block 6 achieves peak performance (34.03\% mIoU), validating that SIMPLER's automated selection of Block 5 captures semantically optimal depth without gradient-based search.

\begin{table}[t!]
\centering
\caption{Ablation Studies on MADOS (300M): (Top) Similarity metrics; (Bottom) Pre-training impact.}
\label{tab:ablation_300m}
\resizebox{0.7\textwidth}{!}{
\begin{tabular}{llcrcc}
\toprule
\textbf{Experiment} & \textbf{Configuration} & \textbf{Cutoff} & \textbf{Params (M)} & \textbf{mIoU (\%)} & \textbf{Acc (\%)} \\
\midrule
\multirow{3}{*}{\textbf{Metric}} & Baseline (Full) & 24 bl. & 303.90 & \cellcolor{best}66.9$\pm$2.5 & \cellcolor{best}95.3$\pm$1.2 \\
 & SIMPLER (Jaccard/SVCCA) & 2 bl. & \cellcolor{best}26.78 & 50.7$\pm$3.4 & 84.6$\pm$0.8 \\
 & SIMPLER (CKA) & 5 bl. & \cellcolor{second}64.57 & \cellcolor{second}62.8$\pm$1.2 & \cellcolor{second}94.2$\pm$1.1 \\
\midrule
\multirow{6}{*}{\textbf{Pre-training}} & Baseline (From Scratch) & 24 bl. & 303.90 & 46.7$\pm$2.4 & 83.0$\pm$0.8 \\
 & Baseline (Fine-tuning) & 24 bl. & 303.90 & 66.9$\pm$2.5 & 95.3$\pm$1.2 \\
 & Baseline (LoRA) & 24 bl. & 306.32 & 59.6$\pm$1.5 & 94.1$\pm$0.6 \\
 \cmidrule{2-6}
 & SIMPLER (From Scratch) & 5 bl. & 64.57 & 44.1$\pm$1.9 & 81.9$\pm$1.3 \\
 & SIMPLER (Fine-tuning) & 5 bl. & 64.57 & 62.8$\pm$1.2 & 94.2$\pm$1.1 \\
 & SIMPLER (LoRA) & 5 bl. & 65.12 & 60.4$\pm$1.4 & 91.8$\pm$1.2 \\
\bottomrule
\end{tabular}
}
\end{table}

\begin{figure}[ttp!]
\centering
\includegraphics[width=0.65\linewidth]{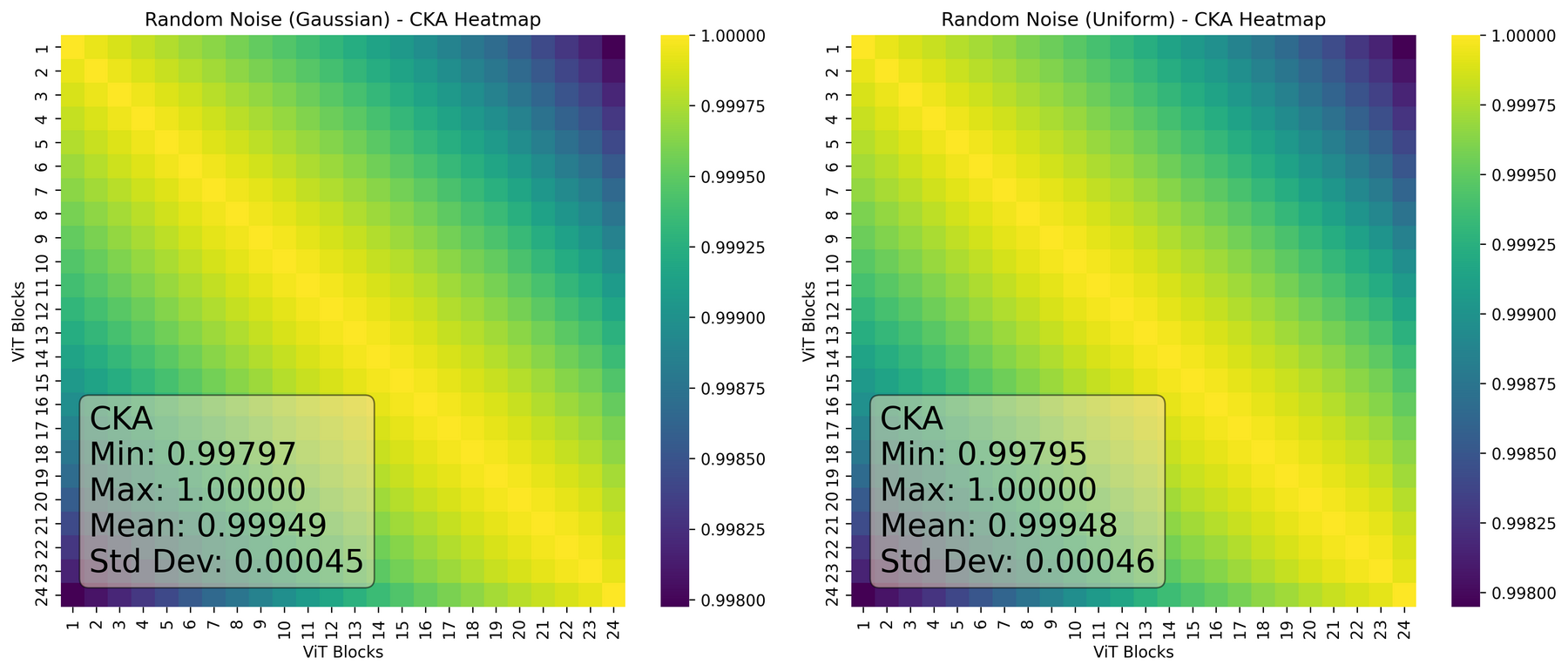}
\caption{CKA similarity for random noise (Gaussian, Uniform). Uniformly high similarity (0.998-1.000) with 115$\times$ narrower range vs. real data confirms learned features, not architectural artifacts.}
\label{fig:random_noise_cka}
\end{figure}

\textbf{Random Data Ablation.} Representation stabilization arises from learned features, not architectural artifacts (\cref{fig:random_noise_cka}). Random noise inputs produce uniformly high similarity (0.998-1.000) with 115$\times$ narrower range than real data (0.77-1.00), confirming SIMPLER exploits genuine learned hierarchies requiring downstream task samples.

\crh{\textbf{Block-Selection Ablation.} At matched depth, selecting random or last-$k$ blocks instead of the early blocks chosen by SIMPLER degrades performance to the from-scratch range, confirming the gains stem from retaining the transferable early layers rather than from depth reduction alone (Suppl.~Mat.~Sec.~G).}

\subsection{Generalization Across Foundation Models}

To validate generalization beyond Prithvi-EO-2, we evaluate SIMPLER on TerraMind~\cite{terramind}, a multimodal EO foundation model, across three scales: Large, Small, and Tiny  on MADOS (\cref{tab:mados_terramind_all_results}).

\begin{table*}[t!]
\centering
\caption{Task 1 - Semantic Segmentation: Results on MADOS dataset using TerraMind-Large, TerraMind-Small and TerraMind-Tiny models comparing SIMPLER against baseline methods. Results show mean $\pm$ std over 5 runs. Best results are highlighted with \colorbox{best}{light green}, second best with \colorbox{second}{light gray}.}
\label{tab:mados_terramind_all_results}
\resizebox{\textwidth}{!}{%
\begin{tabular}{ @{}llrrrrrrrrlll @{}}
\toprule
\multicolumn{3}{c}{}                                                                                                                     & \multicolumn{3}{c}{\textbf{Training Cost}}                                                                                                                                           & \multicolumn{3}{c}{\textbf{Inference Cost}}                                                                                                                                           & \multicolumn{2}{c}{\textbf{Performance}}            \\ \midrule
\multicolumn{1}{c}{\textbf{Model}} & \multicolumn{1}{c}{\textbf{Method}} & \multicolumn{1}{c|}{\begin{tabular}[c]{@{}c@{}}Params\\ (M)\end{tabular}} & \multicolumn{1}{c}{\begin{tabular}[c]{@{}c@{}}Train\\ (M)\end{tabular}} & \multicolumn{1}{c}{\begin{tabular}[c]{@{}c@{}}Time\\ (min)\end{tabular}} & \multicolumn{1}{c|}{\begin{tabular}[c]{@{}c@{}}Mem\\ (GB)\end{tabular}} & \multicolumn{1}{c}{\begin{tabular}[c]{@{}c@{}}FLOPs\\ (G)\end{tabular}} & \multicolumn{1}{c}{\begin{tabular}[c]{@{}c@{}}Thr.\\ (img/s)\end{tabular}} & \multicolumn{1}{c|}{\begin{tabular}[c]{@{}c@{}}Inf\\ (s)\end{tabular}} & \multicolumn{1}{c}{\begin{tabular}[c]{@{}c@{}}mIoU\\ (\%)\end{tabular}} & \multicolumn{1}{c}{\begin{tabular}[c]{@{}c@{}}Acc\\ (\%)\end{tabular}} \\ \midrule
\multirow{4}{*}{\textbf{TerraMind-L}} & Baseline                       & \multicolumn{1}{r|}{304.89} & 304.89 & 8.84$\pm$1.76 & \multicolumn{1}{r|}{8.38$\pm$0.51} & 61.69 & 124.91$\pm$10.20 & \multicolumn{1}{r|}{3.18$\pm$1.08} & \cellcolor{best}70.1$\pm$2.7 & \cellcolor{best}97.1$\pm$0.3 \\
 & Baseline with LoRA             & \multicolumn{1}{r|}{320.82} & \cellcolor{second}15.93 & 10.34$\pm$1.37 & \multicolumn{1}{r|}{6.11$\pm$0.50} & 64.81 & 69.73$\pm$3.10 & \multicolumn{1}{r|}{3.10$\pm$0.30} & \cellcolor{second}66.1$\pm$2.9 & \cellcolor{second}96.3$\pm$0.3 \\
 & SIMPLER (Ours)                 & \multicolumn{1}{r|}{\cellcolor{best}53.23} & 53.23 & \cellcolor{second}4.79$\pm$0.54 & \multicolumn{1}{r|}{\cellcolor{second}1.62$\pm$0.09} & \cellcolor{best}10.76 & \cellcolor{best}661.24$\pm$19.94 & \multicolumn{1}{r|}{\cellcolor{second}1.05$\pm$0.63} & 58.8$\pm$2.1 & 92.7$\pm$2.6 \\
 & SIMPLER with LoRA (Ours)       & \multicolumn{1}{r|}{\cellcolor{second}56.05} & \cellcolor{best}2.82 & \cellcolor{best}4.25$\pm$0.89 & \multicolumn{1}{r|}{\cellcolor{best}1.24$\pm$0.08} & \cellcolor{second}11.31 & \cellcolor{second}386.94$\pm$7.23 & \multicolumn{1}{r|}{\cellcolor{best}0.99$\pm$0.47} & 59.0$\pm$2.7 & 94.2$\pm$0.7 \\
\midrule
\multirow{4}{*}{\textbf{TerraMind-S}} & Baseline                       & \multicolumn{1}{r|}{22.37} & 22.37 & \cellcolor{second}6.48$\pm$2.01 & \multicolumn{1}{r|}{1.02$\pm$0.04} & 4.74 & 279.23$\pm$6.25 & \multicolumn{1}{r|}{1.06$\pm$0.37} & \cellcolor{second}53.0$\pm$1.1 & \cellcolor{second}87.9$\pm$1.9 \\
 & Baseline with LoRA             & \multicolumn{1}{r|}{26.07} & \cellcolor{second}3.70 & 10.57$\pm$3.18 & \multicolumn{1}{r|}{0.95$\pm$0.04} & 5.47 & 158.57$\pm$0.60 & \multicolumn{1}{r|}{1.14$\pm$0.14} & 40.4$\pm$7.2 & 86.7$\pm$4.3 \\
 & SIMPLER (Ours)                 & \multicolumn{1}{r|}{\cellcolor{best}9.96} & 9.96 & \cellcolor{best}6.06$\pm$1.15 & \multicolumn{1}{r|}{\cellcolor{second}0.55$\pm$0.02} & \cellcolor{best}2.10 & \cellcolor{best}588.62$\pm$7.43 & \multicolumn{1}{r|}{\cellcolor{best}0.76$\pm$0.20} & \cellcolor{best}53.6$\pm$3.7 & \cellcolor{best}88.9$\pm$1.6 \\
 & SIMPLER with LoRA (Ours)       & \multicolumn{1}{r|}{\cellcolor{second}11.59} & \cellcolor{best}1.63 & 8.18$\pm$1.87 & \multicolumn{1}{r|}{\cellcolor{best}0.52$\pm$0.02} & \cellcolor{second}2.42 & \cellcolor{second}338.56$\pm$3.46 & \multicolumn{1}{r|}{\cellcolor{second}0.81$\pm$0.19} & 41.3$\pm$4.7 & 85.7$\pm$4.0 \\
\midrule
\multirow{4}{*}{\textbf{TerraMind-T}} & Baseline                       & \multicolumn{1}{r|}{5.88} & 5.88 & \cellcolor{second}5.15$\pm$0.48 & \multicolumn{1}{r|}{0.53$\pm$0.01} & 1.33 & 236.13$\pm$30.85 & \multicolumn{1}{r|}{8.14$\pm$5.83} & \cellcolor{best}56.3$\pm$2.5 & \cellcolor{second}91.4$\pm$2.3 \\
 & Baseline with LoRA             & \multicolumn{1}{r|}{7.79} & \cellcolor{second}1.92 & 10.16$\pm$3.56 & \multicolumn{1}{r|}{0.54$\pm$0.02} & 1.71 & 151.15$\pm$12.20 & \multicolumn{1}{r|}{1.92$\pm$0.43} & 53.5$\pm$4.1 & \cellcolor{best}93.1$\pm$1.3 \\
 & SIMPLER (Ours)                 & \multicolumn{1}{r|}{\cellcolor{best}2.32} & 2.32 & \cellcolor{best}4.85$\pm$0.90 & \multicolumn{1}{r|}{\cellcolor{best}0.32$\pm$0.01} & \cellcolor{best}0.52 & \cellcolor{best}762.76$\pm$7.53 & \multicolumn{1}{r|}{\cellcolor{best}0.54$\pm$0.13} & \cellcolor{second}53.8$\pm$4.0 & 89.5$\pm$1.4 \\
 & SIMPLER with LoRA (Ours)       & \multicolumn{1}{r|}{\cellcolor{second}3.06} & \cellcolor{best}0.74 & 7.75$\pm$1.91 & \multicolumn{1}{r|}{\cellcolor{second}0.32$\pm$0.00} & \cellcolor{second}0.66 & \cellcolor{second}433.44$\pm$6.89 & \multicolumn{1}{r|}{\cellcolor{second}0.66$\pm$0.25} & 50.5$\pm$5.0 & 85.8$\pm$0.6 \\
\bottomrule
\end{tabular}%
}
\end{table*}

SIMPLER consistently achieves 55--83\% parameter reduction across all three TerraMind scales while retaining 84--101\% of baseline mIoU (\cref{tab:mados_terramind_all_results}). Two findings stand out. First, for TerraMind-Small, SIMPLER \textit{improves} over the baseline (mIoU 53.6\% vs 53.0\%) despite halving the parameters. Second, cross-scale comparison reveals that reducing a larger model with SIMPLER outperforms natively smaller architectures: TerraMind-Large reduced to 53.23M parameters (mIoU 58.8\%) surpasses TerraMind-Small baseline (22.37M, mIoU 53.0\%) by 5.8 points, indicating that richer representations learned during large-scale pre-training are preserved after layer selection. At the smallest scale, SIMPLER enables ultra-lightweight deployment (2.32M parameters, 762 img/s) while retaining 96\% baseline mIoU, opening practical avenues for satellite on-board processing. These results advocate for a ``reduce large'' strategy, selecting depth from a single well-trained foundation model, over training multiple smaller models independently. \crh{Beyond the optical modality, SIMPLER also generalizes to Sentinel-1 SAR (BigEarthNet-S1 with TerraMind-Large), retaining 93\% of baseline mAP at $\sim$12$\times$ fewer parameters (Suppl.~Mat.~Sec.~G).}

\textbf{Validation on RGB-Only Foundation Models (ViT-MAE + CIFAR-100).} Beyond multispectral EO data, SIMPLER generalizes to standard RGB vision transformers. On ViT-MAE (ImageNet pre-trained, CIFAR-100 fine-tuned), SIMPLER achieves 87\% parameter reduction (38.88M vs 303.40M) while retaining 82\% baseline accuracy, with 81.7\% memory reduction (9.06 GB-VRAM vs 49.59 GB-VRAM), 1.1$\times$ training speedup, 7.9$\times$ FLOPs reduction, and 6.9$\times$ inference throughput improvement (\cref{tab:vit_cifar100}).

\begin{table*}[t!]
\centering
\caption{Generalization validation on ViT-MAE (pre-trained on ImageNet) for CIFAR-100 classification. Results show mean $\pm$ std over 5 runs. Best results highlighted with \colorbox{best}{light green}, second best with \colorbox{second}{light gray}. Training from scratch (without pre-training) results are provided in Suppl.~Mat.~Sec.~B for comparison.}
\label{tab:vit_cifar100}
\resizebox{0.9\textwidth}{!}{%
\begin{tabular}{@{}lrrrrrrrrr@{}}
\toprule
 & \multicolumn{4}{c}{\textbf{Training Cost}} & \multicolumn{3}{c}{\textbf{Inference Cost}} & \multicolumn{1}{c}{\textbf{Performance}} \\ \midrule
\textbf{Method} & \multicolumn{1}{r}{\begin{tabular}[c]{@{}r@{}}Params\\ (M)\end{tabular}} & \begin{tabular}[c]{@{}r@{}}Train\\ (M)\end{tabular} & \begin{tabular}[c]{@{}r@{}}Time\\ (min)\end{tabular} & \multicolumn{1}{r|}{\begin{tabular}[c]{@{}r@{}}Mem\\ (GB-VRAM)\end{tabular}} & \begin{tabular}[c]{@{}r@{}}FLOPs\\ (G)\end{tabular} & \begin{tabular}[c]{@{}r@{}}Thr.\\ (img/s)\end{tabular} & \multicolumn{1}{r|}{\begin{tabular}[c]{@{}r@{}}Inf\\ (s)\end{tabular}} & Accuracy (\%) \\ \midrule
Baseline & \multicolumn{1}{r}{303.40} & 303.40 & 49.31$\pm$5.37 & \multicolumn{1}{r|}{49.59$\pm$1.46} & 59.70 & 119.11$\pm$1.19 & \multicolumn{1}{r|}{1.02$\pm$0.15} & \cellcolor{second}88.8$\pm$0.4 \\
Baseline with LoRA & \multicolumn{1}{r}{309.72} & \cellcolor{second}6.42 & 116.50$\pm$12.61 & \multicolumn{1}{r|}{57.75$\pm$0.59} & 60.94 & 88.78$\pm$0.49 & \multicolumn{1}{r|}{1.49$\pm$0.26} & \cellcolor{best}91.8$\pm$0.2 \\
SIMPLER (Ours) & \multicolumn{1}{r}{\cellcolor{best}38.88} & 38.88 & \cellcolor{second}42.97$\pm$1.37 & \multicolumn{1}{r|}{\cellcolor{best}9.06$\pm$1.02} & \cellcolor{best}7.60 & \cellcolor{best}822.51$\pm$3.48 & \multicolumn{1}{r|}{\cellcolor{best}0.22$\pm$0.01} & 72.8$\pm$0.3 \\
SIMPLER with LoRA (Ours) & \multicolumn{1}{r}{\cellcolor{second}40.51} & \cellcolor{best}1.73 & \cellcolor{best}28.51$\pm$11.57 & \multicolumn{1}{r|}{\cellcolor{second}9.62$\pm$0.13} & \cellcolor{second}7.92 & \cellcolor{second}583.11$\pm$26.54 & \multicolumn{1}{r|}{\cellcolor{second}0.78$\pm$0.73} & 67.9$\pm$2.1 \\ \bottomrule
\end{tabular}%
}
\end{table*}

For ViT-MAE on CIFAR-100, SIMPLER achieves 87\% parameter reduction while retaining 82\% baseline accuracy (72.8\% vs 88.8\%), with 81.7\% memory reduction and 6.9$\times$ inference throughput improvement. The from-scratch comparison (Suppl.~Mat.~Sec.~B) confirms SIMPLER preserves layers most enriched by pre-training rather than selecting based on architectural capacity alone. These results demonstrate that SIMPLER's representation-based layer selection generalizes beyond multispectral EO data to standard RGB vision transformers.

\subsection{Discussion}

SIMPLER exploits a key property of pre-trained vision transformers: consecutive layers producing highly similar representations on downstream task data perform redundant transformations for that distribution. Our ablation studies (\cref{tab:ablation_300m}) validate this principle empirically. The pruned 5-block architecture achieves nearly identical from-scratch performance as the full 24-block baseline (mIoU 44.1\% vs 46.7\%), confirming that removed layers contribute minimal architectural capacity. However, pre-training provides substantial gains (42-43\% relative improvement for both architectures), demonstrating that SIMPLER preserves layers most enriched with pre-trained features rather than merely selecting based on capacity.

The multi-scale TerraMind evaluation (\cref{tab:mados_terramind_all_results}) reinforces this finding: at every scale, a SIMPLER-reduced model retains or even surpasses the baseline of the next-smaller architecture, while inheriting the richer representations of the larger pre-training. This consistent cross-scale pattern suggests that investing in a single large-scale pre-training and deriving task-specific reduced models via SIMPLER is more cost-effective than training multiple smaller foundation models independently.

The method delivers greatest benefits when deployment efficiency is critical for edge devices, satellites, or resource-constrained environments requiring low-latency inference and both training and inference costs matter simultaneously. For scenarios prioritizing absolute accuracy where inference costs are unconstrained (e.g., cloud-based batch processing), parameter-efficient methods like LoRA may be preferable as they maintain full model depth. SIMPLER addresses the common case in Earth observation applications requiring on-board processing, where no existing method simultaneously reduces both training and deployment costs. \crh{These gains transfer to embedded hardware: on an NVIDIA Jetson Orin, SIMPLER delivers a $\sim$3.9$\times$ on-device inference speedup (Suppl.~Mat.~Sec.~G).}

Representation stabilization naturally emerges in foundation models trained via \crh{masked-autoencoding} self-supervised objectives. However, \crh{architectures employing contrastive (e.g., SoftCon~\cite{wanyan2024extending}) or explicit collapse-prevention (e.g., DINOv3~\cite{dinov3} with KoLeo regularization) objectives may instead exhibit oscillating or non-stabilizing similarity patterns} (Suppl.~Mat.~Sec.~C). Practitioners can verify applicability via CKA heatmap analysis on 500 task samples: clear block-diagonal structure in deep layers indicates suitability, while uniform or oscillating patterns suggest alternative approaches.

\section{Conclusion}
\label{sec:conclusions}

SIMPLER introduces a paradigm shift in foundation model compression: rather than pruning task-adapted weights post-hoc or assuming all layers are necessary during training, we analyze pre-trained representations on downstream task data to select architecture depth before fine-tuning begins. By computing layer-wise similarity through Centered Kernel Alignment, our automated scoring criterion identifies redundant depth without setting any hyperparameters, achieving 55--87\% parameter reduction while retaining 82--101\% performance across semantic segmentation, multi-label classification, and time series analysis on multispectral EO data, and RGB classification on natural images. Validation across diverse foundation models (Prithvi-EO-2, TerraMind, ViT-MAE) and datasets (MADOS, BigEarthNetv2, Sen4Map, CIFAR-100) establishes representation similarity as a predictive signal for layer importance that simultaneously reduces training costs, inference costs, and memory footprint, enabling deployment on resource-constrained hardware.

Multi-scale evaluation on TerraMind (Large, Small, Tiny) shows that \linebreak SIMPLER-reduced models consistently match or outperform natively smaller architectures, with depth reduction even acting as implicit regularization at smaller scales. These findings advocate for a ``reduce once'' paradigm: investing in a single large-scale pre-training and deriving task-specific reduced architectures via similarity-based layer selection, rather than training multiple smaller foundation models independently.

SIMPLER generalizes to foundation models trained with \crh{MAE-style} self-supervised objectives, which naturally exhibit the representation stabilization the method exploits. For models employing alternative pre-training strategies, practitioners can verify applicability through CKA heatmap analysis as detailed in Suppl.~Mat.~Sec.~C. \crh{Having validated SIMPLER across architectures, sensing modalities (multispectral, RGB, SAR), and task types, we leave its extension to paradigms without deep-layer stabilization, such as contrastive and collapse-prevention models, as the principal direction for future work (Suppl.\ Mat., Sec.~C).}

\section*{Acknowledgements}
\crh{This work was supported in part by grants PID2022--141623NB--I00 funded by MCIN{\slash}AEI{\slash}10.13039{\slash}501100011033 and by ``European Union NextGenerationEU{\slash}PRTR''. It was also supported by Xunta de Galicia - Conseller{\'\i}a de Cultura, Educaci{\'o}n, Formaci{\'o}n Profesional e Universidades [Centro de investigaci{\'o}n de Galicia accreditation 2024-2027 ED431G-2023{\slash}04 and Reference Competitive Group accreditation, ED431C-2022{\slash}16], and by ``ERDF{\slash}EU''. The work of V{\'\i}ctor Barreiro was supported by the Ministerio de Universidades, under a FPU Grant [grant number FPU2022-04364].}

\bibliographystyle{splncs04}
\bibliography{main}

\begin{thebibliography}{10}
\providecommand{\url}[1]{\texttt{#1}}
\providecommand{\urlprefix}{URL }
\providecommand{\doi}[1]{https://doi.org/#1}

\bibitem{alain2016understanding}
Alain, G., Bengio, Y.: Understanding intermediate layers using linear
  classifier probes. arXiv preprint arXiv:1610.01644  (2016).
  \doi{10.48550/arXiv.1610.01644}

\bibitem{early_exit_vit}
Bakhtiarnia, A., Zhang, Q., Iosifidis, A.: Single-layer vision transformers for
  more accurate early exits with less overhead. Neural Networks  \textbf{153},
  461--473 (2022). \doi{10.1016/j.neunet.2022.06.038}

\bibitem{bommasani2021}
Bommasani, R., Hudson, D.A., Adeli, E., Altman, R., Arora, S., von Arx, S.,
  et~al.: On the opportunities and risks of foundation models. arXiv preprint
  arXiv:2108.07258  (2021). \doi{10.48550/arXiv.2108.07258}

\bibitem{dino}
Caron, M., Touvron, H., Misra, I., J{\'e}gou, H., Mairal, J., Bojanowski, P.,
  Joulin, A.: Emerging properties in self-supervised vision transformers. In:
  IEEE/CVF International Conference on Computer Vision (ICCV). pp. 9650--9660
  (2021). \doi{10.1109/ICCV48922.2021.00951}

\bibitem{adaptformer}
Chen, S., Ge, C., Tong, Z., Wang, J., Song, Y., Wang, J., Luo, P.:
  {AdaptFormer}: Adapting vision transformers for scalable visual recognition.
  In: Advances in Neural Information Processing Systems (NeurIPS) (2022),
  \url{https://arxiv.org/abs/2205.13535}

\bibitem{bigearthnetv2}
Clasen, K.N., Hackel, L., Burgert, T., Sumbul, G., Demir, B., Markl, V.:
  {reBEN}: Refined {BigEarthNet} dataset for remote sensing image analysis. In:
  IEEE International Geoscience and Remote Sensing Symposium (IGARSS) (2025).
  \doi{10.5281/zenodo.10891137}

\bibitem{satmae}
Cong, Y., Khanna, S., Meng, C., Liu, P., Rozi, E., He, Y., Burke, M., Lobell,
  D.B., Ermon, S.: Satmae: pre-training transformers for temporal and
  multi-spectral satellite imagery. In: Proceedings of the 36th International
  Conference on Neural Information Processing Systems. NIPS '22 (2022)

\bibitem{vit}
Dosovitskiy, A., Beyer, L., Kolesnikov, A., Weissenborn, D., Zhai, X.,
  Unterthiner, T., Dehghani, M., Minderer, M., Heigold, G., Gelly, S.,
  Uszkoreit, J., Houlsby, N.: An image is worth 16x16 words: Transformers for
  image recognition at scale. In: International Conference on Learning
  Representations (ICLR) (2021),
  \url{https://openreview.net/forum?id=YicbFdNTTy}

\bibitem{layerdrop}
Fan, A., Grave, E., Joulin, A.: Reducing transformer depth on demand with
  structured dropout. In: International Conference on Learning Representations
  (2020), \url{https://openreview.net/forum?id=SylO2yStDr}

\bibitem{fang2023depgraph}
Fang, G., Ma, X., Song, M., Mi, M.B., Wang, X.: {DepGraph}: Towards any
  structural pruning. In: IEEE/CVF Conference on Computer Vision and Pattern
  Recognition (CVPR). pp. 16091--16101 (2023).
  \doi{10.1109/CVPR52729.2023.01544}

\bibitem{lottery}
Frankle, J., Carbin, M.: The lottery ticket hypothesis: Finding sparse,
  trainable neural networks. In: International Conference on Learning
  Representations (ICLR) (2019),
  \url{https://openreview.net/forum?id=rJl-b3RcF7}

\bibitem{gromov2024unreasonable}
Gromov, A., Tirumala, K., Shapourian, H., Glorioso, P., Roberts, D.A.: The
  unreasonable ineffectiveness of the deeper layers (2024)

\bibitem{mae}
He, K., Chen, X., Xie, S., Li, Y., Doll{\'a}r, P., Girshick, R.: Masked
  autoencoders are scalable vision learners. In: IEEE/CVF Conference on
  Computer Vision and Pattern Recognition (CVPR). pp. 15979--15986 (2022).
  \doi{10.1109/CVPR52688.2022.01553}

\bibitem{distillation}
Hinton, G., Vinyals, O., Dean, J.: Distilling the knowledge in a neural
  network. arXiv preprint arXiv:1503.02531  (2015).
  \doi{10.48550/arXiv.1503.02531}

\bibitem{adapters}
Houlsby, N., Giurgiu, A., Jastrzebski, S., Morrone, B., De~Laroussilhe, Q.,
  Gesmundo, A., Attariyan, M., Gelly, S.: Parameter-efficient transfer learning
  for {NLP}. In: Chaudhuri, K., Salakhutdinov, R. (eds.) Proceedings of the
  36th International Conference on Machine Learning. Proceedings of Machine
  Learning Research, vol.~97, pp. 2790--2799. PMLR (09--15 Jun 2019),
  \url{https://proceedings.mlr.press/v97/houlsby19a.html}

\bibitem{lora}
Hu, E.J., Shen, Y., Wallis, P., Allen-Zhu, Z., Li, Y., Wang, S., Wang, L.,
  Chen, W.: {LoRA}: Low-rank adaptation of large language models. In:
  International Conference on Learning Representations (ICLR) (2022),
  \url{https://openreview.net/forum?id=nZeVKeeFYf9}

\bibitem{platonic}
Huh, M., Cheung, B., Wang, T., Isola, P.: Position: The platonic representation
  hypothesis. In: International Conference on Machine Learning (ICML) (2024),
  \url{https://proceedings.mlr.press/v235/huh24a.html}

\bibitem{terramind}
Jakubik, J., Yang, F., Blumenstiel, B., Scheurer, E., Sedona, R.,
  Maurogiovanni, S., Bosmans, J., Dionelis, N., Marsocci, V., Kopp, N., et~al.:
  {TerraMind}: Large-scale generative multimodality for earth observation. In:
  IEEE/CVF International Conference on Computer Vision (ICCV) (2025)

\bibitem{vpt}
Jia, M., Tang, L., Chen, B.C., Cardie, C., Belongie, S., Hariharan, B., Lim,
  S.N.: Visual prompt tuning. In: European Conference on Computer Vision
  (ECCV). pp. 709--727. Springer (2022). \doi{10.1007/978-3-031-19827-4_41}

\bibitem{mados}
Kikaki, K., Kakogeorgiou, I., Hoteit, I., Karantzalos, K.: Detecting marine
  pollutants and sea surface features with deep learning in {Sentinel-2}
  imagery. ISPRS Journal of Photogrammetry and Remote Sensing  \textbf{210},
  39--57 (2024). \doi{10.1016/j.isprsjprs.2024.02.017}, mADOS dataset available
  at \url{https://marine-pollution.github.io/}

\bibitem{similarity_survey}
Klabunde, M., Schumacher, T., Strohmaier, M., Lemmerich, F.: Similarity of
  neural network models: A survey of functional and representational measures.
  ACM Comput. Surv.  \textbf{57}(9) (May 2025). \doi{10.1145/3728458},
  \url{https://doi.org/10.1145/3728458}

\bibitem{cka}
Kornblith, S., Norouzi, M., Lee, H., Hinton, G.: Similarity of neural network
  representations revisited. In: Chaudhuri, K., Salakhutdinov, R. (eds.)
  Proceedings of the 36th International Conference on Machine Learning.
  Proceedings of Machine Learning Research, vol.~97, pp. 3519--3529. PMLR
  (09--15 Jun 2019), \url{https://proceedings.mlr.press/v97/kornblith19a.html}

\bibitem{geobench}
Lacoste, A., Lehmann, N., Rodriguez, P., Sherwin, E.D., Kerner, H.,
  L{\"u}tjens, B., Irvin, J.A., Dao, D., Alemohammad, H., Drouin, A., et~al.:
  {GEO-Bench}: Toward foundation models for earth monitoring. In: Advances in
  Neural Information Processing Systems (NeurIPS) (2023),
  \url{https://arxiv.org/abs/2306.03831}

\bibitem{lester-etal-2021-power}
Lester, B., Al-Rfou, R., Constant, N.: The power of scale for
  parameter-efficient prompt tuning. In: Proceedings of the 2021 Conference on
  Empirical Methods in Natural Language Processing. pp. 3045--3059. Association
  for Computational Linguistics, Online and Punta Cana, Dominican Republic (Nov
  2021). \doi{10.18653/v1/2021.emnlp-main.243},
  \url{https://aclanthology.org/2021.emnlp-main.243/}

\bibitem{li2017pruning}
Li, H., Kadav, A., Durdanovic, I., Samet, H., Graf, H.P.: Pruning filters for
  efficient {ConvNets}. In: International Conference on Learning
  Representations (2017), \url{https://openreview.net/forum?id=rJqFGTslg}

\bibitem{li2021prefix}
Li, X.L., Liang, P.: Prefix-tuning: Optimizing continuous prompts for
  generation. arXiv preprint arXiv:2101.00190  (2021),
  \url{https://arxiv.org/abs/2101.00190}

\bibitem{qvit}
Li, Y., Xu, S., Zhang, B., Cao, X., Gao, P., Guo, G.: {Q-ViT}: Accurate and
  fully quantized low-bit vision transformer. In: Advances in Neural
  Information Processing Systems (2022),
  \url{https://proceedings.neurips.cc/paper_files/paper/2022/hash/deb921bff461a7b0a5c344a4871e7101-Abstract-Conference.html}

\bibitem{fqvit}
Lin, Y., Zhang, T., Sun, P., Li, Z., Zhou, S.: {FQ-ViT}: Post-training
  quantization for fully quantized vision transformer. In: Raedt, L.D. (ed.)
  Proceedings of the Thirty-First International Joint Conference on Artificial
  Intelligence, {IJCAI-22}. pp. 1173--1179. International Joint Conferences on
  Artificial Intelligence Organization (7 2022). \doi{10.24963/ijcai.2022/164},
  \url{https://doi.org/10.24963/ijcai.2022/164}, main Track

\bibitem{michel2019sixteen}
Michel, P., Levy, O., Neubig, G.: Are sixteen heads really better than one? In:
  Advances in Neural Information Processing Systems (NeurIPS). pp. 14014--14024
  (2019),
  \url{https://proceedings.neurips.cc/paper/2019/hash/2c601ad9d2ff9bc8b282670cdd54f69f-Abstract.html}

\bibitem{nguyen2021wide}
Nguyen, T., Raghu, M., Kornblith, S.: Do wide and deep networks learn the same
  things? {U}ncovering how neural network representations vary with width and
  depth. In: International Conference on Learning Representations (ICLR)
  (2021), \url{https://openreview.net/forum?id=KJNcAkY8tY4}

\bibitem{svcca}
Raghu, M., Gilmer, J., Yosinski, J., Sohl-Dickstein, J.N.: {SVCCA}: Singular
  vector canonical correlation analysis for deep learning dynamics and
  interpretability. In: Neural Information Processing Systems (2017).
  \doi{10.48550/arXiv.1706.05806}

\bibitem{see}
Raghu, M., Unterthiner, T., Kornblith, S., Zhang, C., Dosovitskiy, A.: Do
  vision transformers see like convolutional neural networks? In: Beygelzimer,
  A., Dauphin, Y., Liang, P., Vaughan, J.W. (eds.) Advances in Neural
  Information Processing Systems (2021),
  \url{https://openreview.net/forum?id=Gl8FHfMVTZu}

\bibitem{fvcore}
Research, F.: {fvcore}: A light-weight core library for pytorch.
  \url{https://github.com/facebookresearch/fvcore} (2019), accessed: 2024

\bibitem{sen4map}
Sharma, S., Sedona, R., Riedel, M., Cavallaro, G., Paris, C.: {Sen4Map}:
  Advancing mapping with {Sentinel-2} by providing detailed semantic
  descriptions and customizable land-use and land-cover data. IEEE Journal of
  Selected Topics in Applied Earth Observations and Remote Sensing
  \textbf{17},  13893--13907 (2024). \doi{10.1109/JSTARS.2024.3435081}

\bibitem{dinov3}
Sim{\'e}oni, O., Vo, H.V., Seitzer, M., Baldassarre, F., Oquab, M., Jose, C.,
  Khalidov, V., Szafraniec, M., Yi, S., Ramamonjisoa, M., Massa, F., Haziza,
  D., Wehrstedt, L., Wang, J., Darcet, T., Moutakanni, T., Sentana, L.,
  Roberts, C., Vedaldi, A., Tolan, J., Brandt, J., Couprie, C., Mairal, J.,
  J{\'e}gou, H., Labatut, P., Bojanowski, P.: {DINOv3}  (2025),
  \url{https://arxiv.org/abs/2508.10104}

\bibitem{strubell2019energy}
Strubell, E., Ganesh, A., McCallum, A.: Energy and policy considerations for
  deep learning in {NLP}. arXiv preprint arXiv:1906.02243  (2019).
  \doi{10.48550/arXiv.1906.02243}, seminal work on carbon footprint of training
  large models

\bibitem{prithvi2}
Szwarcman, D., Roy, S., Fraccaro, P., Þorsteinn Elí~Gíslason, Blumenstiel,
  B., Ghosal, R., de~Oliveira, P.H., de~Sousa~Almeida, J.L., Sedona, R., Kang,
  Y., Chakraborty, S., Wang, S., Gomes, C., Kumar, A., Truong, M., Godwin, D.,
  Lee, H., Hsu, C.Y., Asanjan, A.A., Mujeci, B., Shidham, D., Keenan, T.,
  Arevalo, P., Li, W., Alemohammad, H., Olofsson, P., Hain, C., Kennedy, R.,
  Zadrozny, B., Bell, D., Cavallaro, G., Watson, C., Maskey, M., Ramachandran,
  R., Moreno, J.B.: Prithvi-eo-2.0: A versatile multi-temporal foundation model
  for earth observation applications  (2025),
  \url{https://arxiv.org/abs/2412.02732}

\bibitem{tay2022efficient}
Tay, Y., Dehghani, M., Bahri, D., Metzler, D.: Efficient transformers: A
  survey. ACM Computing Surveys  \textbf{55}(6),  1--28 (2022).
  \doi{10.1145/3530811}

\bibitem{branchynet}
Teerapittayanon, S., McDanel, B., Kung, H.T.: {BranchyNet}: Fast inference via
  early exiting from deep neural networks. In: International Conference on
  Pattern Recognition (ICPR). pp. 2464--2469 (2016).
  \doi{10.1109/ICPR.2016.7900006}

\bibitem{Thoreau_2025_ICCV}
Thoreau, R., Marsocci, V., Derksen, D.: Parameter-efficient adaptation of
  geospatial foundation models through embedding deflection. In: Proceedings of
  the IEEE/CVF International Conference on Computer Vision (ICCV). pp.
  9594--9604 (October 2025)

\bibitem{deit}
Touvron, H., Cord, M., Douze, M., Massa, F., Sablayrolles, A., Jegou, H.:
  Training data-efficient image transformers and distillation through
  attention. In: Meila, M., Zhang, T. (eds.) Proceedings of the 38th
  International Conference on Machine Learning. Proceedings of Machine Learning
  Research, vol.~139, pp. 10347--10357. PMLR (18--24 Jul 2021),
  \url{https://proceedings.mlr.press/v139/touvron21a.html}

\bibitem{wanyan2024extending}
Wanyan, X., Seneviratne, S., Shen, S., Kirley, M.: Extending global-local view
  alignment for self-supervised learning with remote sensing imagery. In:
  Proceedings of the IEEE/CVF Conference on Computer Vision and Pattern
  Recognition Workshops (CVPRW). pp. 2443--2453 (2024)

\bibitem{tinyvit}
Wu, K., Zhang, J., Peng, H., Liu, M., Xiao, B., Fu, J., Yuan, L.: {TinyViT}:
  Fast pretraining distillation for small vision transformers. In: European
  Conference on Computer Vision (ECCV). pp. 68--85. Springer (2022).
  \doi{10.1007/978-3-031-19803-8_5}

\bibitem{eo_models}
Xiao, A., Xuan, W., Wang, J., Huang, J., Tao, D., Lu, S., Yokoya, N.:
  Foundation models for remote sensing and earth observation: A survey (2025).
  \doi{10.1109/MGRS.2025.3576766}

\bibitem{lgvit}
Xu, G., Hao, J., Shen, L., Hu, H., Luo, Y., Lin, H., Shen, J.: {LGViT}: Dynamic
  early exiting for accelerating vision transformer. In: Proceedings of the
  31st ACM International Conference on Multimedia. p. 9103–9114. MM '23,
  Association for Computing Machinery, New York, NY, USA (2023).
  \doi{10.1145/3581783.3611762}

\bibitem{nvit}
Yang, H., Yin, H., Shen, M., Molchanov, P., Li, H., Kautz, J.: Global vision
  transformer pruning with hessian-aware saliency. In: IEEE/CVF Conference on
  Computer Vision and Pattern Recognition (CVPR). pp. 18547--18557 (June 2023)

\bibitem{vitkd}
Yang, Z., Li, Z., Zeng, A., Li, Z., Yuan, C., Li, Y.: {ViTKD}: Feature-based
  knowledge distillation for vision transformers. In: 2024 IEEE/CVF Conference
  on Computer Vision and Pattern Recognition Workshops (CVPRW). pp. 1379--1388
  (2024). \doi{10.1109/CVPRW63382.2024.00145}

\bibitem{xpruner}
Yu, L., Xiang, W.: X-pruner: explainable pruning for vision transformers. In:
  Proceedings of the IEEE/CVF Conference on Computer Vision and Pattern
  Recognition (CVPR). pp. 24355--24363 (June 2023)

\bibitem{ptq4vit}
Yuan, Z., Xue, C., Chen, Y., Wu, Q., Sun, G.: {PTQ4ViT}: Post-training
  quantization for vision transformers with twin uniform quantization. In:
  European Conference on Computer Vision (ECCV). pp. 191--207. Springer (2022).
  \doi{10.1007/978-3-031-19775-8_12}

\end{thebibliography}

% ================================================================
% SUPPLEMENTARY MATERIAL (merged as appendices for arXiv)
% ================================================================
\clearpage
\appendix
\title{SIMPLER: Efficient Foundation Model Adaptation via Similarity-Guided Layer Pruning for Earth Observation\\
\large Supplementary Material}

\titlerunning{SIMPLER - Supplementary Material}

\author{V{\'\i}ctor Barreiro\inst{1,2} \and
Johannes Jakubik\inst{3} \and
Francisco Arg{\"u}ello\inst{2} \and
Dora B.~Heras\inst{1,2}}
\authorrunning{V.~Barreiro et al.}
\institute{Centro Singular de Investigación en Tecnoloxías Intelixentes (CiTIUS), Universidade de Santiago de Compostela, Spain \and
Departmento de Electrónica e Computación, Universidade de Santiago de Compostela, Spain \and
IBM Research Europe, Zurich, Switzerland}

\maketitle

% Reset section and equation numbering for supplementary
\setcounter{section}{0}
\renewcommand{\thesection}{\Alph{section}}
\setcounter{figure}{0}
\renewcommand{\thefigure}{S\arabic{figure}}
\setcounter{table}{0}
\renewcommand{\thetable}{S\arabic{table}}
\setcounter{equation}{0}
\renewcommand{\theequation}{S\arabic{equation}}
\setcounter{algorithm}{0}
\renewcommand{\thealgorithm}{S\arabic{algorithm}}

% ============================================================
% SECTION A: Complete Similarity Metric Comparison
% ============================================================
\section{Complete Similarity Metric Comparison}
\label{app:metric_complete}

For completeness, we provide the full comparison of similarity metrics (CKA, Jaccard, SVCCA) for layer selection on MADOS with Prithvi-EO-2 300M model, including all training and inference metrics. \Cref{tab:metric_complete_300m} shows the comprehensive results across all efficiency and performance dimensions.

\begin{table*}[htbp]
\centering
\caption{Complete similarity metric comparison (CKA vs Jaccard/SVCCA) for layer selection on MADOS 300M model. Results show mean $\pm$ std over 5 runs.}
\label{tab:metric_complete_300m}
\resizebox{\textwidth}{!}{%
\begin{tabular}{ @{}l|c|c|c|c|c|c|c|c|c|c@{}}
\toprule
\textbf{Similarity Metric} & \textbf{Cutoff} & \begin{tabular}[c]{ @{}r @{}}\textbf{Params}\\(M)\end{tabular} & \begin{tabular}[c]{ @{}r @{}}\textbf{Train}\\(M)\end{tabular} & \begin{tabular}[c]{ @{}r @{}}\textbf{Time}\\(min)\end{tabular} & \begin{tabular}[c]{ @{}r @{}}\textbf{Mem}\\(GB-VRAM)\end{tabular} & \begin{tabular}[c]{ @{}r @{}}\textbf{FLOPs}\\(G)\end{tabular} & \begin{tabular}[c]{ @{}r @{}}\textbf{Thr.}\\(img/s)\end{tabular} & \begin{tabular}[c]{ @{}r @{}}\textbf{Inf}\\(s)\end{tabular} & \textbf{mIoU (\%)} & \textbf{Acc (\%)} \\
\midrule
Baseline (Full) & 24 bl. & 303.90 & 303.90 & 15.90$\pm$4.80 & 11.70$\pm$0.47 & 238.50 & 33.02$\pm$1.94 & 3.04$\pm$0.18 & 66.9$\pm$2.5 & 95.3$\pm$1.2 \\
\midrule
SIMPLER (Jaccard/SVCCA) & 2 bl. & 26.78 & 26.78 & 5.98$\pm$1.88 & 1.48$\pm$0.05 & 21.01 & 125.55$\pm$23.98 & 0.83$\pm$0.17 & 50.7$\pm$3.4 & 84.6$\pm$0.8 \\
SIMPLER (CKA) & 5 bl. & 64.57 & 64.57 & 7.46$\pm$1.62 & 2.83$\pm$0.08 & 50.67 & 88.72$\pm$15.04 & 1.16$\pm$0.21 & 62.8$\pm$1.2 & 94.2$\pm$1.1 \\
\bottomrule
\end{tabular}%
}
\end{table*}

The complete results show that while Jaccard/SVCCA achieve the most aggressive compression (9\% parameters, 2.7$\times$ training speedup), they severely degrade performance to 76\% of baseline mIoU (50.7\% vs 66.9\%). In contrast, CKA achieves a better balance: 21\% parameters retained, 2.1$\times$ training speedup, while preserving 94\% of baseline mIoU (62.8\% vs 66.9\%) and maintaining competitive accuracy (94.2\% vs 95.3\%). The 3-layer difference in selected cutoff (2 blocks vs 5 blocks) yields an 18\% performance gap, demonstrating CKA's superior capability for identifying task-relevant depth through its smooth gradient properties and noise robustness.

\subsection{SIMPLER Cutoff Selection Algorithm}
\label{app:algorithm}

For implementation clarity, we provide the complete pseudocode for SIMPLER's automated layer selection procedure. This algorithm takes the pre-computed similarity matrix $Z$ and automatically identifies the optimal cutoff layer $c^*$ without hyperparameter tuning.

\begin{algorithm}[h]
\caption{SIMPLER Cutoff Selection}
\label{alg:simpler}
\begin{algorithmic}[1]
\Require $Z\in\mathbb{R}^{L\times L}$ (similarity matrix)
\Ensure $c^\star$ (optimal cutoff layer)
\State $C \gets \{2,\dots,L-2\}$
\ForAll{$c\in C$}
  \State $TL \gets Z[0{:}c-1,\, 0{:}c-1]$ \Comment{Top-left block ($c \times c$)}
  \State $BR \gets Z[c{:}L,\, c{:}L]$ \Comment{Bottom-right block ($(L{-}c) \times (L{-}c)$)}
  \State $\Delta_{\text{TL}} \gets \delta(TL)$ \Comment{Mean abs.\ consecutive-row difference}
  \State $\Delta_{\text{BR}} \gets \delta(BR)$
  \State $s(c) \gets \Delta_{\text{TL}} - \Delta_{\text{BR}}$
\EndFor
\State $c^\star \gets \arg\max_{c\in C} s(c)$
\State \Return $c^\star$
\end{algorithmic}
\end{algorithm}

The algorithm explores all candidate cutoffs $c \in \{2,\ldots,L-2\}$ and partitions the similarity matrix into two non-overlapping diagonal blocks: the top-left block $TL = Z[0{:}c{-}1,\, 0{:}c{-}1]$ (retained layers, indices $0$ to $c{-}1$, size $c \times c$) and the bottom-right block $BR = Z[c{:}L,\, c{:}L]$ (pruned layers, indices $c$ to $L$, size $(L{-}c) \times (L{-}c)$). The ranges are inclusive and the two blocks share no indices. For a square block $M \in \mathbb{R}^{k \times k}$, the mean absolute consecutive-row difference is $\delta(M) = \frac{1}{(k{-}1)\,k}\sum_{i=0}^{k-2}\sum_{j=0}^{k-1}|M_{i,j} - M_{i+1,j}|$. The algorithm selects the cutoff maximizing $\Delta_{TL} - \Delta_{BR}$.

Maximizing alone $\Delta_{TL}$ alone would always prefer keeping more layers, degenerating toward the full model. Minimizing $\Delta_{BR}$ alone would always prefer cutting more aggressively, degenerating toward a single block. Only the joint criterion balances compression against representational richness, which is why CKA with this scoring outperforms Jaccard/SVCCA which effectively collapse to aggressive $\Delta_{BR}$ minimization.

\subsection{Representational Similarity Metrics}

\subsubsection*{Centered Kernel Alignment (CKA)}

CKA~\cite{cka} measures the similarity between two representations $R, R' \in \mathbb{R}^{N \times D}$ by comparing their linear kernel representational similarity matrices (RSMs) $S, S'$, where $S_{i,j} = R_i^\top R_j$, via the Hilbert--Schmidt Independence Criterion (HSIC):
\begin{equation}
    m_{\text{CKA}}(R, R') = \frac{\text{HSIC}(S, S')}{\sqrt{\text{HSIC}(S,S)\,\text{HSIC}(S',S')}},
\end{equation}
where $\text{HSIC}(S, S') = \frac{1}{(N-1)^2}\,\text{tr}(S H_N S' H_N)$, with centering matrix $H_N = I_N - \frac{1}{N}\mathbf{1}_N\mathbf{1}_N^\top$, computed on mean-centered representations. CKA is bounded in $[0, 1]$, with a value of $1$ indicating identical relational structure, and is invariant to orthogonal transformations and isotropic scaling.

Intuitively, CKA captures the \emph{global pairwise structure} of representations: two representations are considered similar if the relative distances and similarities among all $N$ instances are preserved. This makes it sensitive to broad organizational patterns across the full representation space.

\subsubsection*{$k$-NN Jaccard Similarity}

$k$-NN Jaccard similarity~\cite{similarity_survey} assesses similarity through the overlap of local neighborhoods. For each instance $i$, let $\mathcal{N}_R^k(i)$ denote its $k$ nearest neighbors in representation $R$ under cosine similarity. The measure is:
\begin{equation}
    m_{\text{Jaccard}}^k(R, R') = \frac{1}{N} \sum_{i=1}^{N} \frac{|\mathcal{N}_R^k(i) \cap \mathcal{N}_{R'}^k(i)|}{|\mathcal{N}_R^k(i) \cup \mathcal{N}_{R'}^k(i)|}.
\end{equation}
Scores lie in $[0, 1]$, with $1$ indicating identical neighborhoods across both representations.

In contrast to CKA, Jaccard similarity captures \emph{local neighborhood structure}: two representations are similar if each instance retains roughly the same set of nearby instances. This reflects whether models cluster semantically related inputs consistently, and is particularly informative about local topological organization. In all experiments we set $k=20$, following standard practice in representational similarity analysis.

\subsubsection*{Singular Vector CCA (SVCCA)}

SVCCA~\cite{svcca} first denoises each representation via truncated SVD, retaining singular vectors that explain a fixed proportion of variance (typically $t = 0.99$), yielding denoised representations $\tilde{R}$ and $\tilde{R}'$. Canonical correlation analysis is then applied to identify the directions of maximum covariation, producing canonical correlations $\rho_1, \ldots, \rho_{D_{\min}}$. Similarity is defined as their mean:
\begin{equation}
    m_{\text{SVCCA}}(R, R') = \frac{1}{D_{\min}} \sum_{i=1}^{D_{\min}} \rho_i,
\end{equation}
where $D_{\min}$ is the number of retained components. SVCCA is bounded in $[0, 1]$ and is invariant to orthogonal transformations, isotropic scaling, and translation.

The SVD denoising step filters out low-variance dimensions assumed to carry noise rather than signal, making SVCCA more robust to representation noise than standard CCA. The subsequent canonical correlation step then identifies the maximally correlated subspaces, providing a \emph{denoised global view} of representational alignment. The variance threshold $t$ is a hyperparameter that controls the trade-off between noise robustness and information retention.

\begin{figure*}[!htbp]
    \centering
    \includegraphics[width=1\linewidth]{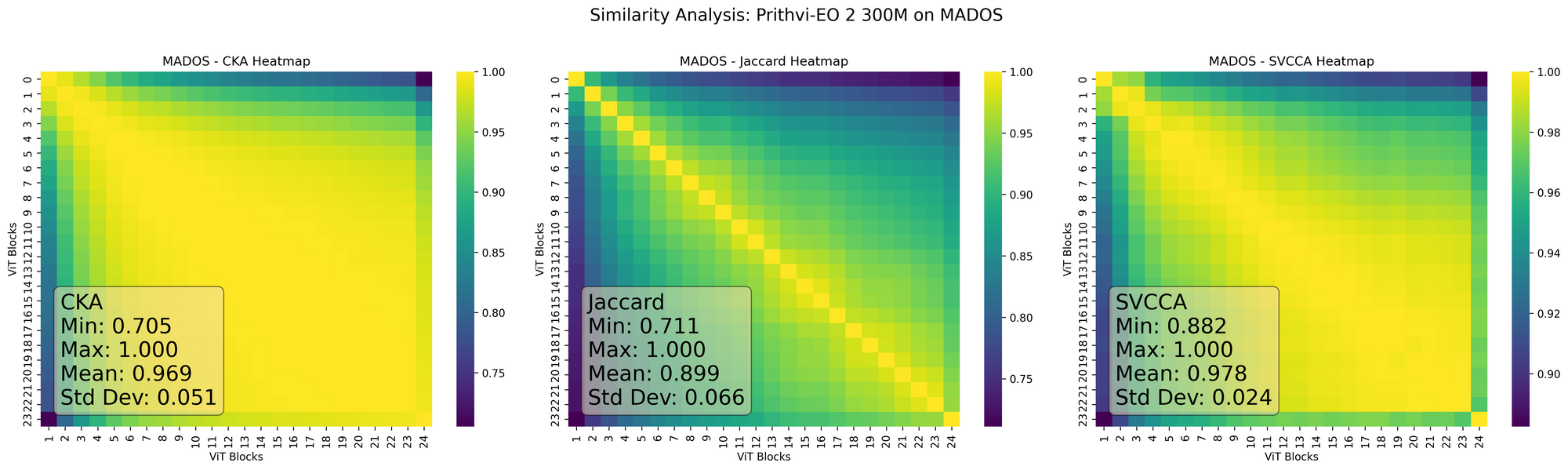}

    \includegraphics[width=1\linewidth]{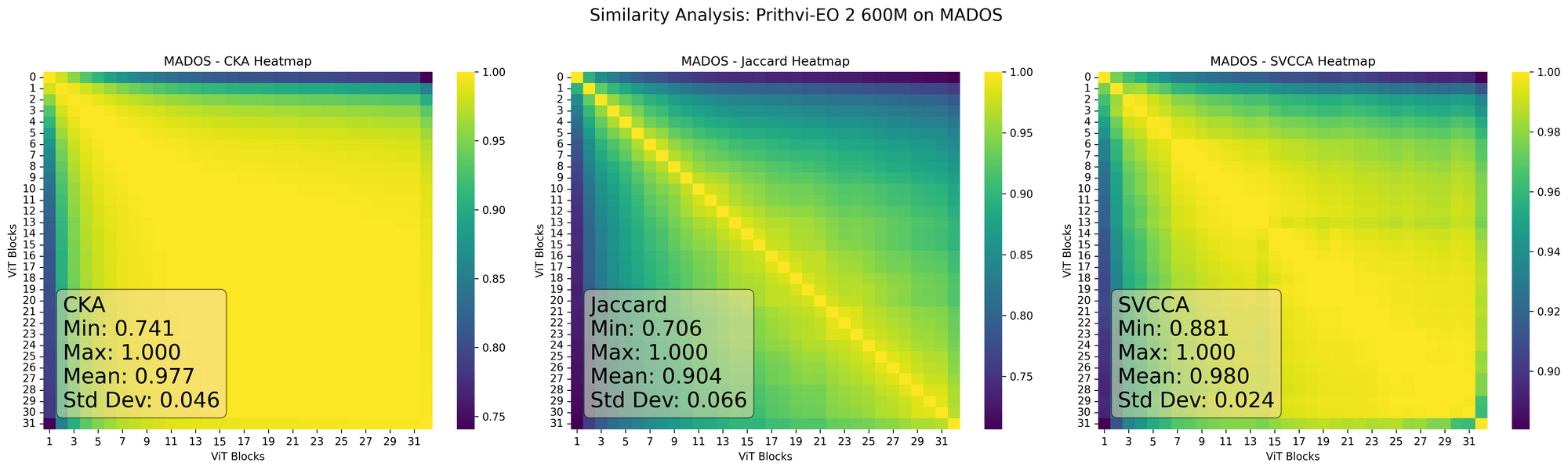}

    \includegraphics[width=1\linewidth]{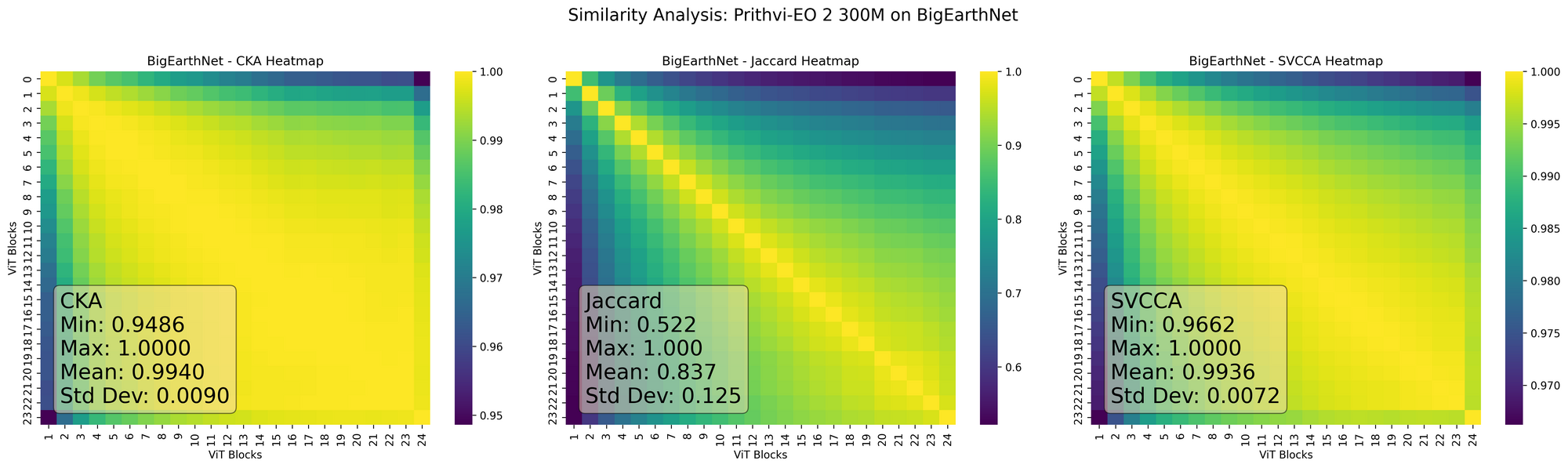}

    \includegraphics[width=1\linewidth]{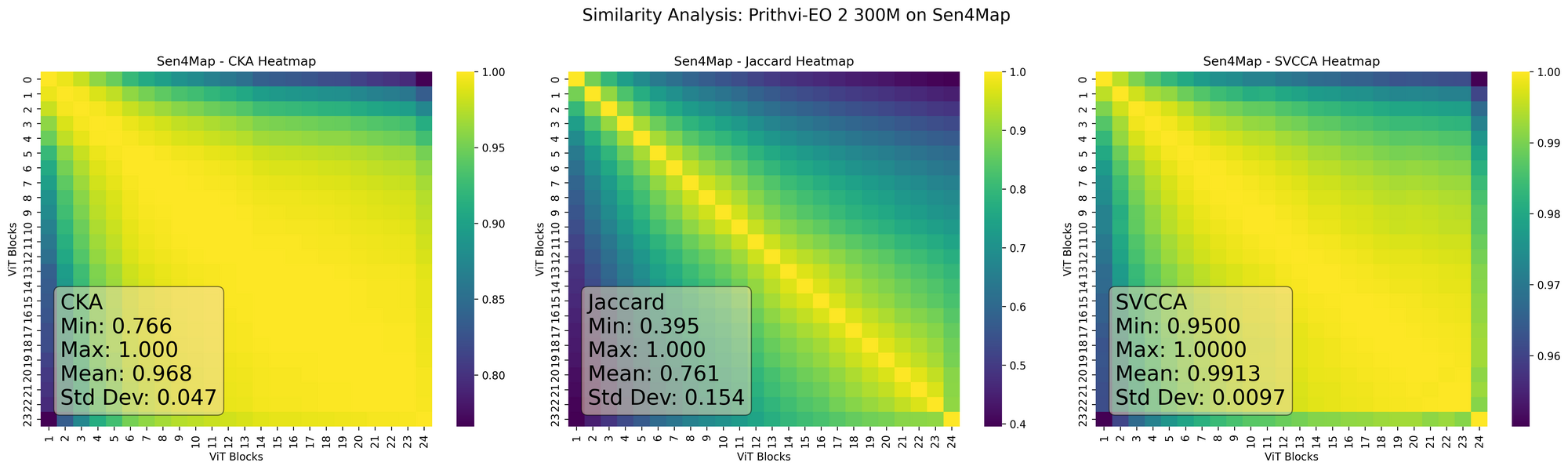}

    \caption{CKA, Jaccard and SVCCA heatmaps for the considered combinations of models and datasets in the remote sensing domain and Prithvi-EO 2. }
    \label{fig:heatmpas_detailed_remote_sensing}
\end{figure*}

\begin{figure*}[!htbp]
    \centering
    \includegraphics[width=1\linewidth]{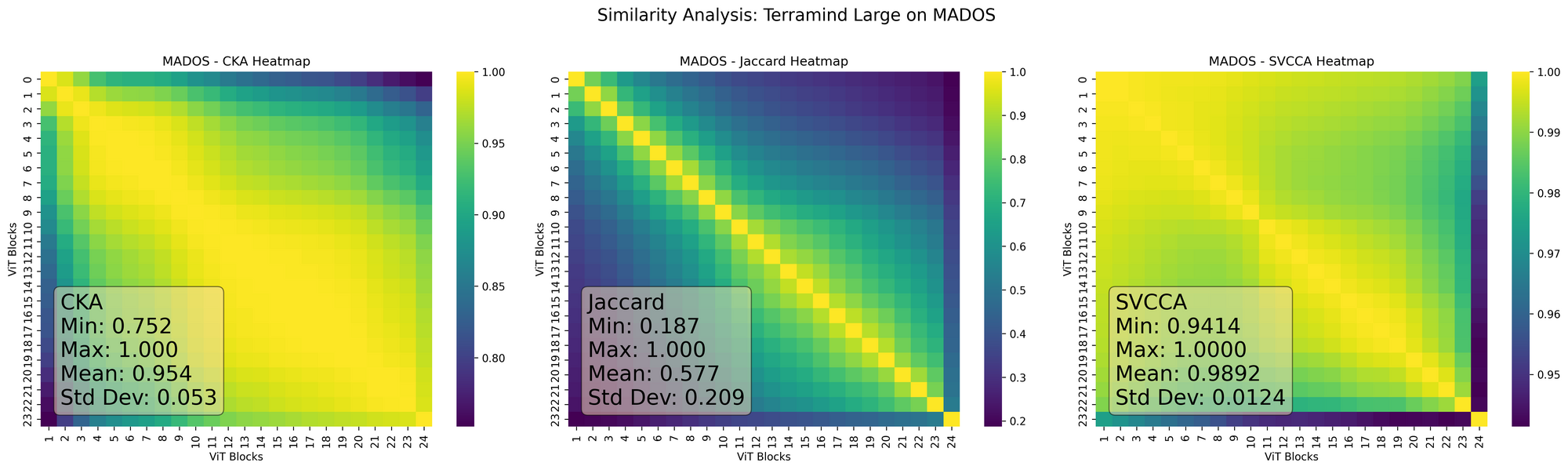}
    \includegraphics[width=1\linewidth]{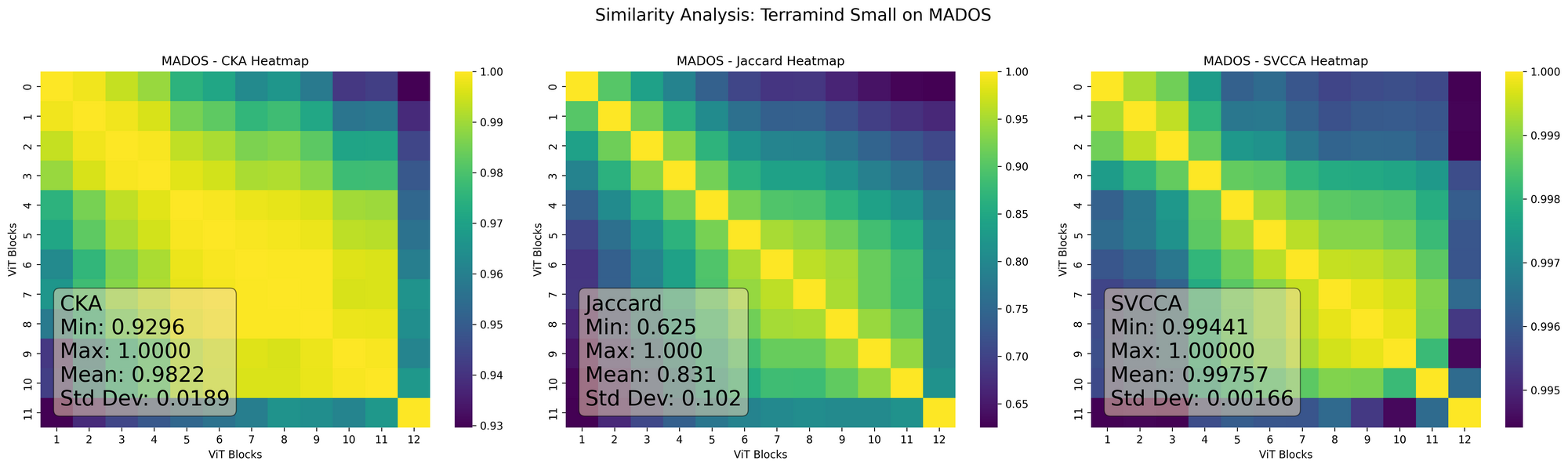}
    \includegraphics[width=1\linewidth]{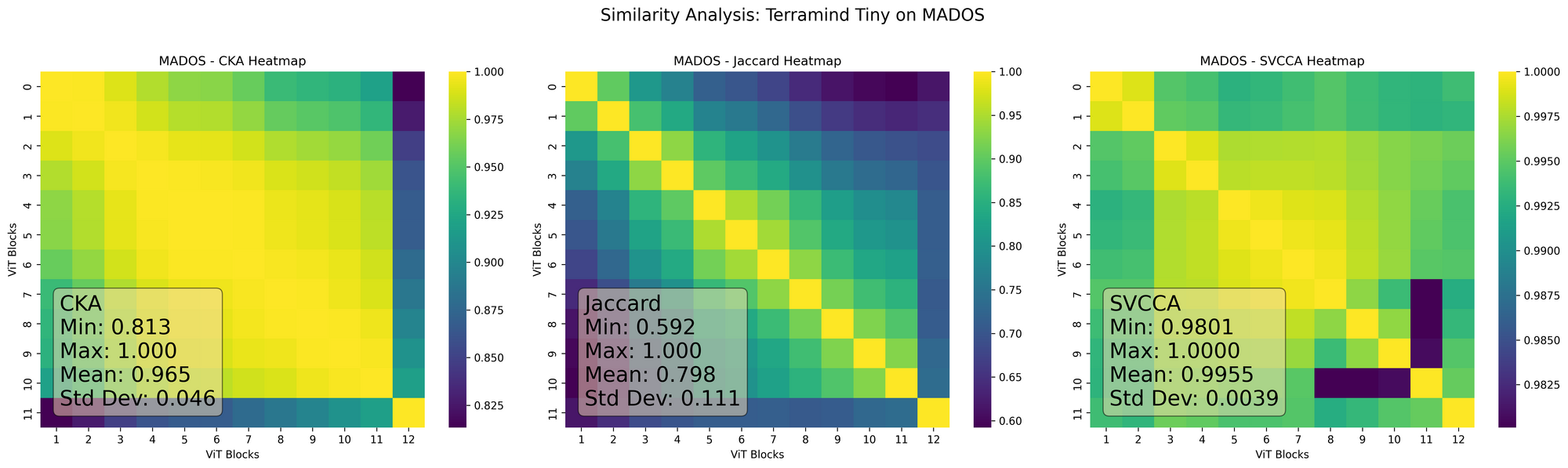}
    \caption{CKA, Jaccard and SVCCA heatmaps for the considered combinations of TerraMind family and MADOS dataset. }
    \label{fig:heatmpas_detailed_terramind_family}
\end{figure*}

\begin{figure*}[!htbp]
    \centering
    \includegraphics[width=1\linewidth]{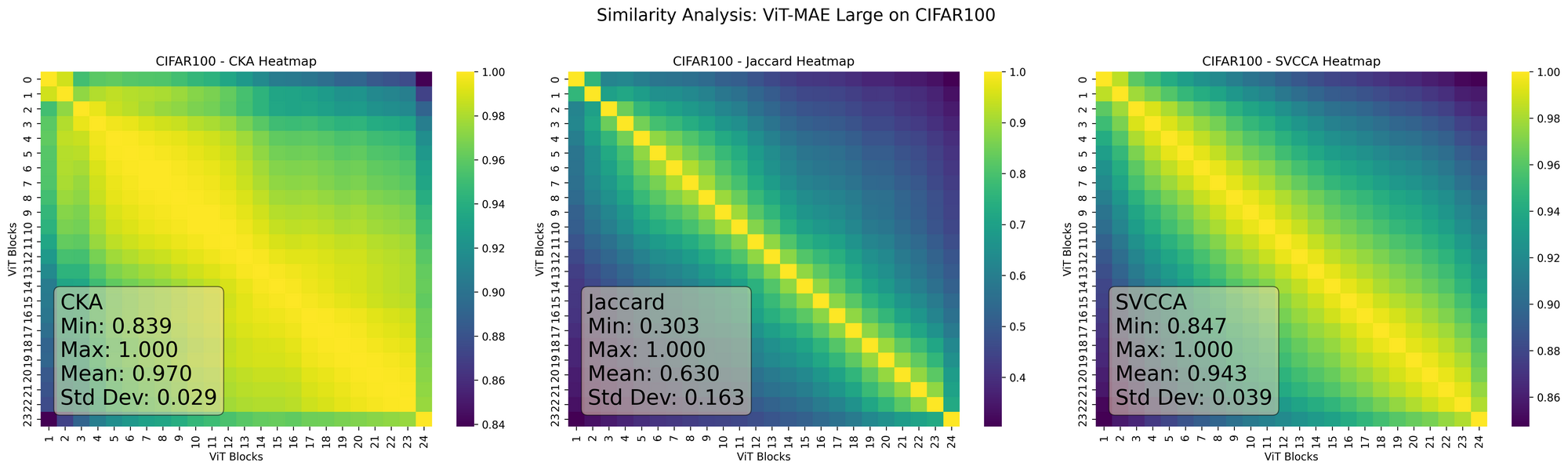}
    \caption{CKA, Jaccard and SVCCA heatmaps for ViT-MAE (trained on ImageNet) over the CIFAR-100 dataset. }
    \label{fig:heatmpas_detailed_vitmae_cifar}
\end{figure*}

\begin{figure*}[!htbp]
    \centering
    \includegraphics[width=1\linewidth]{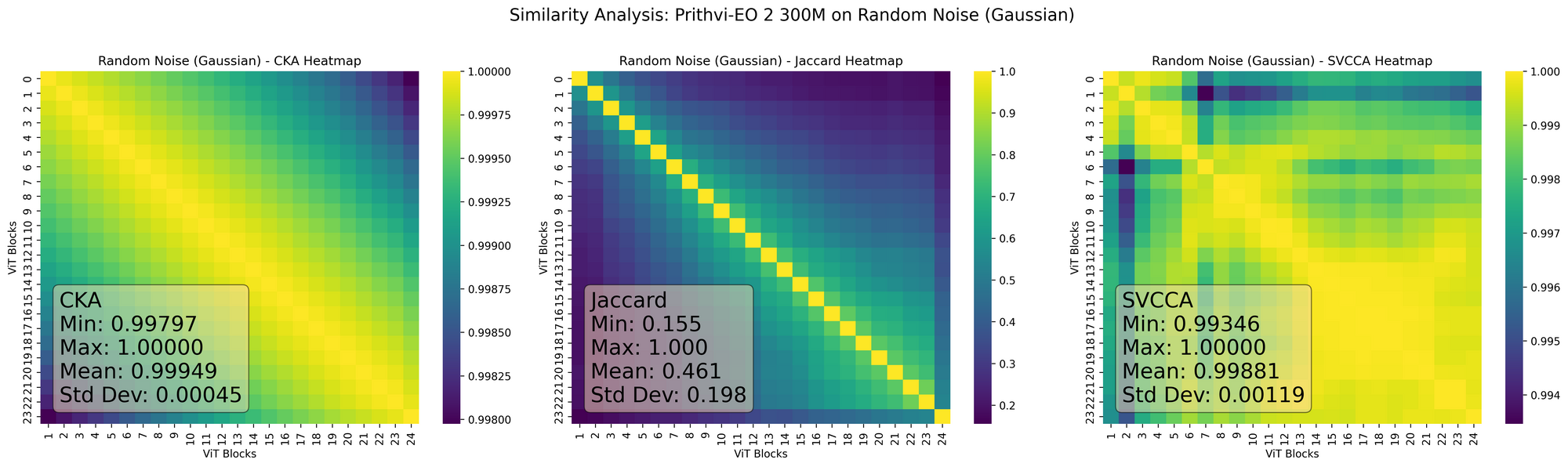}

    \includegraphics[width=1\linewidth]{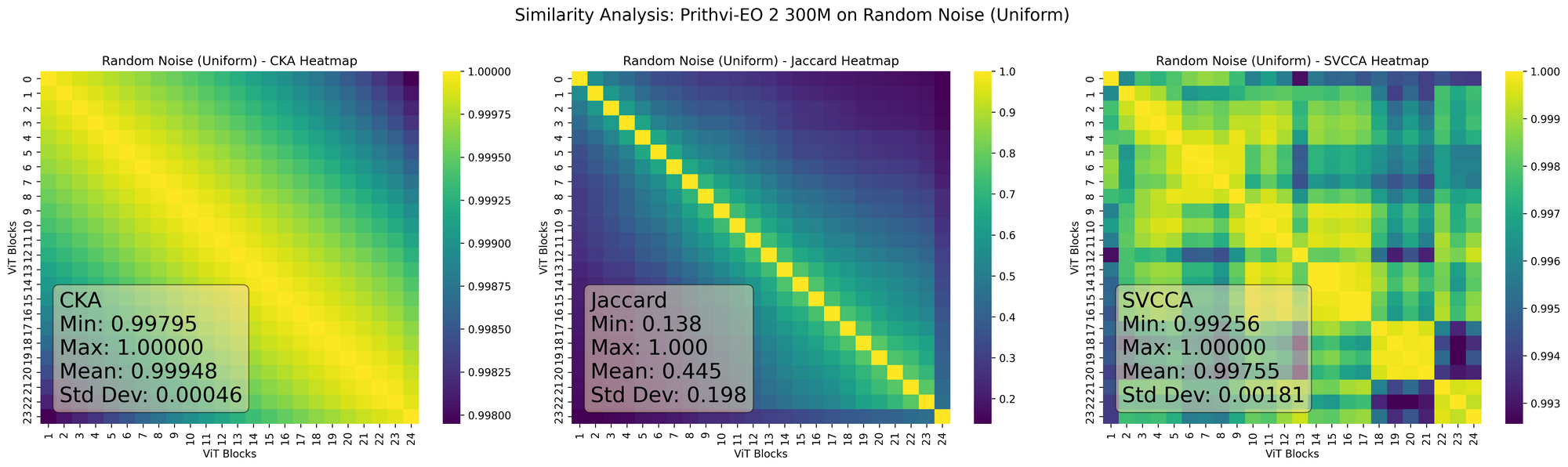}
    \caption{CKA, Jaccard and SVCCA heatmaps for the ablation analysis with noise input for the Prithvi-EO-2 300M. }
    \label{fig:heatmpas_detailed_noise}
\end{figure*}

\FloatBarrier
\clearpage

% ============================================================
% SECTION B: Training From Scratch: ViT-MAE on CIFAR-100
% ============================================================
\section{Training From Scratch: ViT-MAE on CIFAR-100}
\label{app:vit_scratch}

To provide complete context for interpreting the ViT-MAE results on CIFAR-100, we trained both the full baseline and SIMPLER's pruned architecture from scratch (without ImageNet pre-training). \Cref{tab:vit_cifar100_complete} presents the comprehensive comparison across both training regimes.

\begin{table*}[htbp]
\centering
\caption{Complete comparison of ViT-MAE on CIFAR-100 including training from scratch (without pre-training) and with ImageNet pre-training. Results show mean $\pm$ std over 5 runs.}
\label{tab:vit_cifar100_complete}
\resizebox{\textwidth}{!}{%
\begin{tabular}{@{}l|l|r|r|r|r|r|r|r|r|r@{}}
\toprule
 & & \multicolumn{4}{c}{\textbf{Training Cost}} & \multicolumn{3}{c}{\textbf{Inference Cost}} & \multicolumn{1}{c}{\textbf{Performance}} \\ \midrule
\textbf{Pre-training} & \textbf{Method} & \multicolumn{1}{r}{\begin{tabular}[c]{@{}r@{}}\textbf{Params}\\ (M)\end{tabular}} & \begin{tabular}[c]{@{}r@{}}\textbf{Train}\\ (M)\end{tabular} & \begin{tabular}[c]{@{}r@{}}\textbf{Time}\\ (min)\end{tabular} & \multicolumn{1}{r|}{\begin{tabular}[c]{@{}r@{}}\textbf{Mem}\\ (GB-VRAM)\end{tabular}} & \begin{tabular}[c]{@{}r@{}}\textbf{FLOPs}\\ (G)\end{tabular} & \begin{tabular}[c]{@{}r@{}}\textbf{Thr.}\\ (img/s)\end{tabular} & \multicolumn{1}{r|}{\begin{tabular}[c]{@{}r@{}}\textbf{Inf}\\ (s)\end{tabular}} & \textbf{Accuracy (\%)} \\ \midrule
\multirow{2}{*}{\begin{tabular}[c]{@{}l@{}}From Scratch\\ (No Pre-train)\end{tabular}} & Baseline & \multicolumn{1}{r}{303.40} & 303.40 & 247.94$\pm$57.49 & \multicolumn{1}{r|}{49.59$\pm$1.46} & 59.70 & 117.80$\pm$1.40 & \multicolumn{1}{r|}{1.09$\pm$0.18} & 63.8$\pm$2.8 \\
 & SIMPLER (Ours) & \multicolumn{1}{r}{38.88} & 38.88 & 39.51$\pm$22.17 & \multicolumn{1}{r|}{9.06$\pm$1.02} & 7.60 & 823.31$\pm$2.78 & \multicolumn{1}{r|}{0.22$\pm$0.00} & 52.1$\pm$23.8 \\
\midrule
\multirow{4}{*}{\begin{tabular}[c]{@{}l@{}}Pre-trained\\ (ImageNet)\end{tabular}} & Baseline & \multicolumn{1}{r}{303.40} & 303.40 & 49.31$\pm$5.37 & \multicolumn{1}{r|}{49.59$\pm$1.46} & 59.70 & 119.11$\pm$1.19 & \multicolumn{1}{r|}{1.02$\pm$0.15} & 88.8$\pm$0.4 \\
 & Baseline with LoRA & \multicolumn{1}{r}{309.72} & 6.42 & 116.50$\pm$12.61 & \multicolumn{1}{r|}{57.75$\pm$0.59} & 60.94 & 88.78$\pm$0.49 & \multicolumn{1}{r|}{1.49$\pm$0.26} & 91.8$\pm$0.2 \\
 & SIMPLER (Ours) & \multicolumn{1}{r}{38.88} & 38.88 & 42.97$\pm$1.37 & \multicolumn{1}{r|}{9.06$\pm$1.02} & 7.60 & 822.51$\pm$3.48 & \multicolumn{1}{r|}{0.22$\pm$0.01} & 72.8$\pm$0.3 \\
 & SIMPLER with LoRA (Ours) & \multicolumn{1}{r}{40.51} & 1.73 & 28.51$\pm$11.57 & \multicolumn{1}{r|}{9.62$\pm$0.13} & 7.92 & 583.11$\pm$26.54 & \multicolumn{1}{r|}{0.78$\pm$0.73} & 67.9$\pm$2.1 \\ \bottomrule
\end{tabular}%
}
\end{table*}

The from-scratch results establish the architectural capacity baseline. Without pre-training, the full 303.40M-parameter model achieves 63.8\% accuracy, while SIMPLER's 38.88M-parameter pruned architecture achieves 52.1\% (82\% retention). This demonstrates that the removed layers provide limited additional representational capacity for CIFAR-100. Notably, the from-scratch SIMPLER configuration exhibits high variance ($\pm$23.8pp), indicating training instability when the shallow 3-block architecture is initialized from random weights. This is consistent with the Lottery Ticket Hypothesis~\cite{lottery}: larger networks contain more favorable initialization subsets (``winning tickets'') that facilitate stable convergence, whereas the reduced architecture has fewer such subsets, leading to seed-dependent convergence. Crucially, this instability vanishes with pre-training ($\pm$0.3pp), confirming that pre-trained weights provide reliable initialization that compensates for the limited depth.

Pre-training provides benefits for both architectures: the full model improves from 63.8\% to 88.8\% (1.39$\times$ gain), while SIMPLER improves from 52.1\% to 72.8\% (1.40$\times$ gain). Critically, SIMPLER retains 82\% of the pre-trained baseline's performance despite having only 13\% of the parameters and 18\% of the memory footprint (9.06 GB-VRAM vs 49.59 GB-VRAM). This matches the 82\% architectural capacity retention observed from scratch, validating that SIMPLER successfully identifies and preserves the layers most enriched by ImageNet pre-training rather than merely selecting based on architectural capacity alone. The 81.7\% memory reduction enables deployment on consumer-grade GPUs (e.g., RTX 3090, 24 GB-VRAM), whereas the full baseline requires high-end datacenter hardware.

% ============================================================
% SECTION C: Scope -- Representation Patterns in Non-MAE Foundation Models
% (Referenced from main paper Discussion as "Suppl. Mat. Sec. C")
% ============================================================
\section{Scope: Representation Patterns in Non-MAE Foundation Models}
\label{app:collapse}

\crh{SIMPLER exploits the progressive stabilization of representations in deep layers, a pattern validated across the masked-autoencoding (MAE) style foundation models studied in the main paper (Prithvi-EO-2, TerraMind, ViT-MAE; Fig.~2 of main paper). This section characterizes the scope boundary of the method by examining two families of \emph{non-MAE} models, contrastive and collapse-prevention, where this stabilization does not emerge.}

\subsection{Collapse-Prevention Models (DINOv3-SAT)}

Recent foundation models such as DINOv2/DINOv3~\cite{dinov3} employ explicit regularization (e.g., KoLeo) to maintain feature diversity across all layers.

\begin{figure}[!htbp]
    \centering
    \includegraphics[width=0.7\linewidth]{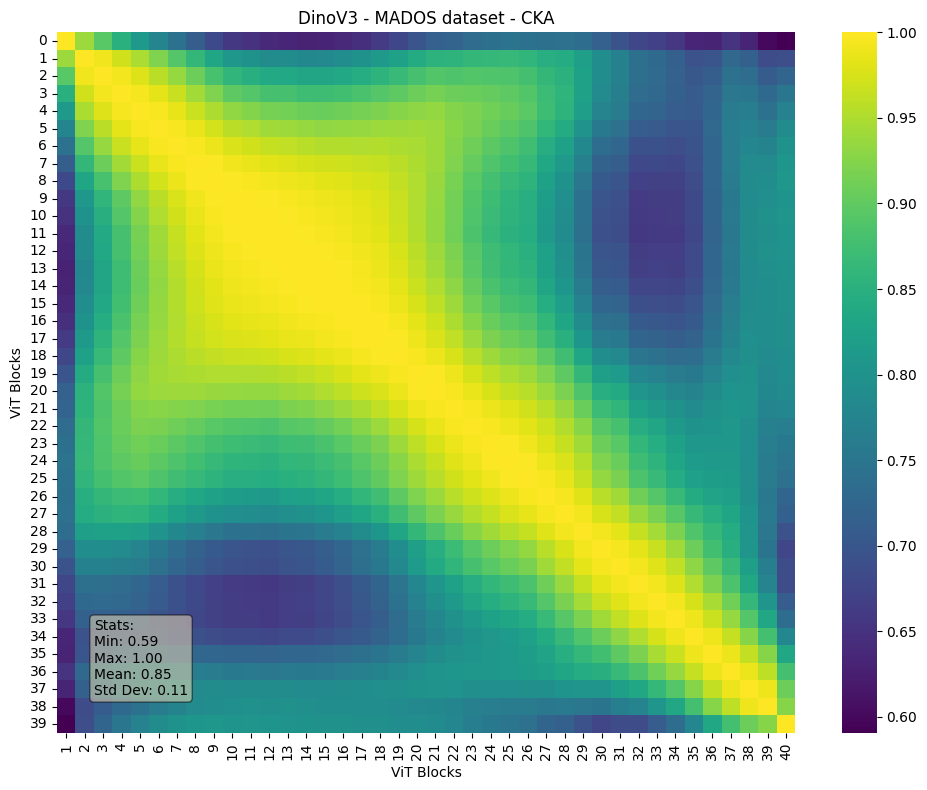}
    \caption{CKA similarity matrix for DINOv3-SAT on MADOS samples. }
    \label{fig:dinov3_cka}
\end{figure}

\Cref{fig:dinov3_cka} shows that DINOv3-SAT exhibits oscillating similarity patterns across depth, with no clear stabilization zone (std=0.11 vs. Prithvi's block-diagonal structure). This behavior is expected given DINOv3's explicit collapse prevention mechanisms.

\subsection{Contrastive Models (SoftCon)}

\begin{figure}[!htbp]
    \centering
    \includegraphics[width=0.62\linewidth]{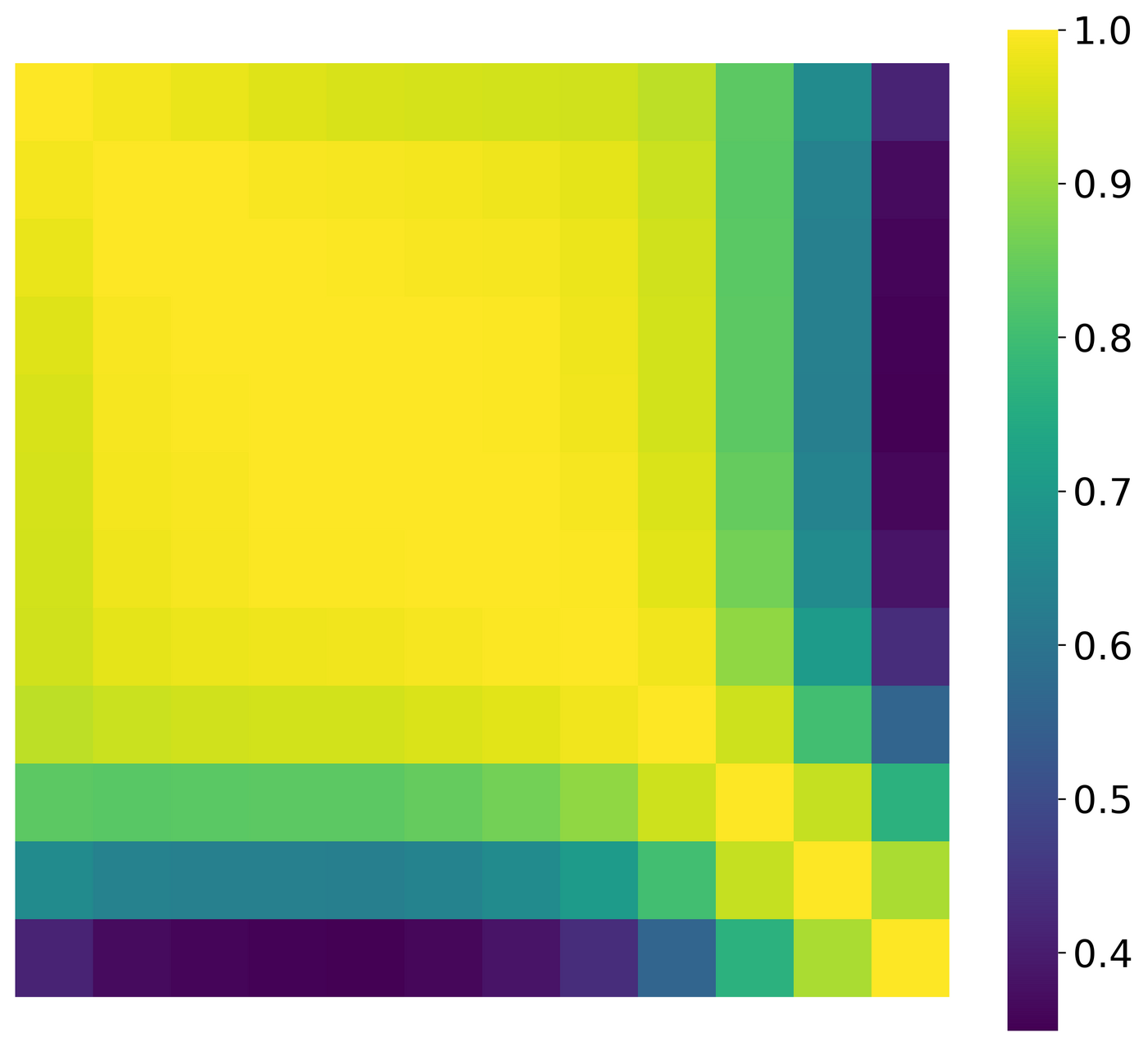}
    \caption{CKA similarity matrix for SoftCon (ViT-B/14, contrastive) on MADOS samples. Unlike the MAE-style models in Fig.~2 of the main paper, no deep-layer stabilization zone emerges.}
    \label{fig:softcon_cka}
\end{figure}

\crh{We additionally examined SoftCon~\cite{wanyan2024extending} (ViT-B/14), a contrastively pre-trained EO foundation model, on MADOS using the same 500-sample CKA protocol. As shown in \cref{fig:softcon_cka}, SoftCon also lacks deep-layer stabilization: the last layers are \emph{less} similar to one another than the early ones, mirroring the DINOv3-SAT behavior rather than the block-diagonal redundancy of MAE-style models. Consequently, SIMPLER's scoring criterion finds no redundant tail to exploit in either contrastive model.}

\crh{\textbf{Scope boundary and practitioner guidance.} Together, the contrastive (SoftCon) and collapse-prevention (DINOv3-SAT) results indicate that the deep-layer stabilization SIMPLER relies on is characteristic of MAE-style pre-training and is \emph{not} a universal property of self-supervised models. This defines a principled scope boundary rather than a failure mode: before applying SIMPLER to a new foundation model, practitioners should compute the CKA heatmap on task samples. Clear block-diagonal structure in deep layers (as in Fig.~2 of main paper) indicates suitability for similarity-based layer selection, whereas oscillating or uniform patterns indicate the method does not apply. Extending similarity-guided layer selection to non-MAE pre-training paradigms is left as future work.}

% ============================================================
% SECTION D: Qualitative Segmentation Results on MADOS
% ============================================================

\section{Qualitative Segmentation Results on MADOS}
\label{app:mados_qualitative}
We present qualitative segmentation outputs on the MADOS dataset comparing the full baseline model (Prithvi-EO-2 300M, 24 blocks), SIMPLER (5 blocks, CKA-guided), and magnitude-based post-hoc pruning (40\% compression). Within each panel, columns show the RGB input (left), ground-truth annotation (center), and predicted segmentation mask (right). The class colour legend is provided below.

% ====================
% SAMPLES 1 -- 5
% ====================
\begin{figure}[H]
    \centering
    \begin{subfigure}[t]{\linewidth}
        \centering
    \includegraphics[width=0.9\linewidth]{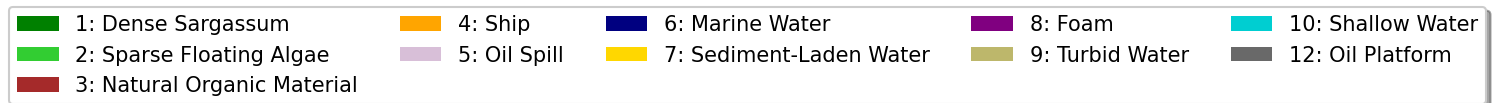}
    \caption*{\textit{Class colour legend}}

    \end{subfigure}\\[4pt]
    \begin{subfigure}[t]{0.49\linewidth}
    \includegraphics[width=\linewidth]{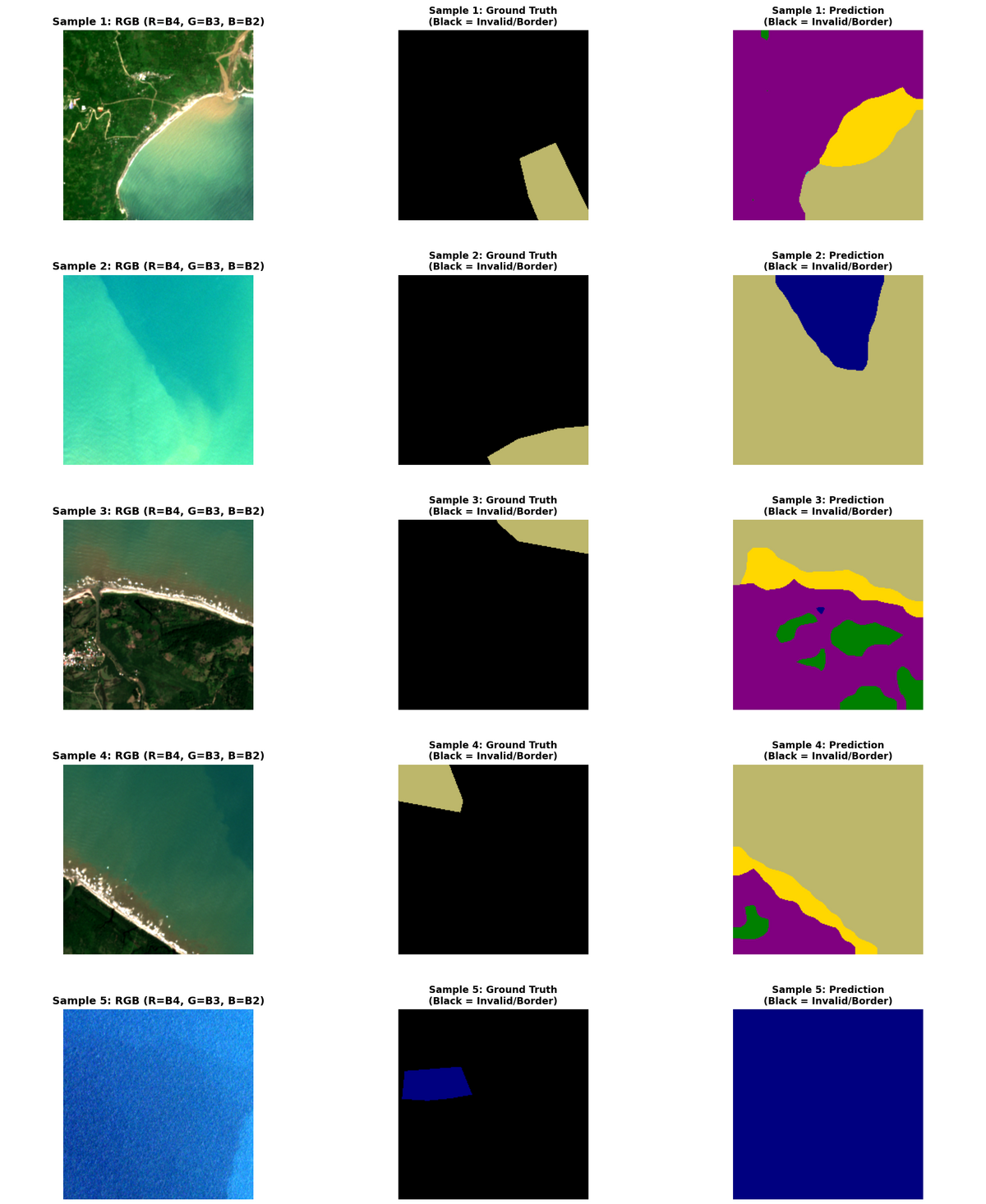}
    \caption{Full baseline (24 blocks)}

    \end{subfigure}\hfill
    \begin{subfigure}[t]{0.49\linewidth}

\includegraphics[width=\linewidth]{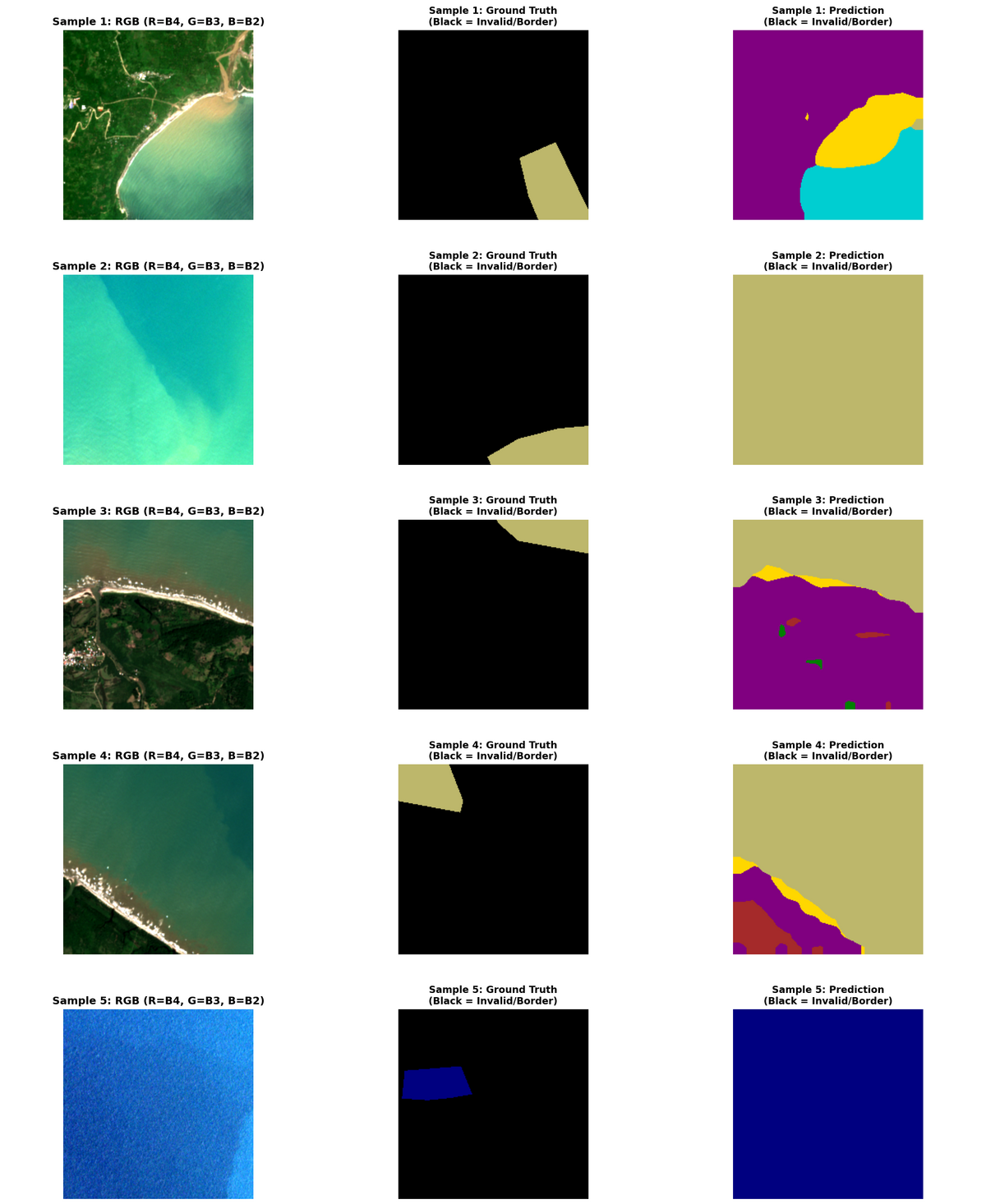}
    \caption{Post-hoc pruning (40\%)}
        
    \end{subfigure}\\[6pt]
    \begin{subfigure}[t]{0.49\linewidth}

\includegraphics[width=\linewidth]{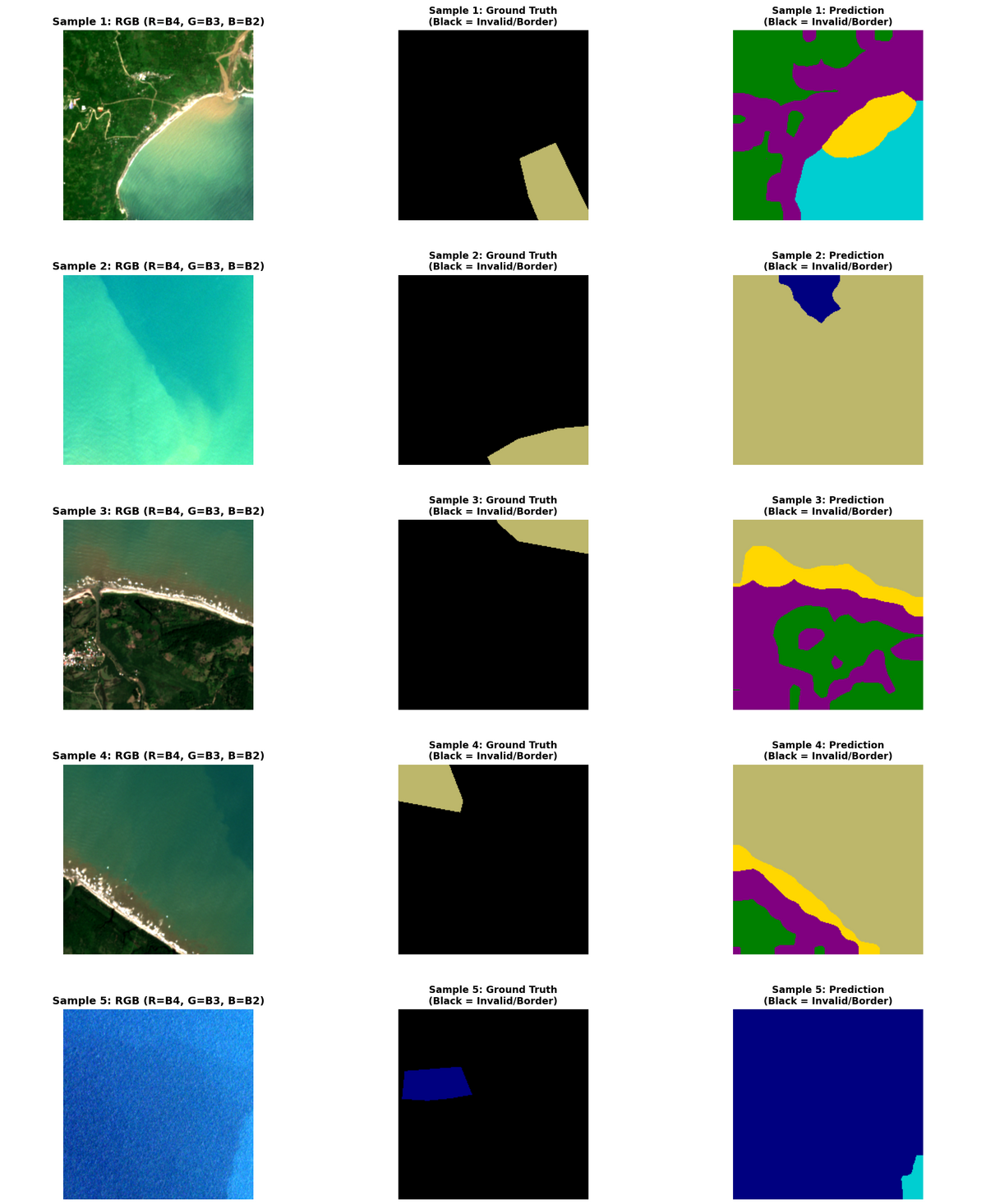}
    \caption{SIMPLER (5 blocks)}
    
    \end{subfigure}
    \caption{One sample per row: RGB input (left), ground-truth (center), prediction (right) for each model.}
    \label{fig:qual_strip1}
\end{figure}

\clearpage

% ====================
% SAMPLES 6 -- 10
% ====================
\begin{figure}[H]
    \centering
    \begin{subfigure}[t]{\linewidth}
        \centering
        \includegraphics[width=0.95\linewidth]{qual_crops/legend.png}
        \caption*{\textit{Class colour legend}}
    \end{subfigure}\\[4pt]
    \begin{subfigure}[t]{0.49\linewidth}
        \includegraphics[width=\linewidth]{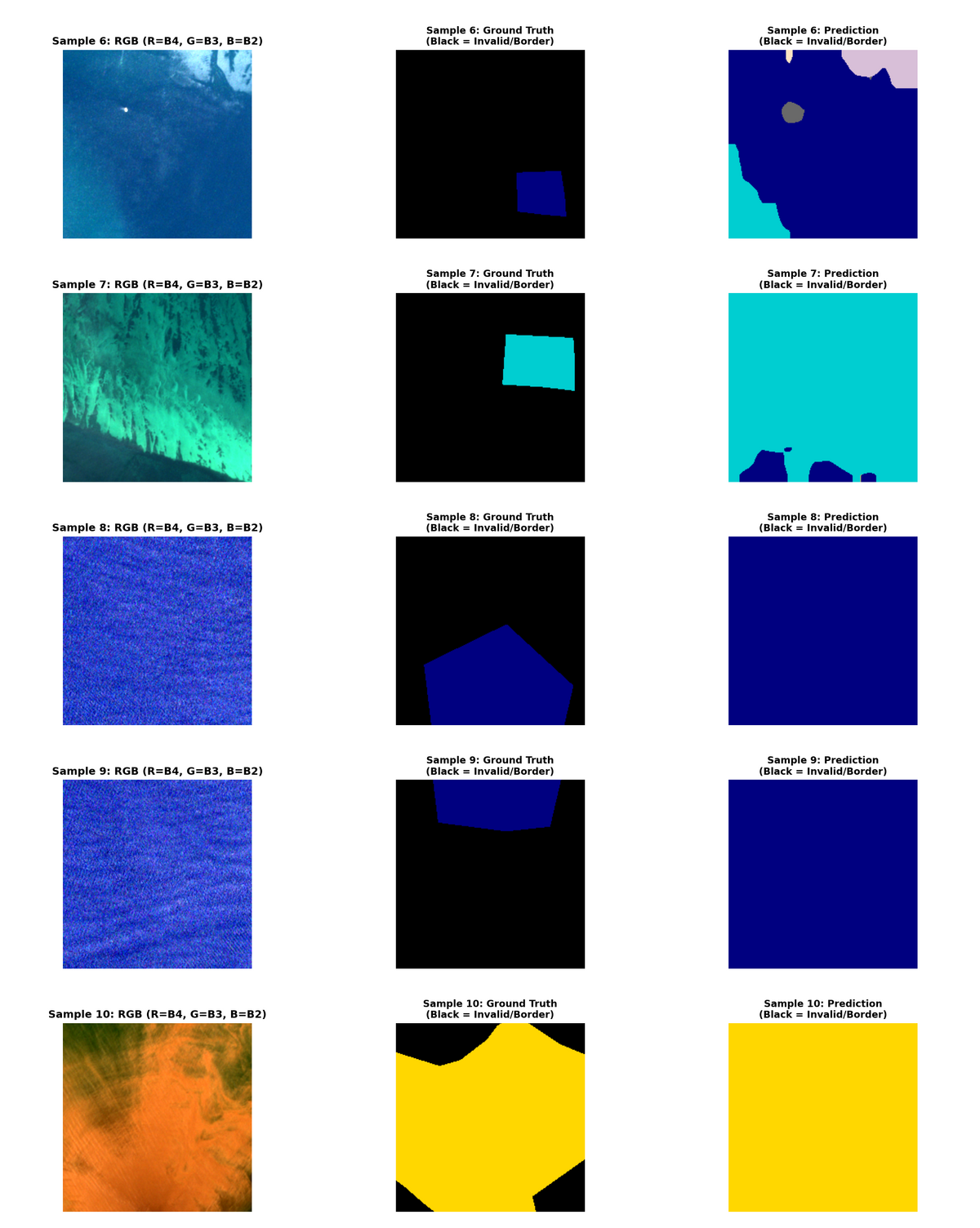}
        \caption{Full baseline (24 blocks)}
    \end{subfigure}\hfill
    \begin{subfigure}[t]{0.49\linewidth}
        \includegraphics[width=\linewidth]{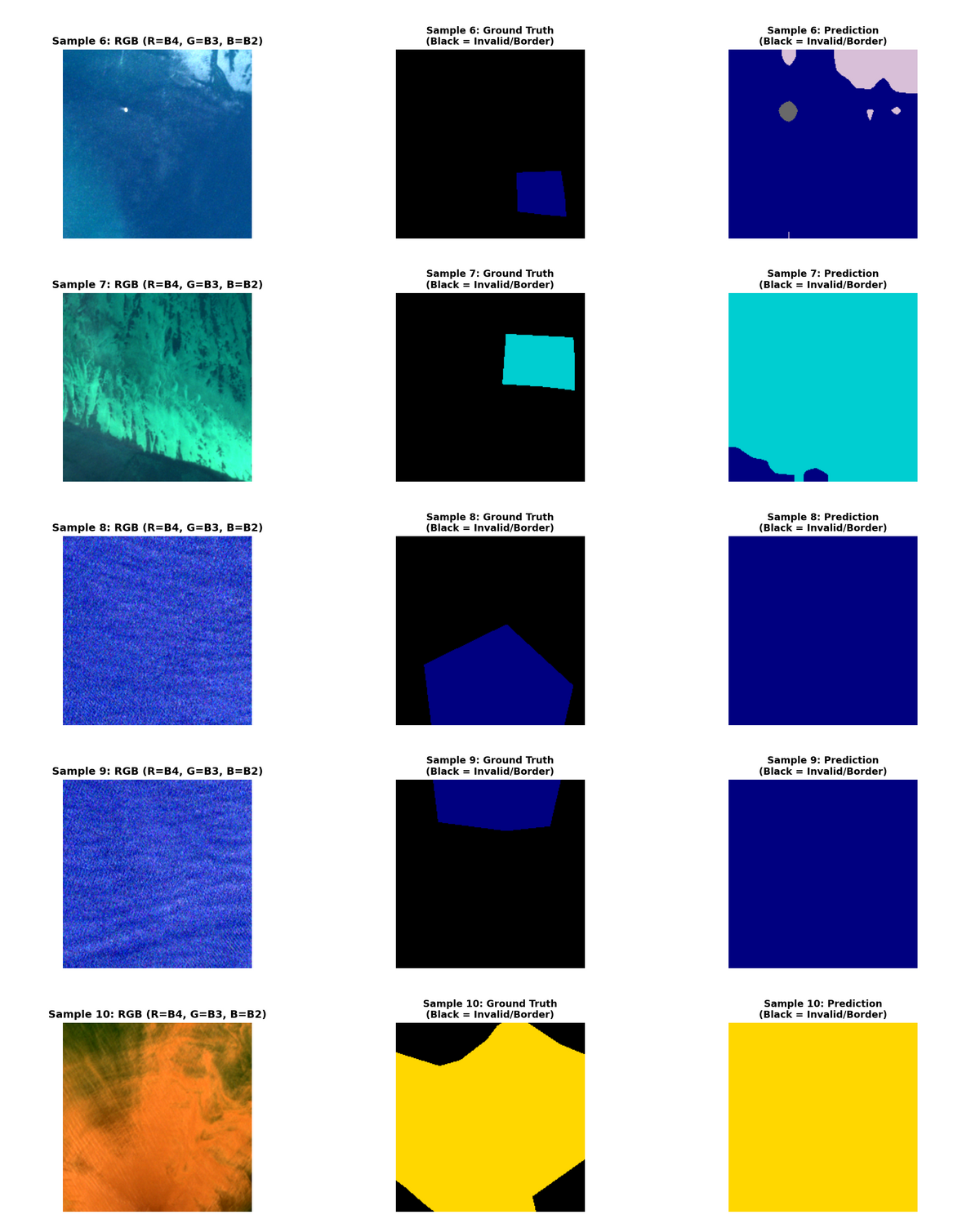}
        \caption{Post-hoc pruning (40\%)}

    \end{subfigure}\\[6pt]
    \begin{subfigure}[t]{0.49\linewidth}
        \includegraphics[width=\linewidth]{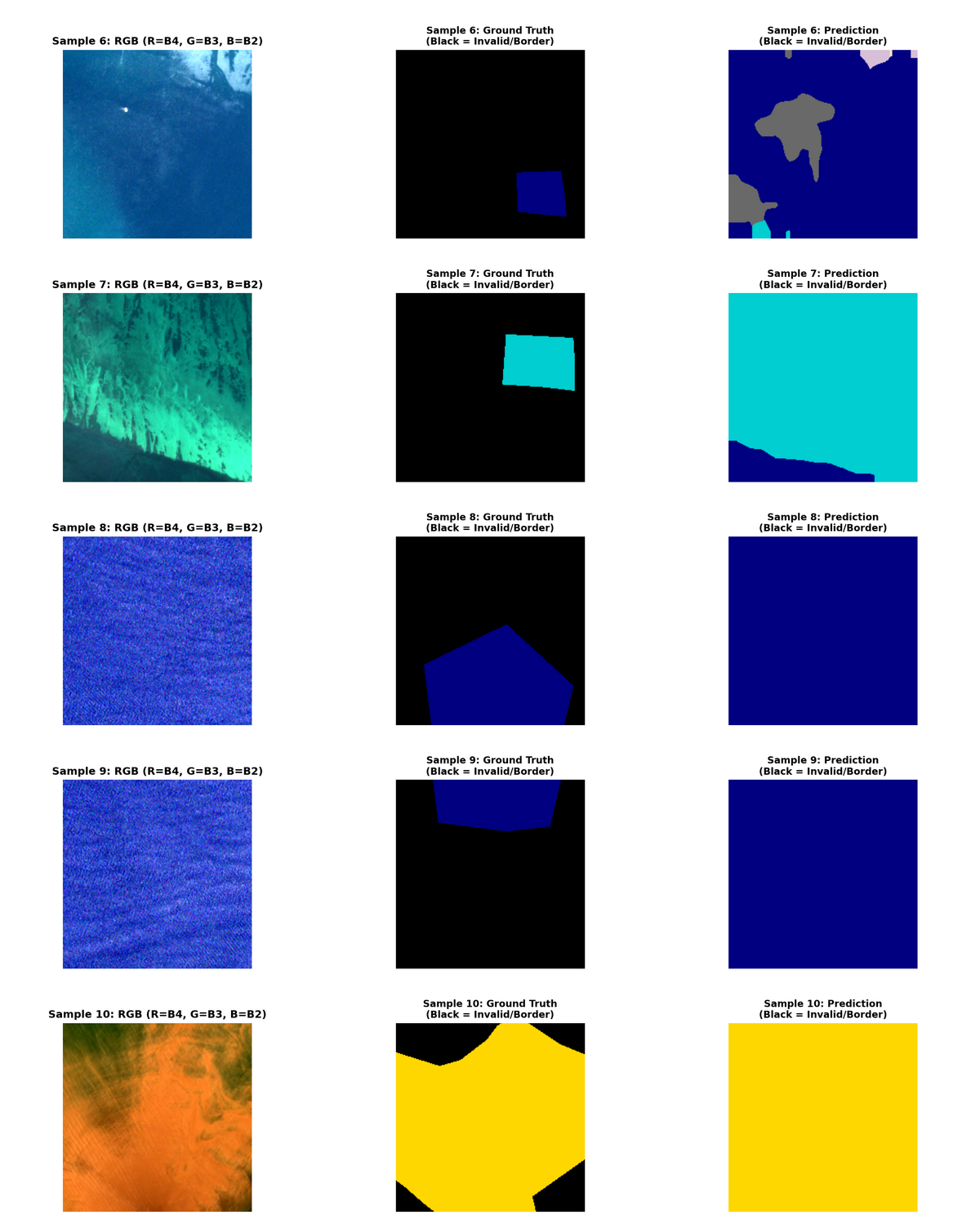}
        \caption{SIMPLER (5 blocks)}
    \end{subfigure}
    \caption{One sample per row: RGB input (left), ground-truth (center), prediction (right) for each model.}
    \label{fig:qual_strip2}
\end{figure}

\clearpage

% ====================
% SAMPLES 11 -- 15
% ====================
\begin{figure}[H]
    \centering
    \begin{subfigure}[t]{\linewidth}
        \centering
        \includegraphics[width=0.95\linewidth]{qual_crops/legend.png}
        \caption*{\textit{Class colour legend}}
    \end{subfigure}\\[4pt]
    \begin{subfigure}[t]{0.49\linewidth}
        \includegraphics[width=\linewidth]{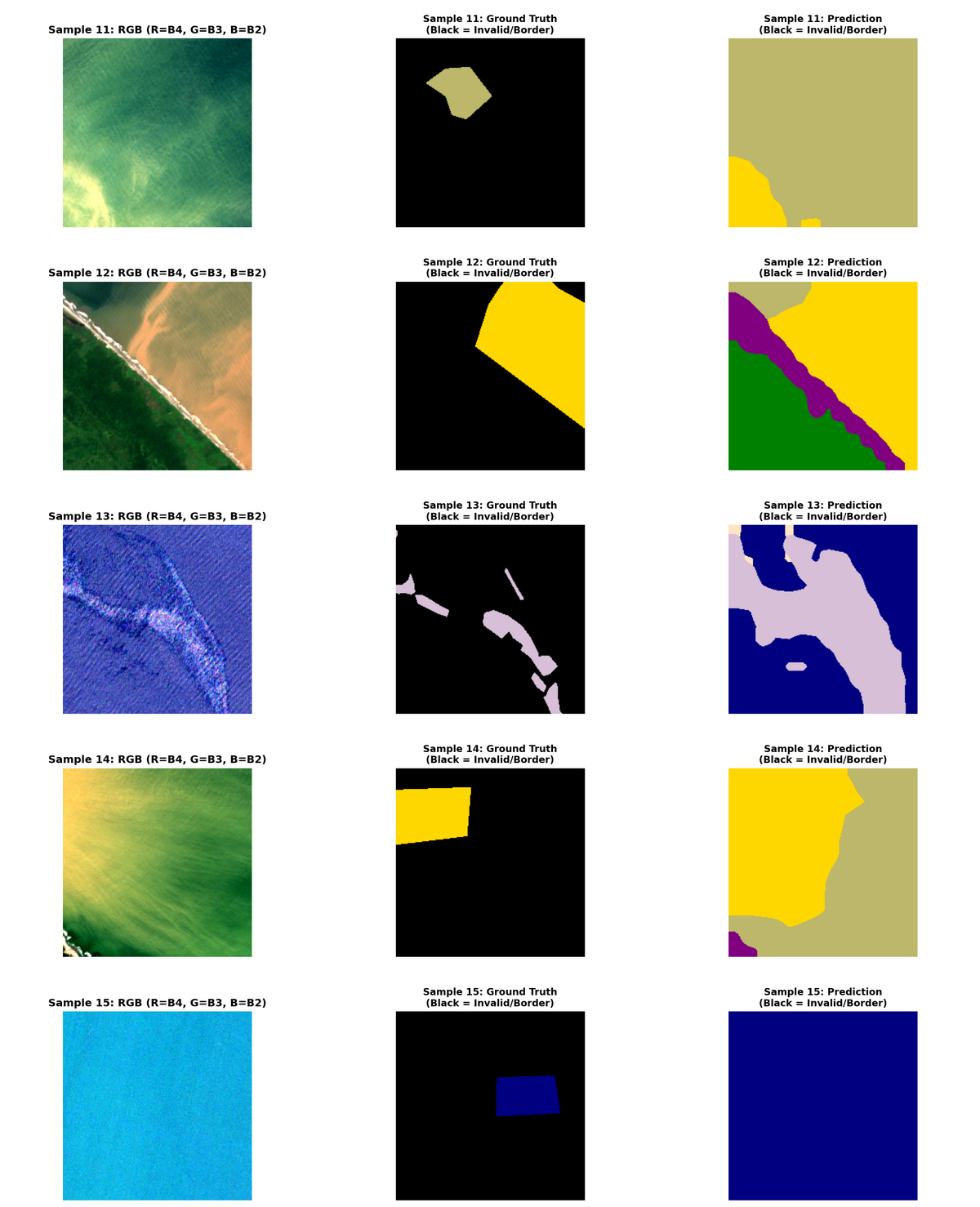}
        \caption{Full baseline (24 blocks)}
    \end{subfigure}\hfill
    \begin{subfigure}[t]{0.49\linewidth}
        \includegraphics[width=\linewidth]{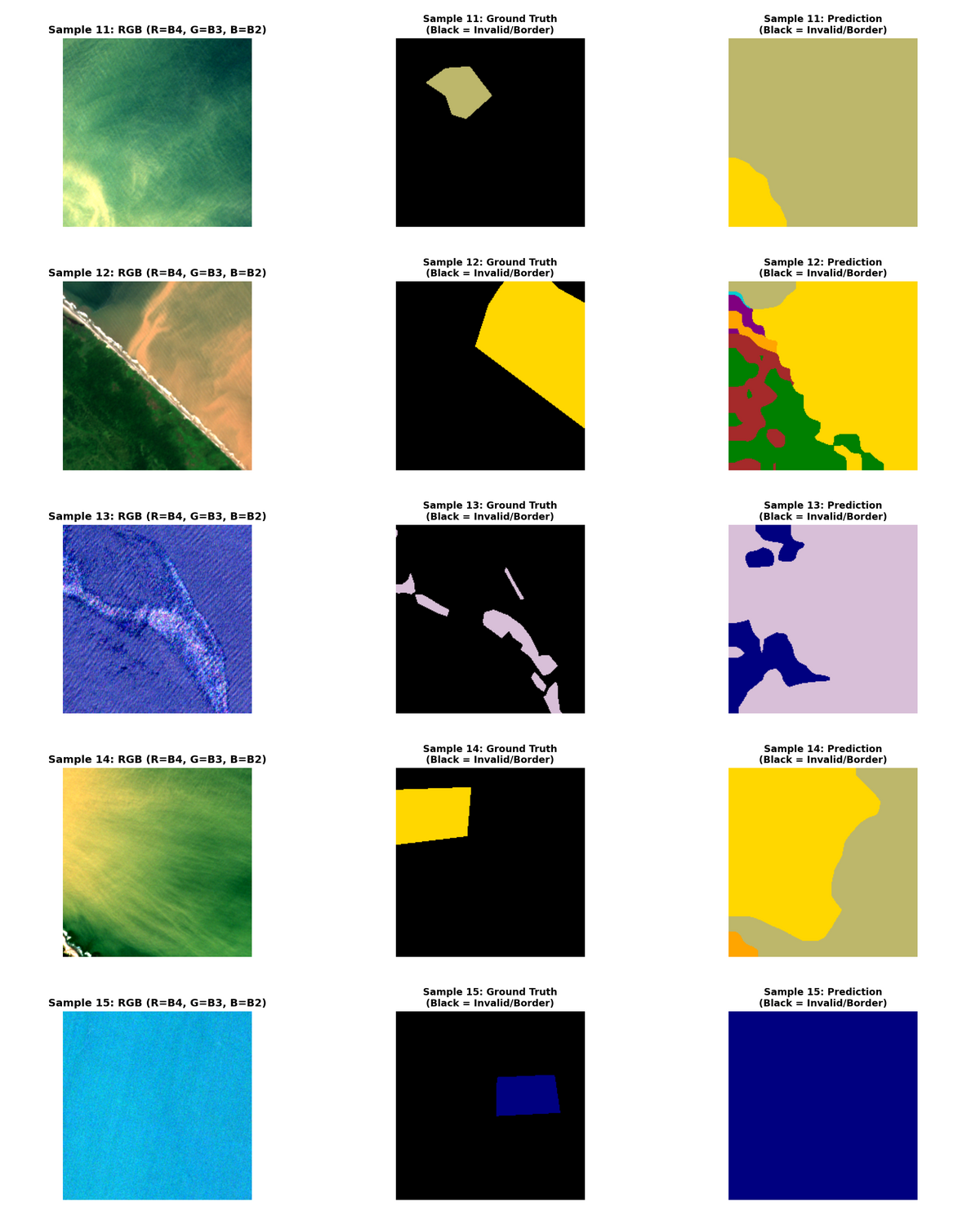}
        \caption{Post-hoc pruning (40\%)}

    \end{subfigure}\\[6pt]
    \begin{subfigure}[t]{0.49\linewidth}
        \includegraphics[width=\linewidth]{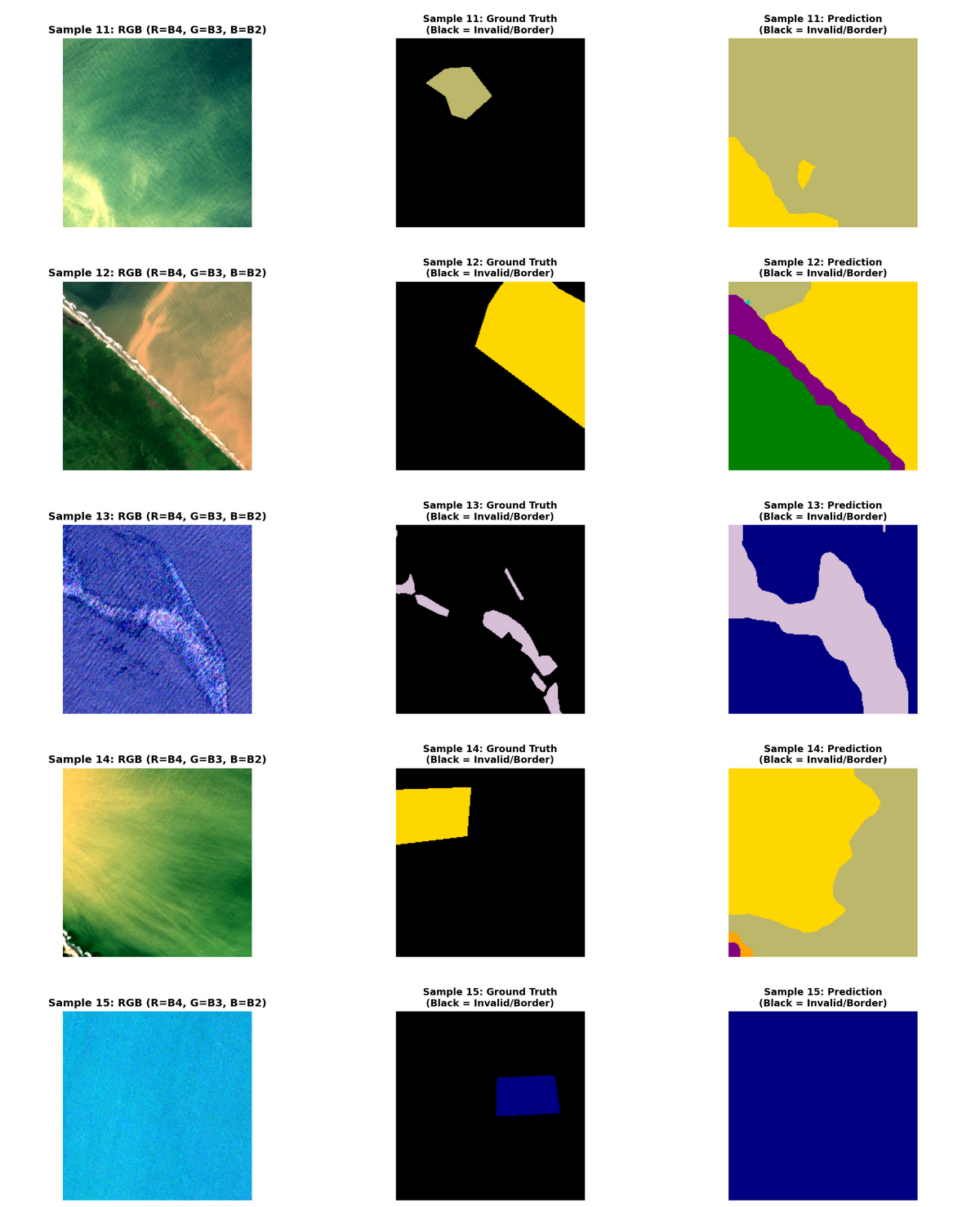}
        \caption{SIMPLER (5 blocks)}
    \end{subfigure}
    \caption{One sample per row: RGB input (left), ground-truth (center), prediction (right) for each model.}
    \label{fig:qual_strip3}
\end{figure}

\clearpage

% ====================
% SAMPLES 16 -- 20
% ====================
\begin{figure}[H]
    \centering
    \begin{subfigure}[t]{\linewidth}
        \centering
        \includegraphics[width=0.95\linewidth]{qual_crops/legend.png}
        \caption*{\textit{Class colour legend}}
    \end{subfigure}\\[4pt]
    \begin{subfigure}[t]{0.49\linewidth}
        \includegraphics[width=\linewidth]{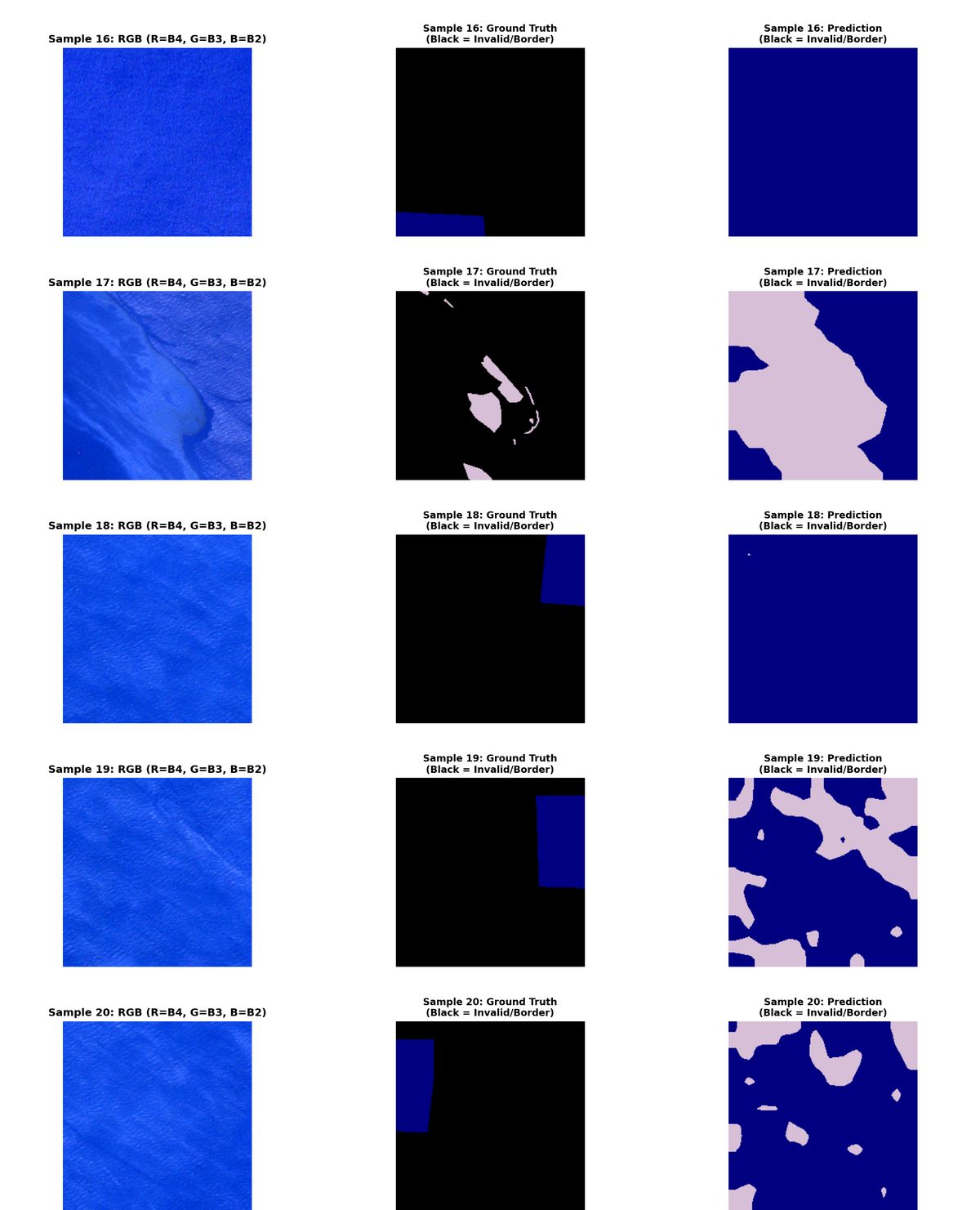}
        \caption{Full baseline (24 blocks)}
    \end{subfigure}\hfill
    \begin{subfigure}[t]{0.49\linewidth}
        \includegraphics[width=\linewidth]{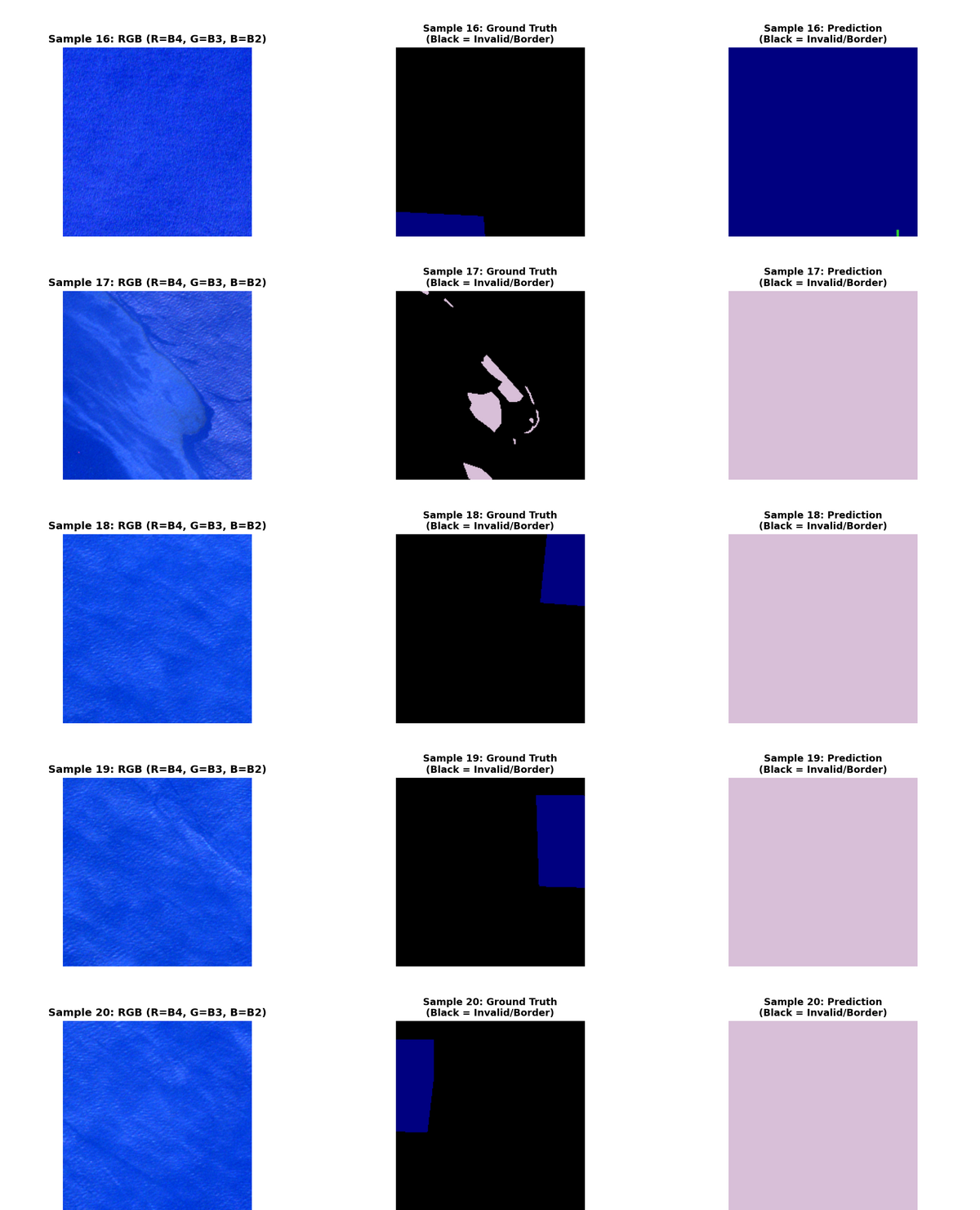}
        \caption{Post-hoc pruning (40\%)}

    \end{subfigure}\\[6pt]
    \begin{subfigure}[t]{0.49\linewidth}
        \includegraphics[width=\linewidth]{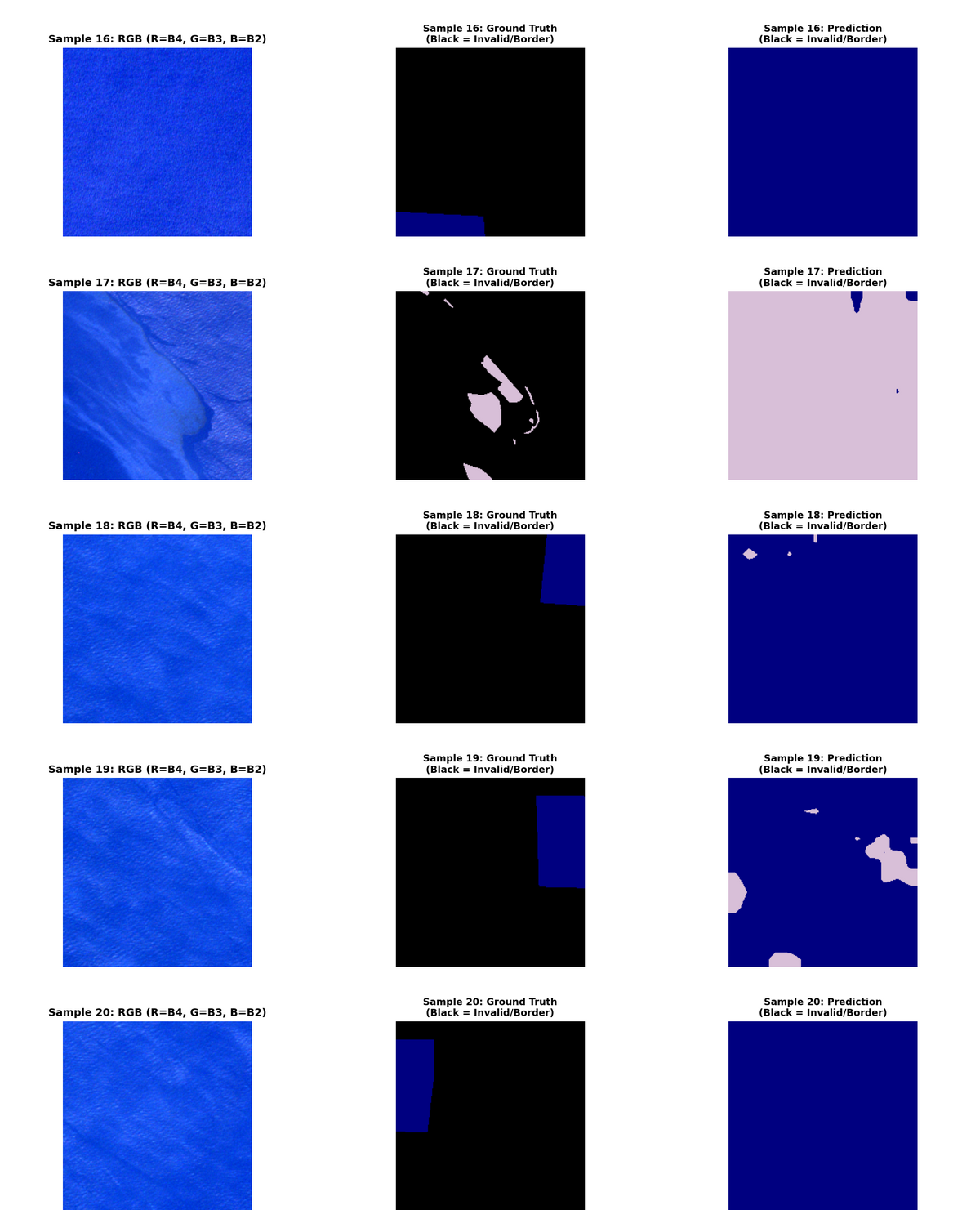}
        \caption{SIMPLER (5 blocks)}
    \end{subfigure}
    \caption{One sample per row: RGB input (left), ground-truth (center), prediction (right) for each model.}
    \label{fig:qual_strip4}
\end{figure}

\textbf{Discussion.} The qualitative results reveal consistent differences between compression strategies. In \cref{fig:qual_strip2}, the baseline produces well-defined segmentation masks that closely match the ground-truth spatial patterns; SIMPLER preserves this behavior, yielding visually comparable predictions, whereas post-hoc pruning introduces noticeably degraded boundaries and misclassified regions. A more pronounced failure mode of post-hoc pruning is visible in \cref{fig:qual_strip4}: the pruned model saturates the Oil Spill class across the last four patches (Samples 17--20), assigning it uniformly to regions that the ground truth labels as clean water. In contrast, SIMPLER correctly identifies most of these areas as water, producing predictions that are clearly closer to the baseline despite occasional minor deviations. These observations are consistent with the quantitative gap reported in Table~2 of the main paper (mIoU 62.8\% for SIMPLER vs.\ 47.9\% for 40\% pruning) and illustrate that magnitude-based pruning can catastrophically discard features critical for fine-grained class discrimination, while CKA-guided layer selection better preserves the semantic representations needed for accurate segmentation.

% ============================================================
% SECTION E: Layer-wise Linear Probing Analysis
% ============================================================
\section{Layer-wise Linear Probing Analysis}
\label{sec:appendix-linear-probing}

We validate SIMPLER's CKA-based layer selection through comprehensive linear probing experiments on Prithvi-300M ~\cite{alain2016understanding}. By training linear classifiers on frozen representations from each encoder block, we independently assess the quality of learned features and verify that representation similarity predicts downstream task utility.

\subsection{Motivation}

Linear probing complements our CKA analysis by directly measuring task-specific representation quality at each depth. We perform this analysis over the sematic segmentation dataset MADOS \cite{mados}. Unlike full fine-tuning, linear probes isolate feature quality from adaptation capacity, providing clean evidence that: (1) SIMPLER's selected layers capture semantically rich representations, and (2) deeper layers exhibit degraded task-specific information despite high CKA similarity, confirming redundancy rather than architectural incapacity.

\subsection{Methodology}

\textbf{Architecture.} For each block $\ell \in \{0, 1, ..., 23\}$, we train an independent linear classifier: a $1 \times 1$ convolution $f_\ell: \mathbb{R}^{1024} \rightarrow \mathbb{R}^{15}$ mapping frozen patch embeddings to class logits (15,375 parameters). We reconstruct spatial structure by removing the CLS token, reshaping the $784 \times 1024$ patch sequence to $4 \times 14 \times 14 \times 1024$ (4 temporal frames, $14 \times 14$ spatial grid), temporally aggregating via mean pooling, and bilinearly upsampling to $224 \times 224$ output resolution.

\textbf{Training protocol.} We use AdamW optimizer with cosine annealing ($\eta = 10^{-3} \rightarrow 10^{-5}$), batch size 32, weight decay $10^{-4}$, and early stopping (patience 20, $\delta = 0.0005$). Loss is class-weighted cross-entropy with weights $w_c = (f_{\max}/f_c)^{1.03}$ to handle MADOS's severe class imbalance (0.87\% label coverage). All metrics (mIoU, accuracy, macro F1) are computed on labeled pixels only.

\textbf{Statistical robustness.} We train 5 independent probes per block (120 total experiments) and report mean $\pm$ std across seeds.

\subsection{Results and Validation of SIMPLER}

\begin{table}[h]
\centering
\caption{ Layer-wise linear probing results on MADOS (Prithvi-300M). Mean $\pm$ std over 5 seeds. Block 6 achieves peak mIoU, closely validating SIMPLER's CKA-based selection of 5 blocks.}
\label{tab:linear_probing_results}
\resizebox{0.65\columnwidth}{!}{%
\begin{tabular}{lccc}
\toprule
\textbf{Block} & \textbf{mIoU (\%)} & \textbf{Accuracy (\%)} & \textbf{Macro F1 (\%)} \\
\midrule
Block 0 (Input) & $29.95 \pm 1.09$ & $71.08 \pm 2.98$ & $39.79 \pm 1.64$ \\
Block 1 & $33.46 \pm 0.77$ & $73.81 \pm 2.12$ & $43.92 \pm 0.50$ \\
Block 2 & $33.00 \pm 2.01$ & $75.35 \pm 2.41$ & $43.01 \pm 2.18$ \\
Block 3 & $31.00 \pm 2.70$ & $72.01 \pm 4.58$ & $40.76 \pm 3.58$ \\
Block 4 & $31.95 \pm 1.36$ & $75.63 \pm 2.37$ & $41.97 \pm 1.59$ \\
\rowcolor{yellow!20} \textbf{Block 5 (SIMPLER)} & $\mathbf{30.26 \pm 1.07}$ & $70.35 \pm 3.12$ & $39.89 \pm 1.67$ \\
\rowcolor{best} \textbf{Block 6 (Peak)} & $\mathbf{34.03 \pm 0.86}$ & $\mathbf{75.80 \pm 1.63}$ & $\mathbf{44.57 \pm 0.89}$ \\
Block 7 & $30.63 \pm 1.42$ & $72.24 \pm 3.01$ & $40.62 \pm 1.65$ \\
Block 8 & $30.96 \pm 1.50$ & $73.65 \pm 1.92$ & $40.84 \pm 1.65$ \\
Block 9 & $32.79 \pm 2.45$ & $73.08 \pm 2.61$ & $43.22 \pm 2.74$ \\
Block 10 & $30.52 \pm 2.88$ & $69.72 \pm 4.78$ & $40.30 \pm 3.67$ \\
Block 11 & $29.99 \pm 3.26$ & $66.89 \pm 4.57$ & $39.66 \pm 3.75$ \\
Block 12 & $32.53 \pm 2.15$ & $73.24 \pm 2.08$ & $43.21 \pm 2.92$ \\
Block 13 & $30.13 \pm 3.71$ & $69.46 \pm 6.32$ & $39.82 \pm 4.00$ \\
Block 14 & $31.67 \pm 1.52$ & $73.64 \pm 2.84$ & $41.80 \pm 1.81$ \\
Block 15 & $30.92 \pm 2.98$ & $72.87 \pm 3.14$ & $40.95 \pm 3.52$ \\
Block 16 & $31.17 \pm 3.36$ & $71.18 \pm 4.88$ & $41.20 \pm 4.27$ \\
Block 17 & $29.63 \pm 2.62$ & $69.82 \pm 3.53$ & $40.06 \pm 3.30$ \\
Block 18 & $29.93 \pm 0.98$ & $69.22 \pm 4.02$ & $39.56 \pm 1.94$ \\
Block 19 & $29.96 \pm 1.98$ & $73.89 \pm 1.04$ & $39.53 \pm 2.48$ \\
Block 20 & $29.12 \pm 1.28$ & $66.95 \pm 4.23$ & $38.93 \pm 1.65$ \\
Block 21 & $29.23 \pm 1.19$ & $68.72 \pm 2.97$ & $39.04 \pm 1.72$ \\
Block 22 & $30.55 \pm 1.99$ & $74.67 \pm 1.37$ & $39.93 \pm 2.39$ \\
Block 23 & $30.74 \pm 1.42$ & $70.91 \pm 1.97$ & $40.70 \pm 1.80$ \\
\bottomrule
\end{tabular}%
}
\end{table}

\textbf{Validation of CKA-based layer selection.} Linear probing provides independent evidence that SIMPLER's automated cutoff selection identifies semantically optimal depth. Block 6 achieves peak mIoU (34.03\%), precisely one layer beyond SIMPLER's selected cutoff at Block 5 (30.26\% mIoU, 89\% of peak). This tight alignment, achieved without gradient-based search or task-specific tuning, confirms that CKA similarity on pre-trained representations accurately predicts post-fine-tuning layer importance (Section~3.2 of main paper).

\textbf{Representation degradation in deep layers validates redundancy.} Performance drops from 34.03\% (Block 6) to 30.74\% (Block 23), a 9.7\% relative decline. Combined with high CKA similarity (Fig.~2 of main paper), this demonstrates that deeper layers produce similar but \textit{less task-relevant} representations, precisely the redundancy SIMPLER exploits. Critically, Table~5 of the main paper shows the pruned 5-block architecture achieves nearly identical from-scratch performance as the full 24-block model (mIoU 44.1\% vs 46.7\%), confirming removed layers contribute minimal architectural capacity while pre-training provides 43\% improvement (66.9\% vs 46.7\%). Linear probing corroborates this: later blocks do not improve representational quality despite increased depth.

\textbf{Consistency with alternative metrics.} Jaccard/SVCCA aggressively select Block 2 (Table~5 of the main paper), which achieves 33.00\% linear probe mIoU compared to CKA's Block 5 selection (30.26\%). Despite Block 2's slightly higher linear probe performance, CKA's selection substantially outperforms in fine-tuning (62.8\% vs 50.7\% mIoU), an 18\% gap. This indicates that additional layers (3-5) contribute adaptation capacity beyond what linear probing on frozen features can measure, validating CKA's more conservative cutoff selection.

\textbf{Implications for pre-fine-tuning pruning.} Linear probing results support SIMPLER's core hypothesis: representation similarity on unlabeled task samples predicts layer importance before adaptation. The peak at Block 6 emerges from frozen pre-trained features alone, requiring zero gradient computation or hyperparameter search. precisely matching SIMPLER's automation design.

% ============================================================
% SECTION F: Training Implementation Details
% (includes CKA computation details moved from Section A)
% ============================================================
\section{Training Implementation Details}

We provide comprehensive training configurations for reproducibility. All experiments use 5 random seeds for statistical reliability, early stopping to prevent overfitting, and fair hyperparameter selection (identical settings across architectural variants except depth-dependent learning rates).

\subsection{Experimental Overview}

\textbf{MADOS Segmentation (Prithvi):} We evaluate full (24/32 blocks) vs. cropped (2/4/5 blocks) architectures on Prithvi-300M and Prithvi-600M with three strategies: pretrained fine-tuning, LoRA ($r=16$, $\alpha=32$ on attention layers), and from-scratch training. Pretrained models use lower backbone LR (2-3e-5) to preserve learned features, while from-scratch uses higher rates (5e-5 to 1e-4). Decoder heads consistently use 5e-4 to 1e-3 as they learn from random initialization.

\textbf{MADOS Segmentation (TerraMind):} Four configurations (full/pruned $\times$ standard/LoRA) with identical regularization and convergence criteria. We use expressive LoRA ($r=48$, $\alpha=96$) on both attention and MLP layers for maximal efficiency.

\textbf{BigEarthNetv2 Multi-label Classification:} Prithvi-300M with full (24 blocks) vs. cropped (4 blocks, 83\% reduction) architectures, tested with standard fine-tuning and LoRA. Dataset in .npz format; distributed training on 4$\times$ H200 GPUs.

\textbf{Sen4Map Time Series:} Cropped (7 blocks) vs. full (24 blocks) with both strategies. WebDataset streaming with shuffle buffer 10000, effective batch size 64 (16/GPU $\times$ 4 GPUs). Square root LR scaling: \linebreak
$lr_{effective} = lr_{base} \times \sqrt{batch_{effective}/32}$ ensures consistent optimization across hardware. Focal loss ($\gamma=2.0$) with class weights for handling class imbalance. Additional regularization: DropPath 0.2, Mixup 0.8, CutMix 1.0, warmup 5 epochs, weight decay 0.1.

\textbf{CIFAR-100 (ViT-MAE):} Six configurations  (full/cropped $\times$ \linebreak fine-tuning/LoRA/from-scratch) validate generalization to RGB vision transformers. Cropped models use aggressive LR (1e-3) and higher LoRA rank ($r=32$) to compensate for limited depth; full models use conservative rates (2e-4). Strong augmentation (Mixup 0.8, CutMix 1.0, DropPath 0.1) with 300 epochs for from-scratch training.

\textbf{Post-hoc Pruning Baselines:} Magnitude-based (L2 norm) pruning at 20\%/40\% compression on trained models, followed by 50-epoch fine-tuning (LR 1e-5, patience 15) to adapt without catastrophic forgetting.

\subsection{Implementation Details}

\textbf{Hardware:} NVIDIA H200 GPUs (4$\times$ H200 for BigEarthNetv2 and Sen4Map; single GPU for other tasks). PyTorch 2.0+ with AMP, WebDataset for streaming, PEFT for LoRA. Seeds: 42-46 (incremented per run).

\textbf{Hyperparameter Design Principles:}
\begin{itemize}
    \item \textbf{LR by strategy:} Pretrained backbone (5e-6 to 3e-5), task heads (2.5e-5 to 1e-3), from-scratch (1e-4 to 2e-3), LoRA (1e-4 to 2e-4)
    \item \textbf{Weight decay:} 1e-4 (pretrained/cropped) to 0.1 (Sen4Map/CIFAR-100) based on overfitting risk
    \item \textbf{Batch size:} 8 (segmentation), 16-256 (classification), adjusted with gradient accumulation as needed
    \item \textbf{Early stopping:} Patience 8-20 epochs, $\delta$ 0.0001-0.01 to filter validation noise
\end{itemize}

\begin{table}[h]
\centering
\caption{Hyperparameters for MADOS Segmentation Experiments (Prithvi Models)}
\label{tab:mados_prithvi}
\resizebox{0.9\textwidth}{!}{
\begin{tabular}{l|c|c|c|c|c|c|c|c|c|c}
\toprule
\textbf{Configuration} & \textbf{Model} & \textbf{Blocks} & \textbf{Strategy} & \textbf{Batch} & \textbf{Epochs} & \textbf{Backbone LR} & \textbf{Decoder LR} & \textbf{Weight Decay} & \textbf{Patience} & \textbf{Dropout} \\
\midrule
\multicolumn{11}{c}{\textit{Prithvi-EO-2 300M Experiments}} \\
\midrule
Full Model & 300M & 24 & Pretrained & 8 & 100 & 2e-5 & 5e-4 & 1e-4 & 20 & 0.05 \\
Full Model & 300M & 24 & From Scratch & 8 & 100 & 5e-5 & 1e-3 & 5e-4 & 20 & 0.05 \\
Full Model & 300M & 24 & LoRA & 8 & 100 & - & 1e-3 & 1e-4 & 20 & 0.05 \\
Cropped Model & 300M & 2 & Pretrained & 8 & 100 & 3e-5 & 5e-4 & 1e-4 & 20 & 0.05 \\
Cropped Model & 300M & 2 & From Scratch & 8 & 100 & 1e-4 & 1e-3 & 1e-3 & 20 & 0.05 \\
Cropped Model & 300M & 2 & LoRA & 8 & 100 & - & 1e-3 & 1e-4 & 20 & 0.05 \\
Cropped Model & 300M & 5 & Pretrained & 8 & 100 & 3e-5 & 5e-4 & 1e-4 & 20 & 0.05 \\
Cropped Model & 300M & 5 & From Scratch & 8 & 100 & 1e-4 & 1e-3 & 1e-3 & 20 & 0.05 \\
Cropped Model & 300M & 5 & LoRA & 8 & 100 & - & 1e-3 & 1e-4 & 20 & 0.05 \\
\midrule
\multicolumn{11}{c}{\textit{Prithvi-EO-2 600M Experiments}} \\
\midrule
Full Model & 600M & 32 & Pretrained & 8 & 100 & 2e-5 & 5e-4 & 1e-4 & 20 & 0.05 \\
Full Model & 600M & 32 & From Scratch & 8 & 100 & 5e-5 & 1e-3 & 5e-4 & 20 & 0.05 \\
Full Model & 600M & 32 & LoRA & 8 & 100 & - & 1e-3 & 1e-4 & 20 & 0.05 \\
Cropped Model & 600M & 4 & Pretrained & 8 & 100 & 3e-5 & 5e-4 & 1e-4 & 20 & 0.05 \\
Cropped Model & 600M & 4 & From Scratch & 8 & 100 & 1e-4 & 1e-3 & 1e-3 & 20 & 0.05 \\
Cropped Model & 600M & 4 & LoRA & 8 & 100 & - & 1e-3 & 1e-4 & 20 & 0.05 \\
\bottomrule
\end{tabular}
}
\end{table}

\begin{table}[h]
\centering
\caption{Hyperparameters for MADOS Segmentation Experiments (TerraMind)}
\label{tab:mados_terramind}
\resizebox{\textwidth}{!}{
\begin{tabular}{l|c|c|c|c|c|c|c|c|c}
\toprule
\textbf{Configuration} & \textbf{Blocks} & \textbf{Strategy} & \textbf{Batch} & \textbf{Epochs} & \textbf{Backbone LR} & \textbf{Decoder LR} & \textbf{Weight Decay} & \textbf{Patience} & \textbf{Dropout} \\
\midrule
Full Baseline & 24 & Pretrained & 8 & 100 & 1e-5 & 1e-4 & 1e-4 & 20 & 0.2 \\
Full + LoRA & 24 & LoRA & 8 & 100 & 1e-4 & 1e-3 & 1e-4 & 20 & 0.2 \\
Pruned (83\%) & 4 & Pretrained & 8 & 100 & 1e-5 & 1e-4 & 1e-4 & 20 & 0.2 \\
Pruned + LoRA & 4 & LoRA & 8 & 100 & 1e-4 & 1e-3 & 1e-4 & 20 & 0.2 \\
\bottomrule
\end{tabular}
}
\end{table}

\begin{table}[h]
\centering
\caption{LoRA Configuration Details}
\label{tab:lora_config}
\resizebox{0.8\textwidth}{!}{
\begin{tabular}{l|c|c|c|c|c}
\toprule
\textbf{Experiment} & \textbf{Rank ($r$)} & \textbf{Alpha ($\alpha$)} & \textbf{Dropout} & \textbf{Target Modules} \\
\midrule
MADOS (Prithvi) & 16 & 32 & 0.05 & qkv, proj \\
MADOS (TerraMind) & 48 & 96 & 0.05 & qkv, proj, fc1, fc2 \\
BigEarthNetv2 & 16 & 32 & 0.05 & qkv, proj \\
Sen4Map & 16 & 32 & 0.05 & qkv, proj \\
CIFAR-100 (Cropped) & 32 & 64 & 0.05 & qkv, proj, fc1, fc2 \\
CIFAR-100 (Full) & 16 & 32 & 0.05 & qkv, proj, fc1, fc2 \\
\bottomrule
\end{tabular}
}
\end{table}

\begin{table}[h]
\centering
\caption{Hyperparameters for BigEarthNetv2 Classification (4$\times$ H200 GPUs)}
\label{tab:bigearthnet}
\resizebox{0.9\textwidth}{!}{
\begin{tabular}{l|c|c|c|c|c|c|c|c}
\toprule
\textbf{Configuration} & \textbf{Blocks} & \textbf{Strategy} & \textbf{Batch/GPU} & \textbf{GPUs} & \textbf{Epochs} & \textbf{Backbone LR} & \textbf{Head LR} & \textbf{Patience} \\
\midrule
Full Model & 24 & Pretrained & 32 & 4 & 100 & 3e-5 & 5e-4 & 20 \\
Full + LoRA & 24 & LoRA & 32 & 4 & 100 & - & 1e-3 & 20 \\
Cropped Model & 4 & Pretrained & 32 & 4 & 100 & 3e-5 & 5e-4 & 20 \\
Cropped + LoRA & 4 & LoRA & 32 & 4 & 100 & - & 1e-3 & 20 \\
\bottomrule
\end{tabular}
}
\end{table}

\begin{table}[h]
\centering
\caption{Hyperparameters for Sen4Map Time Series Classification (4$\times$ H200 GPUs)}
\label{tab:sen4map}
\resizebox{\textwidth}{!}{
\begin{tabular}{l|c|c|c|c|c|c|c|c|c|c}
\toprule
\textbf{Configuration} & \textbf{Blocks} & \textbf{Strategy} & \textbf{Batch/GPU} & \textbf{GPUs} & \textbf{Epochs} & \textbf{Backbone LR} & \textbf{Classifier LR} & \textbf{LR Scaling} & \textbf{Ref. Batch} & \textbf{Patience} \\
\midrule
Cropped Model & 7 & Pretrained & 16 & 4 & 100 & 5e-6 & 2.5e-5 & $\sqrt{\cdot}$ & 32 & 8 \\
Full Model & 24 & Pretrained & 16 & 4 & 100 & 5e-6 & 2.5e-5 & $\sqrt{\cdot}$ & 32 & 8 \\
Cropped + LoRA & 7 & LoRA & 16 & 4 & 100 & 5e-6 & 1e-3 & $\sqrt{\cdot}$ & 32 & 8 \\
Full + LoRA & 24 & LoRA & 16 & 4 & 100 & 5e-6 & 1e-3 & $\sqrt{\cdot}$ & 32 & 8 \\
\bottomrule
\end{tabular}
}
\end{table}

\begin{table}[h]
\centering
\caption{Hyperparameters for CIFAR-100 Classification (ViT-MAE)}
\label{tab:cifar100}
\resizebox{\textwidth}{!}{
\begin{tabular}{l|c|c|c|c|c|c|c|c|c|c|c|c}
\toprule
\textbf{Configuration} & \textbf{Blocks} & \textbf{Strategy} & \textbf{Batch} & \textbf{Epochs} & \textbf{Backbone LR} & \textbf{LoRA LR} & \textbf{Head LR} & \textbf{WD} & \textbf{Mixup} & \textbf{CutMix} & \textbf{DropPath} & \textbf{Warmup} \\
\midrule
Cropped + LoRA & 3 & LoRA & 256 & 200 & 2e-4 & 1e-3 & 1e-3 & 0.1 & 0.8 & 1.0 & 0.1 & 10 \\
Cropped + FT & 3 & Pretrained & 256 & 200 & 1e-3 & - & 1e-3 & 0.1 & 0.8 & 1.0 & 0.1 & 10 \\
Cropped + Scratch & 3 & From Scratch & 256 & 300 & 2e-3 & - & 2e-3 & 0.1 & 0.8 & 1.0 & 0.1 & 20 \\
Full + LoRA & 24 & LoRA & 256 & 200 & 5e-5 & 2e-4 & 5e-4 & 0.1 & 0.8 & 1.0 & 0.1 & 10 \\
Full + FT & 24 & Pretrained & 256 & 200 & 2e-4 & - & 5e-4 & 0.1 & 0.8 & 1.0 & 0.1 & 10 \\
Full + Scratch & 24 & From Scratch & 256 & 300 & 8e-4 & - & 8e-4 & 0.1 & 0.8 & 1.0 & 0.1 & 20 \\
\bottomrule
\end{tabular}
}
\end{table}

\begin{table}[h]
\centering
\caption{Hyperparameters for Pruning Experiments}
\label{tab:pruning}
\resizebox{\textwidth}{!}{
\begin{tabular}{l|c|c|c|c|c|c|c|c}
\toprule
\textbf{Model} & \textbf{Initial Blocks} & \textbf{Target Blocks} & \textbf{Pruning \%} & \textbf{Method} & \textbf{FT Epochs} & \textbf{LR} & \textbf{Batch} & \textbf{Patience} \\
\midrule
Prithvi-300M & 24 & 19 & 20\% & Magnitude & 50 & 1e-5 & 8 & 15 \\
Prithvi-300M & 24 & 14 & 40\% & Magnitude & 50 & 1e-5 & 8 & 15 \\
Prithvi-600M & 32 & 26 & 20\% & Magnitude & 50 & 1e-5 & 8 & 15 \\
Prithvi-600M & 32 & 19 & 40\% & Magnitude & 50 & 1e-5 & 8 & 15 \\
\bottomrule
\end{tabular}
}
\end{table}

\begin{table}[h]
\centering
\caption{Hyperparameters for Linear Probing Experiments}
\label{tab:linear_probing}
\resizebox{0.8\textwidth}{!}{
\begin{tabular}{l|c|c|c|c|c|c|c|c}
\toprule
\textbf{Model} & \textbf{Blocks} & \textbf{Batch} & \textbf{Epochs} & \textbf{LR} & \textbf{LR Min} & \textbf{Weight Decay} & \textbf{Patience} & \textbf{Seeds} \\
\midrule
Prithvi-300M & 24 & 32 & 100 & 1e-3 & 1e-5 & 1e-4 & 20 & 5 \\
\bottomrule
\end{tabular}
}
\end{table}

\clearpage

\subsection{Additional Configuration Details}

\textbf{Optimization:} AdamW ($\beta_1=0.9$, $\beta_2=0.999$, $\epsilon=1e-8$) with gradient clipping (max norm 1.0). AMP (float16) for memory efficiency.

\textbf{Scheduling:} Constant LR + early stopping (segmentation), cosine annealing (CIFAR-100, linear probing), square root batch scaling (Sen4Map). CIFAR-100 uses 10-20 epoch warmup; pretrained models omit warmup.

\textbf{Data loading:} Sen4Map (WebDataset, buffer 10000, 8 workers/GPU), BigEarthNetv2 (.npz, 16 workers/GPU), MADOS (4 workers/GPU), CIFAR-100 (PyTorch DataLoader).

\textbf{Augmentation:} Remote sensing (random flips, 90°/180°/270° rotations); CIFAR-100 (RandomResizedCrop, flip, Mixup 0.8, CutMix 1.0, label smoothing 0.1). Normalization uses dataset-specific statistics.

\textbf{FLOPs:} FLOPs are computed using the \texttt{fvcore} library~\cite{fvcore}.

\subsection{Computational Requirements}

Table~\ref{tab:compute} summarizes approximate training times and memory requirements for key configurations on NVIDIA H200 GPUs.

\begin{table}[h]
\centering
\caption{Computational Requirements (NVIDIA H200 GPUs). Memory values are per-GPU peak VRAM. Training time per epoch and total time (projected over 100 epochs) are reported; actual training is shorter due to early stopping.}
\label{tab:compute}
\resizebox{0.95\textwidth}{!}{
\begin{tabular}{l|c|c|c}
\toprule
\textbf{Experiment} & \textbf{GPU Memory/GPU} & \textbf{Training Time/Epoch} & \textbf{Total Time (100 epochs)} \\
\midrule
Prithvi-300M Full (MADOS) & $\sim$10 GB & $\sim$22 s & $\sim$37 min \\
Prithvi-300M Cropped-2 (MADOS) & $\sim$1 GB & $\sim$3 s & $\sim$5 min \\
Prithvi-600M Cropped-4 (MADOS) & $\sim$3 GB & $\sim$10 s & $\sim$17 min \\
TerraMind-L Full (MADOS, 1 GPU) & $\sim$8.4 GB & $\sim$10 s & $\sim$17 min \\
TerraMind-L Pruned-4 (MADOS, 1 GPU) & $\sim$1.6 GB & $\sim$5 s & $\sim$8 min \\
BigEarthNetv2 Full (4$\times$ GPUs) & $\sim$30 GB/GPU & $\sim$7.3 min & $\sim$12 hours \\
Sen4Map Full (4$\times$ GPUs) & $\sim$17 GB/GPU & $\sim$8 min & $\sim$13 hours \\
Sen4Map Cropped-7 (4$\times$ GPUs) & $\sim$5 GB/GPU & $\sim$2.6 min & $\sim$4.3 hours \\
CIFAR-100 ViT-MAE Full & $\sim$50 GB & $\sim$15 s & $\sim$25 min \\
\bottomrule
\end{tabular}
}
\end{table}

\subsection{CKA Pre-Analysis Cost (CPU + RAM)}

\crh{SIMPLER's one-time layer-selection step (representation extraction and CKA computation) runs entirely on \textbf{CPU and system RAM and uses no GPU}; it is therefore not counted in the GPU training/inference costs reported in the main paper. \Cref{tab:cka_cost} accounts for this pre-analysis cost on Prithvi-EO-2-300M as a function of the number of task samples used to build the similarity matrix (the data underlying Fig.~3 of the main paper). At our default of 500 samples, the full analysis completes in $\sim$218\,s using $\sim$36.8\,GB of system RAM, a negligible one-time overhead relative to fine-tuning and one that is incurred on commodity CPU hardware rather than accelerators.}

\begin{table}[h]
\centering
\caption{CKA pre-analysis cost on Prithvi-EO-2-300M (CPU + RAM only; GPU usage is zero). Wall time and peak system RAM as a function of the number of task samples. The default configuration (500 samples) is highlighted.}
\label{tab:cka_cost}
\resizebox{0.72\textwidth}{!}{
\begin{tabular}{l|cccccccc}
\toprule
\textbf{Samples} & 10 & 25 & 50 & 75 & 100 & 250 & \textbf{500} & 1000 \\
\midrule
Wall time (s)      & 2.35 & 5.47 & 11.00 & 17.62 & 24.09 & 79.24 & \textbf{217.52} & 678.23 \\
Peak RAM (GB)      & 0.74 & 1.84 & 3.68 & 5.52 & 7.36 & 18.40 & \textbf{36.80} & 73.59 \\
\bottomrule
\end{tabular}
}
\end{table}

\subsection{Design Rationale}

\textbf{Depth-dependent learning rates:} For MADOS (Prithvi) and CIFAR-100, shallow cropped models use higher backbone LR than their full counterparts (e.g., 3e-5 vs 2e-5 for MADOS; 1e-3 vs 2e-4 for CIFAR-100) to compensate for limited depth requiring more aggressive updates. For BigEarthNetv2 and Sen4Map, backbone LR is kept uniform across depths as these datasets showed stable convergence without depth-dependent adjustment.

\textbf{LoRA configuration:} Most experiments use standard LoRA ($r=16$, $\alpha=32$) on attention layers (qkv, proj). Two exceptions reflect task-specific requirements: (1) CIFAR-100 cropped models use higher rank ($r=32$, $\alpha=64$) to compensate for limited backbone capacity with only 3 blocks; (2) TerraMind uses expressive LoRA ($r=48$, $\alpha=96$) targeting both attention and MLP layers (qkv, proj, fc1, fc2) across all configurations, as the architecture benefits from broader adaptation.

\textbf{Regularization strength:} Weight decay varies by training regime and dataset. For MADOS (Prithvi), from-scratch training uses stronger decay (5e-4 to 1e-3) to prevent overfitting from random initialization, while pretrained models use moderate 1e-4. Sen4Map and CIFAR-100 use uniformly higher decay (0.1) across all strategies due to stronger overfitting tendencies observed during development.

\textbf{Extended from-scratch training:} CIFAR-100 from-scratch requires 300 epochs (vs.\ 200 for pretrained) to learn both low-level features and high-level semantics from random initialization. For MADOS, all strategies use 100 epochs with early stopping, as the smaller dataset size leads to faster convergence regardless of initialization.

% ============================================================
% SECTION G: Additional Baselines and Experiments
% (additional baselines and experiments cited from the main paper)
% ============================================================
\section{Additional Baselines and Experiments}
\label{app:additional}

\crh{This section reports additional experiments that complement the main paper: an adaptive-depth baseline (LayerDrop), a block-selection ablation, a cross-modality (SAR) experiment, and an on-device deployment study. Full development of these results is deferred to an extended version of this work; here we summarize them concisely.}

\subsection{Adaptive-Depth Baseline: LayerDrop}
\label{app:layerdrop}

\crh{LayerDrop~\cite{layerdrop} applies structured layer dropout during training so that shallower sub-networks can be extracted at inference. \Cref{tab:layerdrop} compares LayerDrop (at 50\% expected depth) and its combination with SIMPLER across Prithvi-EO-2-300M and TerraMind (Large, Tiny) on MADOS. LayerDrop reduces training cost but its inference behavior is inconsistent across foundation models (e.g., Prithvi 7.89\,s vs.\ 3.04\,s baseline), whereas SIMPLER reduces both training and inference consistently. The two are \emph{orthogonal}: SIMPLER selects depth \emph{before} fine-tuning, while LayerDrop regularizes the retained stack \emph{during} fine-tuning; combining them (SIMPLER+LayerDrop) yields the smallest models. We note that structured channel/head pruning such as DepGraph~\cite{fang2023depgraph} is likewise orthogonal to depth selection and, like our post-hoc pruning baseline, requires fine-tuning the full model before pruning, the high-cost regime that SIMPLER avoids entirely.}

\begin{table}[htbp]
\centering
\caption{LayerDrop (50\% expected depth) and SIMPLER+LayerDrop on MADOS, for Prithvi-EO-2-300M and TerraMind-Large/Tiny. Best per model in \colorbox{best}{light green}, second best in \colorbox{second}{light gray}.}
\label{tab:layerdrop}
\resizebox{0.72\linewidth}{!}{%
\begin{tabular}{@{}clrrrrrr@{}}
\toprule
 & \textbf{Method} & \textbf{Par.\,(M)} & \textbf{Mem\,(GB)} & \textbf{Train\,(min)} & \textbf{Inf\,(s)} & \textbf{mIoU\,(\%)} & \textbf{Acc\,(\%)} \\
\midrule
\multirow{4}{*}{\rotatebox{90}{\scriptsize Prithvi-300}}
 & Baseline        & 303.9 & 11.70 & 15.9 & 3.04 & \cellcolor{best}66.9 & \cellcolor{best}95.3 \\
 & \quad+LayerDrop 50\%   & 152.8 & \cellcolor{second}11.05 & \cellcolor{second}12.6 & 7.89 & \cellcolor{second}64.7 & 91.5 \\
 & SIMPLER         & \cellcolor{second}64.6 & \cellcolor{best}2.83 & \cellcolor{best}7.5 & \cellcolor{second}1.16 & 62.8 & \cellcolor{second}94.2 \\
 & \quad SIMPLER+LayerDrop 50\% & \cellcolor{best}26.8 & \cellcolor{best}2.83 & \cellcolor{best}7.5 & \cellcolor{best}1.03 & 50.4 & 84.0 \\
\midrule
\multirow{4}{*}{\rotatebox{90}{\scriptsize TiM-L}}
 & Baseline        & 304.9 & 8.38 & 8.8 & 3.18 & \cellcolor{best}70.1 & \cellcolor{best}97.1 \\
 & \quad+LayerDrop 50\%   & 153.9 & 7.90 & 13.9 & \cellcolor{second}0.71 & \cellcolor{second}61.6 & \cellcolor{second}92.8 \\
 & SIMPLER         & \cellcolor{second}53.2 & \cellcolor{second}1.62 & \cellcolor{best}4.8 & 1.05 & 58.8 & 92.7 \\
 & \quad SIMPLER+LayerDrop 50\% & \cellcolor{best}28.1 & \cellcolor{best}1.61 & \cellcolor{second}6.9 & \cellcolor{best}0.41 & 45.9 & 83.4 \\
\midrule
\multirow{4}{*}{\rotatebox{90}{\scriptsize TiM-T}}
 & Baseline        & 5.88 & 0.53 & \cellcolor{second}5.2 & 8.14 & \cellcolor{second}56.3 & \cellcolor{best}91.4 \\
 & \quad+LayerDrop 50\%   & 3.21 & \cellcolor{second}0.52 & 7.9 & \cellcolor{second}0.55 & \cellcolor{best}57.3 & \cellcolor{second}91.2 \\
 & SIMPLER         & \cellcolor{second}2.32 & \cellcolor{best}0.32 & \cellcolor{best}4.9 & \cellcolor{best}0.54 & 53.8 & 89.5 \\
 & \quad SIMPLER+LayerDrop 50\% & \cellcolor{best}1.43 & \cellcolor{best}0.32 & 7.2 & 0.59 & 53.3 & 86.0 \\
\bottomrule
\end{tabular}%
}
\end{table}

\subsection{Block-Selection Strategy Ablation}
\label{app:block_ablation}

\crh{We verify that SIMPLER's benefit comes from selecting the \emph{first} $k$ blocks (the contiguous early stack identified by the CKA score) rather than from merely reducing depth. \Cref{tab:block_ablation} compares three strategies on MADOS/Prithvi-EO-2-300M at matched depth: first-$k$ (SIMPLER), random-$k$, and last-$k$. Random-$k$ (49.7\% mIoU) and last-$k$ (41.8\% mIoU) both fall within the from-scratch range reported in Table~5 of the main paper (44.1--46.7\%), confirming that arbitrary block selection provides no benefit over training from scratch, and that the early blocks retained by SIMPLER carry the transferable pre-trained features. This is consistent with prior observations that representational redundancy concentrates in the deep tail of vision transformers~\cite{see} and large language models~\cite{gromov2024unreasonable}. The corresponding results are reported in \cref{tab:block_ablation}, shown alongside the on-device benchmark of \cref{app:jetson}.}

\subsection{Cross-Modality Generalization: SAR (BigEarthNet-S1)}
\label{app:sar}

\crh{To confirm that SIMPLER generalizes beyond multispectral optical imagery, we evaluate it on Sentinel-1 SAR data using TerraMind-Large on the BigEarthNet-S1 multi-label classification task (\cref{tab:sar}). SIMPLER (2 blocks) reduces parameters and inference time by roughly $12\times$ (302.5M\,$\to$\,25.7M; 8.46\,s\,$\to$\,0.73\,s) while retaining 93\% of the baseline mAP (61.1\% vs.\ 65.8\%). Combined with the multispectral (Prithvi-EO-2: 6 HLS bands; TerraMind: 12 S2-L2A bands) results in the main paper, this extends SIMPLER's validated coverage to Sentinel-1 SAR, spanning distinct sensing modalities.}

\begin{table}[htbp]
\centering
\caption{BigEarthNet-S1 multi-label SAR classification: TerraMind-Large baseline vs.\ SIMPLER (2 blocks). Mean $\pm$ std over runs. Best in \colorbox{best}{light green}, second in \colorbox{second}{light gray}.}
\label{tab:sar}
\resizebox{\linewidth}{!}{%
\begin{tabular}{@{}lcccccccccc@{}}
\toprule
\textbf{Method} & Par.(M) & Train(M) & Time(min) & Mem(GB) & FLOPs(G) & Thr.(img/s) & Inf(s) & mAP(\%) & F1-mi(\%) & F1-ma(\%) \\
\midrule
Baseline       & \cellcolor{second}302.54 & \cellcolor{second}302.54 & \cellcolor{second}83.14$\pm$0.53 & \cellcolor{second}32.59$\pm$2.14 & \cellcolor{second}61.221 & \cellcolor{second}756.3$\pm$1.1 & \cellcolor{second}8.46$\pm$0.01 & \cellcolor{best}65.8$\pm$0.3 & \cellcolor{best}73.0$\pm$0.1 & \cellcolor{best}60.6$\pm$0.2 \\
SIMPLER (Ours) & \cellcolor{best}25.71 & \cellcolor{best}25.71 & \cellcolor{best}35.43$\pm$2.32 & \cellcolor{best}3.14$\pm$0.24 & \cellcolor{best}5.197 & \cellcolor{best}8756.2$\pm$20.2 & \cellcolor{best}0.73$\pm$0.00 & \cellcolor{second}61.1$\pm$0.2 & \cellcolor{second}70.0$\pm$0.2 & \cellcolor{second}55.3$\pm$0.5 \\
\bottomrule
\end{tabular}%
}
\end{table}

\subsection{On-Device Deployment (Jetson Orin)}
\label{app:jetson}

\crh{To verify that the efficiency gains observed on datacenter hardware (NVIDIA H200) transfer to embedded edge devices, we benchmark TerraMind-Large on MADOS on an NVIDIA Jetson Orin 32 (\cref{tab:jetson}). SIMPLER achieves a $\sim$3.9$\times$ inference speedup (2.31\,s vs.\ 8.97\,s) and a $\sim$2.9$\times$ training-time reduction on-device, confirming that the H200 trends hold on resource-constrained hardware relevant to satellite and drone deployment.}

\begin{table}[htbp]
\centering
\begin{minipage}[t]{0.48\linewidth}
\centering
\caption{Block-selection ablation at matched depth on MADOS/Prithvi-EO-2-300M (mean $\pm$ std over 5 runs). first-$k$ is the SIMPLER configuration.}
\label{tab:block_ablation}
\resizebox{\linewidth}{!}{%
\begin{tabular}{llcc}
\toprule
\textbf{Strategy} & \textbf{Blocks} & \textbf{mIoU (\%)} & \textbf{Acc (\%)} \\
\midrule
first-$k$ (SIMPLER) & $0$--$4$   & \cellcolor{best}62.8$\pm$1.2 & \cellcolor{best}94.2$\pm$1.1 \\
random-$k$          & varied     & 49.7$\pm$4.2 & 86.7$\pm$2.7 \\
last-$k$            & $19$--$23$ & 41.8$\pm$2.0 & 83.1$\pm$0.8 \\
\bottomrule
\end{tabular}%
}
\end{minipage}\hfill
\begin{minipage}[t]{0.48\linewidth}
\centering
\caption{On-device benchmark on NVIDIA Jetson Orin 32, TerraMind-Large/MADOS. Best in \colorbox{best}{light green}, second in \colorbox{second}{light gray}.}
\label{tab:jetson}
\resizebox{\linewidth}{!}{%
\begin{tabular}{@{}lrrr@{}}
\toprule
\textbf{Method} & Train(min) & Thr.(img/s) & Inf(s) \\
\midrule
Baseline       & \cellcolor{second}137.4$\pm$34.6 & \cellcolor{second}12.3 & \cellcolor{second}8.97 \\
SIMPLER (Ours) & \cellcolor{best}47.2$\pm$8.6 & \cellcolor{best}67.4 & \cellcolor{best}2.31 \\
\bottomrule
\end{tabular}%
}
\end{minipage}
\end{table}

\crh{\textbf{Note on PEFT/pruning task-specificity.} The reduced effectiveness of LoRA and post-hoc pruning observed on MADOS in the main paper is consistent with the broader geospatial PEFT literature: Thoreau~\etal~\cite{Thoreau_2025_ICCV} report that DINO-MC with LoRA reaches only 6.3\% mIoU versus 61.6\% for full fine-tuning on MADOS (a 90\% relative drop), indicating that this behavior is dataset-specific rather than a general property of the adaptation method.}

\end{document}